\documentclass[superscriptaddress,longbibliography,aps,prl,twocolumn,10pt]{revtex4-2}
\usepackage[T1]{fontenc}
\usepackage[utf8]{inputenc}
\usepackage{fouriernc}
\usepackage{graphicx}
\usepackage{booktabs}
\usepackage{amsmath,amssymb,amsfonts,MnSymbol}
\usepackage{silence}
\usepackage{xcolor}
\definecolor{darkmagenta}{HTML}{8b008b}
\usepackage[colorlinks=true, linkcolor=darkmagenta, citecolor=darkmagenta, urlcolor=darkmagenta]{hyperref}
\usepackage{nameref}
\usepackage{microtype}
\usepackage{bm}
\usepackage{color, colortbl}
\usepackage{multirow}
\usepackage{stackengine}
\usepackage{nicefrac}
\newcommand{\comment}[1]{}
\newcommand{\norm}[1]{\left\lVert#1\right\rVert}
\usepackage{siunitx}
\usepackage{tcolorbox}
\usepackage{titlesec}
\usepackage{fancyhdr}

\renewcommand{\arraystretch}{1.0}

\titlespacing*{\section}{0pt}{18pt}{5pt}
\titlespacing*{\subsection}{0pt}{8pt}{2pt}
\usepackage{silence}
\ErrorsOff*

\begin{document}
\title{\LARGE{Novel computational workflows for natural and biomedical  \\ image processing based on hypercomplex algebras}}

\author{\normalsize{Nektarios A. Valous*}}
\affiliation{\scriptsize{Applied Tumor Immunity Clinical Cooperation Unit, National Center for Tumor Diseases (NCT) Heidelberg, German Cancer Research Center (DKFZ), Im Neuenheimer Feld 460, 69120, Heidelberg, Germany}}
\affiliation{\scriptsize{Medical Faculty Heidelberg, Heidelberg University, Department of Medical Oncology, Heidelberg University Hospital (UKHD), Im Neuenheimer Feld 672, 69120, Heidelberg, Germany}}
\affiliation{\scriptsize{Center for Quantitative Analysis of Molecular and Cellular Biosystems (Bioquant), Heidelberg University, Im Neuenheimer Feld 267, 69120, Heidelberg, Germany}}
\author{\normalsize{Eckhard Hitzer}}
\affiliation{\scriptsize{College of Liberal Arts, International Christian University, Osawa 3-10-2, 181-8585 Mitaka, Tokyo, Japan}}
\author{\normalsize{Drago\c{s} Du\c{s}e}}
\affiliation{\scriptsize{Department of Medical Oncology, National Center for Tumor Diseases (NCT) Heidelberg, Heidelberg University Hospital (UKHD), Im Neuenheimer Feld 460, 69120, Heidelberg, Germany}}
\affiliation{\scriptsize{Synaptiq, Str. Clinicilor 23, 400006, Cluj-Napoca, Romania}}
\author{\normalsize{Rodrigo Rojas Moraleda}}
\affiliation{\scriptsize{Applied Tumor Immunity Clinical Cooperation Unit, National Center for Tumor Diseases (NCT) Heidelberg, German Cancer Research Center (DKFZ), Im Neuenheimer Feld 460, 69120, Heidelberg, Germany}}
\affiliation{\scriptsize{Center for Quantitative Analysis of Molecular and Cellular Biosystems (Bioquant), Heidelberg University, Im Neuenheimer Feld 267, 69120, Heidelberg, Germany}}
\author{\normalsize{Ferdinand Popp}}
\affiliation{\scriptsize{Applied Tumor Immunity Clinical Cooperation Unit, National Center for Tumor Diseases (NCT) Heidelberg, German Cancer Research Center (DKFZ), Im Neuenheimer Feld 460, 69120, Heidelberg, Germany}}
\affiliation{\scriptsize{Division of Applied Bioinformatics, German Cancer Research Center (DKFZ), Im Neuenheimer Feld 280, 69120, Heidelberg, Germany}}
\author{\normalsize{Meggy Suarez-Carmona}}
\affiliation{\scriptsize{Department of Cancer Immunology \& Cancer Immunotherapy, German Cancer Research Center (DKFZ), Im Neuenheimer Feld 280, 69120, Heidelberg, Germany}}
\affiliation{\scriptsize{Tumor Immunology and Tumor Immunotherapy Group, Helmholtz Institute for Translational Oncology (HI-TRON), Obere Zahlbacher Stra{\ss}e 63, Building 911, 55131, Mainz, Germany}}
\author{\normalsize{Anna Berthel}}
\affiliation{\scriptsize{Department of Cancer Immunology \& Cancer Immunotherapy, German Cancer Research Center (DKFZ), Im Neuenheimer Feld 280, 69120, Heidelberg, Germany}}
\affiliation{\scriptsize{Tumor Immunology and Tumor Immunotherapy Group, Helmholtz Institute for Translational Oncology (HI-TRON), Obere Zahlbacher Stra{\ss}e 63, Building 911, 55131, Mainz, Germany}}
\author{\normalsize{Ismini Papageorgiou}}
\affiliation{\scriptsize{Institute of Radiology, S{\"u}dharz Hospital Nordhausen, Dr.-Robert-Koch-Stra{\ss}e 39, 99734, Nordhausen, Germany}}
\affiliation{\scriptsize{Institute of Diagnostic and Interventional Radiology, Jena University Hospital, Am Klinikum 1, 07747, Jena, Germany}}
\author{\normalsize{Carlo Fremd}}
\affiliation{\scriptsize{Medical Faculty Heidelberg, Heidelberg University, Department of Medical Oncology, Heidelberg University Hospital (UKHD), Im Neuenheimer Feld 672, 69120, Heidelberg, Germany}}
\affiliation{\scriptsize{Department of Medical Oncology, National Center for Tumor Diseases (NCT) Heidelberg, Heidelberg University Hospital (UKHD), Im Neuenheimer Feld 460, 69120, Heidelberg, Germany}}
\affiliation{\scriptsize{Division of Gynecological Oncology, National Center for Tumor Diseases (NCT) Heidelberg, Heidelberg University Hospital (UKHD), Im Neuenheimer Feld 460, 69120, Heidelberg, Germany}}
\affiliation{\scriptsize{Division of Molecular Genetics, German Cancer Research Center (DKFZ), Im Neuenheimer Feld 280, 69120, Heidelberg, Germany}}
\author{\normalsize{Alexander R{\"o}lle}}
\affiliation{\scriptsize{Medical Faculty Heidelberg, Heidelberg University, Department of Medical Oncology, Heidelberg University Hospital (UKHD), Im Neuenheimer Feld 672, 69120, Heidelberg, Germany}}
\affiliation{\scriptsize{Department of Medical Oncology, National Center for Tumor Diseases (NCT) Heidelberg, Heidelberg University Hospital (UKHD), Im Neuenheimer Feld 460, 69120, Heidelberg, Germany}}
\affiliation{\scriptsize{Applied Tumor Immunity Clinical Cooperation Unit, National Center for Tumor Diseases (NCT) Heidelberg, German Cancer Research Center (DKFZ), Im Neuenheimer Feld 460, 69120, Heidelberg, Germany}}
\author{\normalsize{Christina C. Westhoff}}
\affiliation{\scriptsize{Institute of Pathology, Philipps University of Marburg and University Hospital Giessen and Marburg GmbH (UKGM), Baldingerstrasse, 35033, Marburg, Germany}}
\author{\small{B{\'e}n{\'e}dicte Lenoir}}
\affiliation{\scriptsize{Applied Tumor Immunity Clinical Cooperation Unit, National Center for Tumor Diseases (NCT) Heidelberg, German Cancer Research Center (DKFZ), Im Neuenheimer Feld 460, 69120, Heidelberg, Germany}}
\affiliation{\scriptsize{Systems Immunology and Single-Cell Biology Group, German Cancer Research Center (DKFZ), Im Neuenheimer Feld 280, 69120, Heidelberg, Germany}}
\author{\normalsize{Niels Halama}}
\affiliation{\scriptsize{Department of Cancer Immunology \& Cancer Immunotherapy, German Cancer Research Center (DKFZ), Im Neuenheimer Feld 280, 69120, Heidelberg, Germany}}
\affiliation{\scriptsize{Tumor Immunology and Tumor Immunotherapy Group, Helmholtz Institute for Translational Oncology (HI-TRON), Obere Zahlbacher Stra{\ss}e 63, Building 911, 55131, Mainz, Germany}}
\affiliation{\scriptsize{University Center for Tumor Diseases Mainz (UCT Mainz), University Medical Center of Johannes Gutenberg University Mainz, Langenbeckstr{\ss}e 1, 55131, Mainz, Germany}}
\affiliation{\scriptsize{Department of Hematology and Medical Oncology, III. Medical Clinic and Polyclinic, University Medical Center of Johannes Gutenberg University Mainz, Langenbeckstr{\ss}e 1, 55131, Mainz, Germany}}
\author{\normalsize{Inka Z{\"o}rnig}}
\affiliation{\scriptsize{Medical Faculty Heidelberg, Heidelberg University, Department of Medical Oncology, Heidelberg University Hospital (UKHD), Im Neuenheimer Feld 672, 69120, Heidelberg, Germany}}
\affiliation{\scriptsize{Department of Medical Oncology, National Center for Tumor Diseases (NCT) Heidelberg, Heidelberg University Hospital (UKHD), Im Neuenheimer Feld 460, 69120, Heidelberg, Germany}}
\affiliation{\scriptsize{Applied Tumor Immunity Clinical Cooperation Unit, National Center for Tumor Diseases (NCT) Heidelberg, German Cancer Research Center (DKFZ), Im Neuenheimer Feld 460, 69120, Heidelberg, Germany}}
\author{\normalsize{Dirk J{\"a}ger}}
\affiliation{\scriptsize{Medical Faculty Heidelberg, Heidelberg University, Department of Medical Oncology, Heidelberg University Hospital (UKHD), Im Neuenheimer Feld 672, 69120, Heidelberg, Germany}}
\affiliation{\scriptsize{Department of Medical Oncology, National Center for Tumor Diseases (NCT) Heidelberg, Heidelberg University Hospital (UKHD), Im Neuenheimer Feld 460, 69120, Heidelberg, Germany}}
\affiliation{\scriptsize{Applied Tumor Immunity Clinical Cooperation Unit, National Center for Tumor Diseases (NCT) Heidelberg, German Cancer Research Center (DKFZ), Im Neuenheimer Feld 460, 69120, Heidelberg, Germany}}
\affiliation{\scriptsize{Center for Quantitative Analysis of Molecular and Cellular Biosystems (Bioquant), Heidelberg University, Im Neuenheimer Feld 267, 69120, Heidelberg, Germany}}

\vspace*{3em}

\begin{abstract}
\small{Hypercomplex image processing extends conventional techniques in a unified paradigm encompassing algebraic and geometric principles. This work leverages quaternions and the two-dimensional orthogonal planes split framework (splitting of a quaternion - representing a pixel - into pairs of orthogonal two-dimensional planes) for natural and biomedical image analysis through the following computational workflows and outcomes: natural and biomedical image re-colorization, natural image de-colorization, natural and biomedical image contrast enhancement, computational re-staining and stain separation in histological images, and performance gains in machine/deep learning pipelines for histological images. The workflows are analyzed separately for natural and biomedical images to showcase the effectiveness of the proposed approaches in each instance. The proposed workflows can regulate color appearance (e.g. with alternative renditions and grayscale conversion) and image contrast, be part of automated image processing pipelines (e.g. isolating stain components, boosting learning models), and assist in digital pathology applications (e.g. enhancing biomarker visibility, enabling colorblind-friendly renditions). Employing only basic arithmetic and matrix operations, this work offers a computationally accessible methodology - in the hypercomplex domain - that showcases versatility and consistency across image processing tasks and a range of computer vision and biomedical applications. Furthermore, the proposed non-data-driven methods achieve comparable or better results (particularly in cases involving well-known methods) to those reported in the literature, showcasing the potential of robust theoretical frameworks with practical effectiveness. Results, methods, and limitations are detailed alongside discussion of promising extensions, emphasizing the notable potential of feature-rich mathematical/computational frameworks for natural and biomedical images.}
\\
\\
\small{*\textbf{corresponding author}: Nektarios A. Valous; \textcolor{teal}{nek.valous@nct-heidelberg.de}, \textcolor{teal}{nvllnvll@gmail.com}.}
\end{abstract}

\begin{figure*}[t]
    \centering
    \begin{tcolorbox}[width=\textwidth,colframe=black!25,colback=white!0,boxrule=1pt,coltitle=black]
        \includegraphics[width=1.0\textwidth]{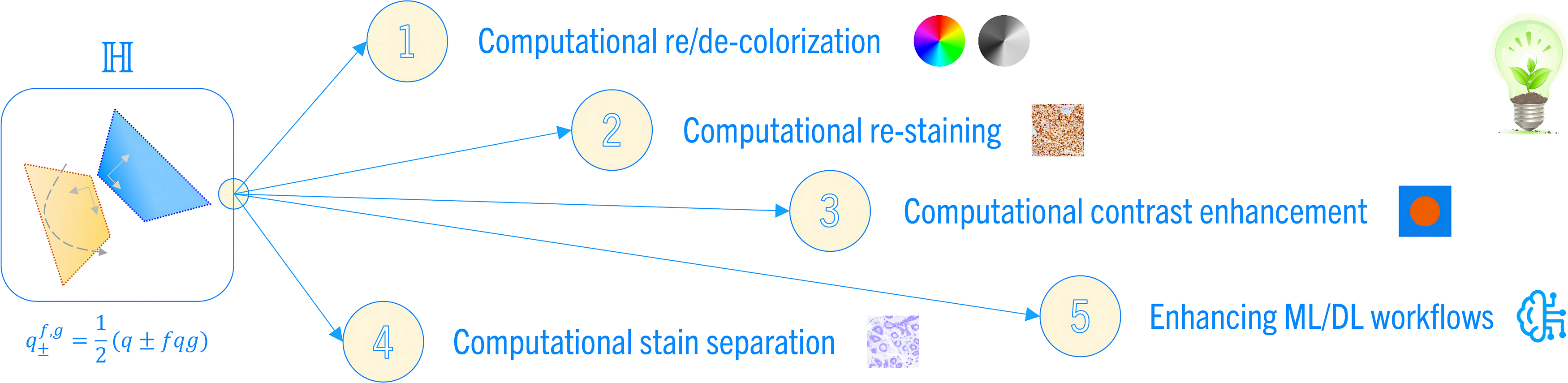}
    \end{tcolorbox}
\end{figure*}

\maketitle

\section*{\large\uppercase{Introduction}}

\begin{tcolorbox}[colframe=black!25,colback=orange!15,boxrule=1pt,coltitle=black]
\small{Quaternions - a type of hypercomplex number - are particularly useful in handling three-dimensional data, i.e. color images. Color pixels can be encoded by a linear combination of the three basis vectors, hence providing the opportunity to process images in a geometric way. Within the feature-rich hypercomplex setting, novel image processing workflows can be realized for natural and biomedical images enabling alternative visual representations, offering effective solutions to current problems in computer vision and digital pathology, and generally expanding the scope and impact of hypercomplex image processing across a wide range of applications.}
\end{tcolorbox}

Emerging computational methodologies for image analysis, grounded in feature-rich mathematical frameworks, are consistently advancing the boundaries of technological innovation \cite{Valous2024_1}. Hypercomplex image processing is a fascinating field that extends conventional methods by the use of hypercomplex numbers in a unified framework for algebra and geometry \cite{Valous2024_1}. Processing images in the hypercomplex domain allows for more complex and intuitive representations with algebraic properties that can lead to new insights and optimizations \cite{Valous2024_1, Valous2024_2}.

Quaternions form one of the four existing normed division algebras (real, complex, quaternions, and octonions), regarded as a special case of hypercomplex number systems \cite{LeBihan2004}. As a generalization, real Clifford algebras \cite{Josipovi2019} can be seen as hypercomplex number systems that incorporate the geometric notion of direction and orientation \cite{deCastro2020}; for example, the quaternion algebra is the even Clifford algebra $Cl3^+$ \cite{Baez2001}. Many applications of hypercomplex numbers are spawned not only in the fields of physics and mathematics but also in a variety of branches of modern science and technology where the utilization of quaternions (and largely Clifford algebras) is constantly expanding \cite{Kharinov2018, Valous2020, Hitzer2023}. These applications include computer graphics, image/signal processing, quantum computing, electromagnetism, satellite navigation, machine/deep learning, electrical engineering, robotics, etc. \cite{Hitzer2013, Hitzer2023}. Quaternions have been utilized extensively to rotate vectors in three dimensions \cite{Goldman2011}. Essentially, real and complex number systems are used to provide arithmetic operations of one-dimensional and two-dimensional data, while quaternions can handle algebraic operations of ternary numbers, expressing color data directly \cite{SooChangPei1997} (refer to the \textit{Quaternion matrix representation} subsection in \textit{Materials and Methods}). This provides the opportunity to process images in a geometric way, and hence the quaternionic representation of color allows image processing to be performed coherently \cite{HitzerSangwine2013}.

\begin{figure}[b]
\centering
\includegraphics[width=1.0\linewidth]{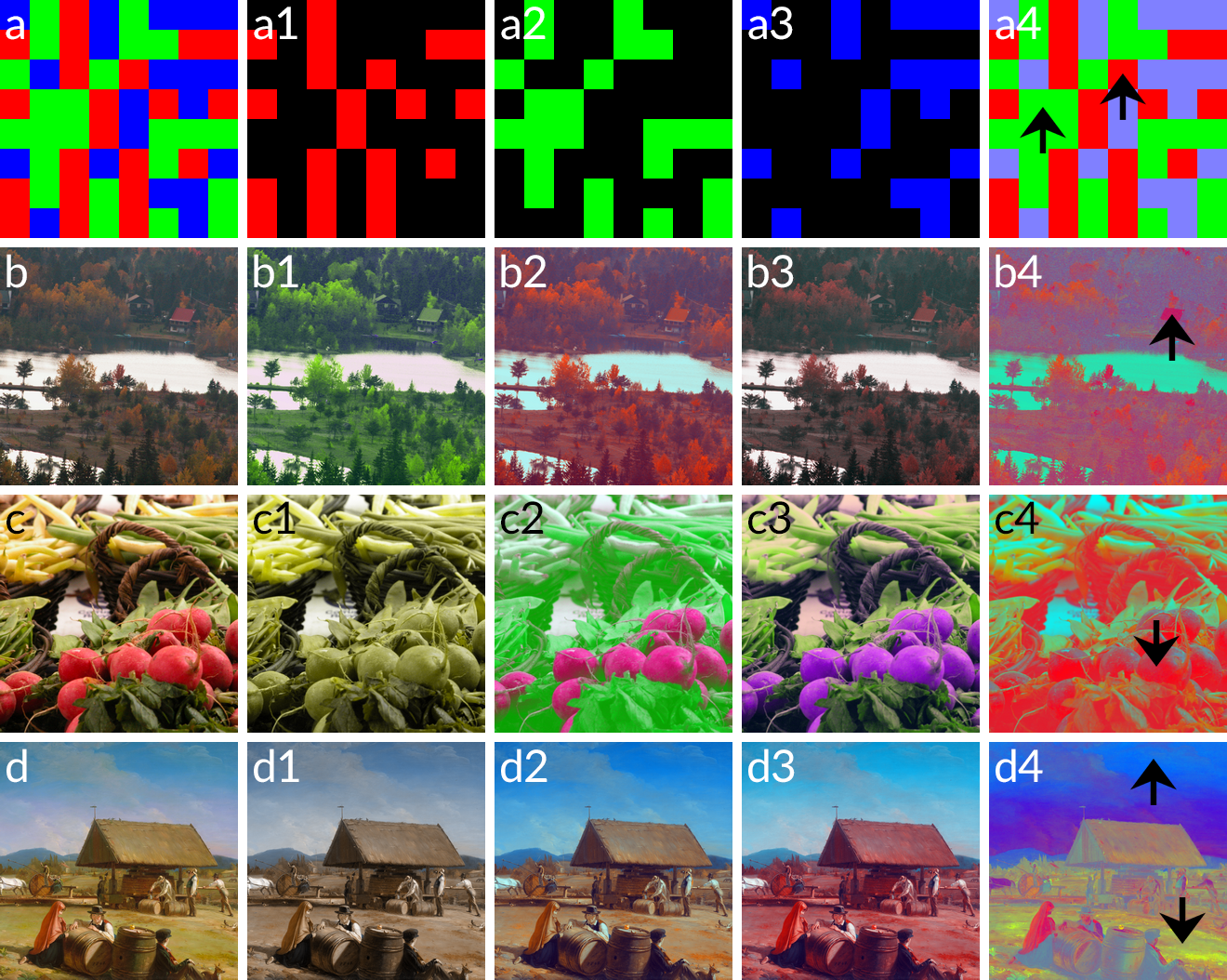}
\vspace{-12pt}
\caption{\footnotesize{Re-colorization in synthetic and natural images (24-bit). Synthetic image (a) containing red, green, and blue color pixel blocks (block size: $32^2$ pixels) with sample renditions: (a1) $q_-$, $\boldsymbol{f} = \boldsymbol{\mu_{1}}$, (a2) $q_-$, $\boldsymbol{f} = \boldsymbol{\mu_{2}}$, and (a3) $q_-$, $\boldsymbol{f} = \boldsymbol{\mu_{3}}$. Natural images (b-c) from the McGill color image database \cite{Olmos2004} with sample renditions: (b1) $q_+$, $\boldsymbol{f} = \boldsymbol{\mu_{2}}$, $\boldsymbol{g} = \boldsymbol{\mu_{11}}$, (b2) $q_-$, $\boldsymbol{f} = \boldsymbol{\mu_{3}}$, $\boldsymbol{g} = \boldsymbol{\mu_{8}}$, (b3) $q_-$, $\boldsymbol{f} = \boldsymbol{\mu_{2}}$, $\boldsymbol{g} = \boldsymbol{\mu_{3}}$, (c1) $q_-$, $\boldsymbol{f} = \boldsymbol{\mu_{1}}$, $\boldsymbol{g} = \boldsymbol{\mu_{2}}$, (c2) $q_-$, $\boldsymbol{f} = \boldsymbol{\mu_{1}}$, $\boldsymbol{g} = \boldsymbol{\mu_{10}}$, and (c3) $q_+$, $\boldsymbol{f} = \boldsymbol{\mu_{3}}$, $\boldsymbol{g} = \boldsymbol{\mu_{11}}$. Painting (d) with sample renditions: (d1) $q_-$, $\boldsymbol{f} = \boldsymbol{\mu_{2}}$, $\boldsymbol{g} = \boldsymbol{\mu_{5}}$, (d2) $q_-$, $\boldsymbol{f} = \boldsymbol{\mu_{5}}$, $\boldsymbol{g} = \boldsymbol{\mu_{8}}$, and (d3) $q_+$, $\boldsymbol{f} = \boldsymbol{\mu_{8}}$, $\boldsymbol{g} = \boldsymbol{\mu_{13}}$. Corresponding re-colorization images with user-defined values ($q_+$, sampling shown by arrows and expressed as pure unit quaternions) are shown in (a4-d4). Arrows correspond to pixel coordinates in (a-d); a single arrow corresponds to $\boldsymbol{f}$ while two arrows to $\boldsymbol{f}$ and $\boldsymbol{g}$. Pure unit quaternions $\boldsymbol{\mu_{1}}$-$\boldsymbol{\mu_{13}}$ are shown in Fig. \ref{fig:m1m13}. Images have $256^2$ pixels.}}
\label{fig:img_recol}
\end{figure}

Natural and biomedical image composition is an amalgam of characteristics comprising the information encoded in the pixels and the emergence of local and global patterns. Based on a hypercomplex framework called the two-dimensional (2D) orthogonal planes split (OPS), which allows the splitting of a quaternion (representing a pixel) into pairs of orthogonal two-dimensional planes \cite{HitzerSangwine2013, Hitzer2015} (refer to the \textit{Quaternion algebra} and \textit{2D orthogonal planes split} subsections in \textit{Materials and Methods}), the following computational workflows and outcomes are presented: i) natural and biomedical image re-colorization, ii) natural image de-colorization, iii) natural and biomedical image contrast enhancement, iv) computational re-staining and stain separation in histological images, and v) performance enhancements in machine learning pipelines for biomedical images. It is important to note that for the workflows presented in this study, only basic arithmetic and matrix operations were utilized to accompany the proposed approaches. 

\begin{figure}[t]
\centering
\includegraphics[width=1.0\linewidth]{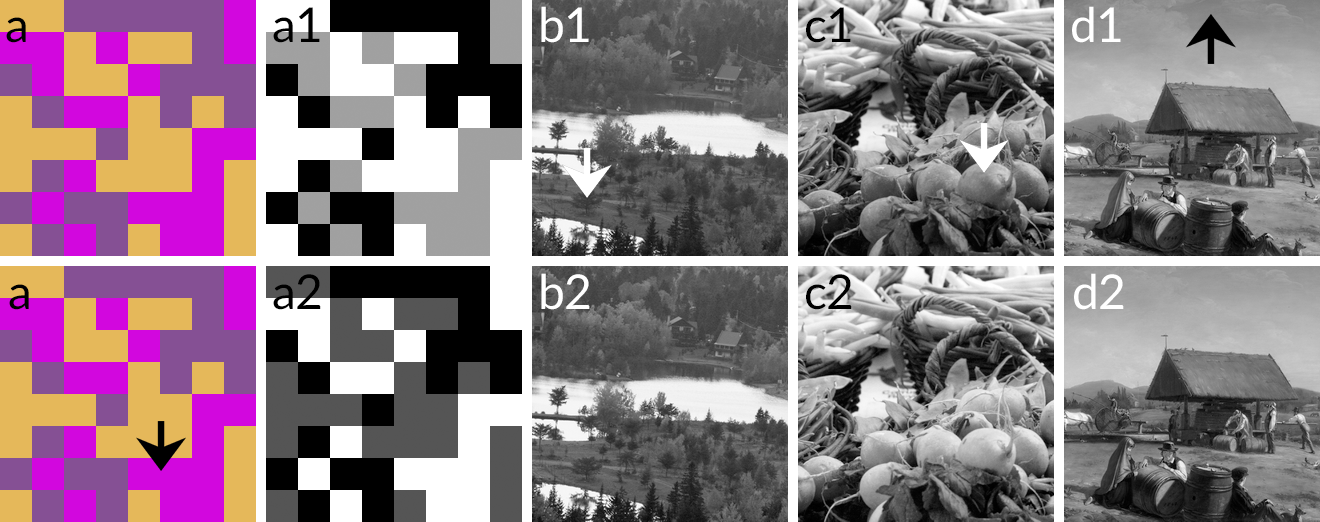}
\vspace{-12pt}
\caption{\footnotesize{De-colorization in synthetic and natural images (24-bit): (a) synthetic image comprising randomly colored pixel blocks (block size: $32^2$ pixels) and its corresponding 8-bit renditions; a1: $q_-$, $\boldsymbol{f} = \boldsymbol{\mu_{7}}$ and a2: $q_-$, $\boldsymbol{f}$, sampling shown by arrow in (a) and expressed as a pure unit quaternion. Likewise, images (b1-d1): $q_-$, $\boldsymbol{f} = \boldsymbol{\mu_{7}}$ and (b2-d2): $q_-$, $\boldsymbol{f}$ (sampling shown by arrows and expressed as pure unit quaternions) are the 8-bit renditions of Fig. \ref{fig:img_recol}(b-d), respectively. The arrows in (b1-d1) correspond to pixel coordinates in the color images of Fig. \ref{fig:img_recol}(b-d) for obtaining (b2-d2). Pure unit quaternions $\boldsymbol{\mu_{1}}$-$\boldsymbol{\mu_{13}}$ are shown in Fig. \ref{fig:m1m13}. Images have $256^2$ pixels.}}
\label{fig:img_decol1}
\end{figure}

\begin{figure}[t]
\centering
\includegraphics[width=1.0\linewidth]{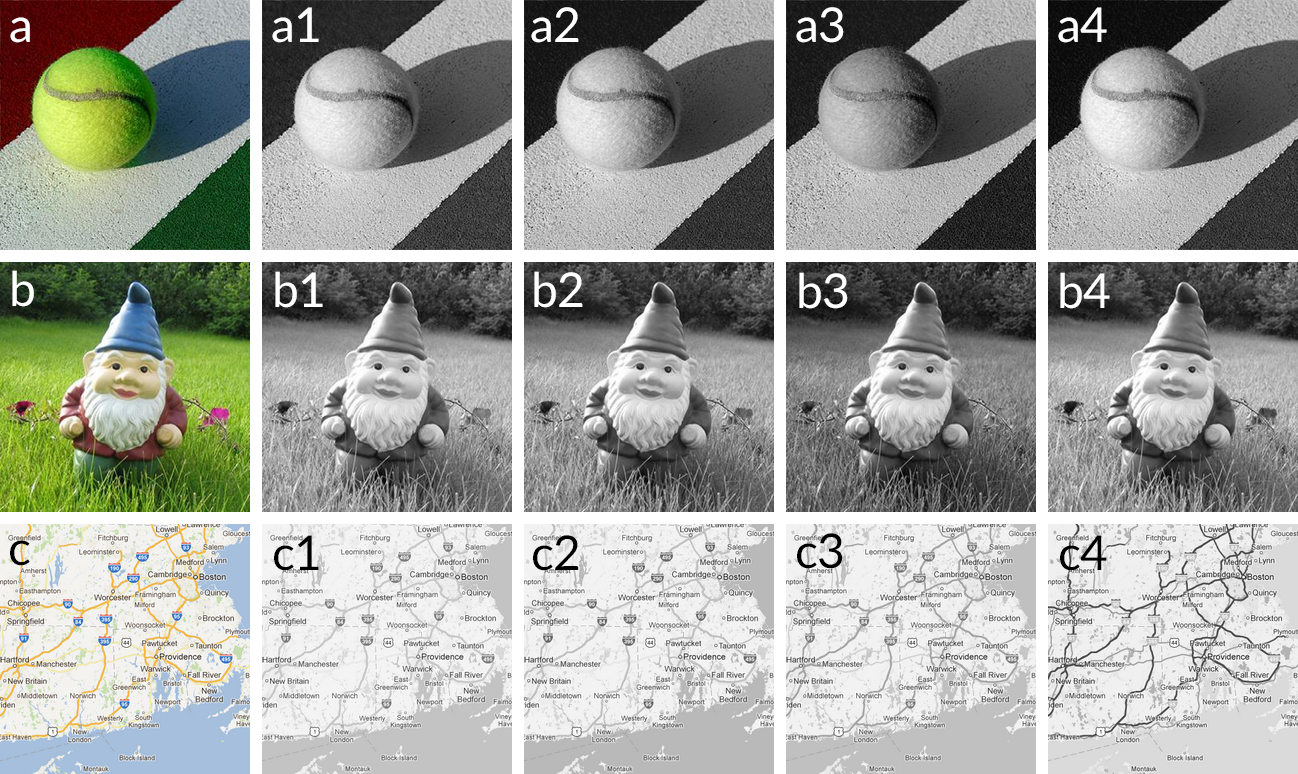}
\vspace{-12pt}
\caption{\footnotesize{De-colorization in natural images (24-bit). The original images shown (a-c) are selected examples from the COLOR250 dataset \cite{Lu2014}. The resulting images (8-bit) in a(1-4), b(1-4), and c(1-4) derive from methods L \cite{Hunt2011}, O \cite{ITU-R_BT.2100}, and proposed methods P1 and P2b (Table \ref{tab:decol_comp}), respectively. Displayed images are cropped and scaled versions ($250^2$ pixels) for presentation purposes.}}
\label{fig:img_decol2}
\end{figure}

\begin{figure}[t]
\centering
\includegraphics[width=1.0\linewidth]{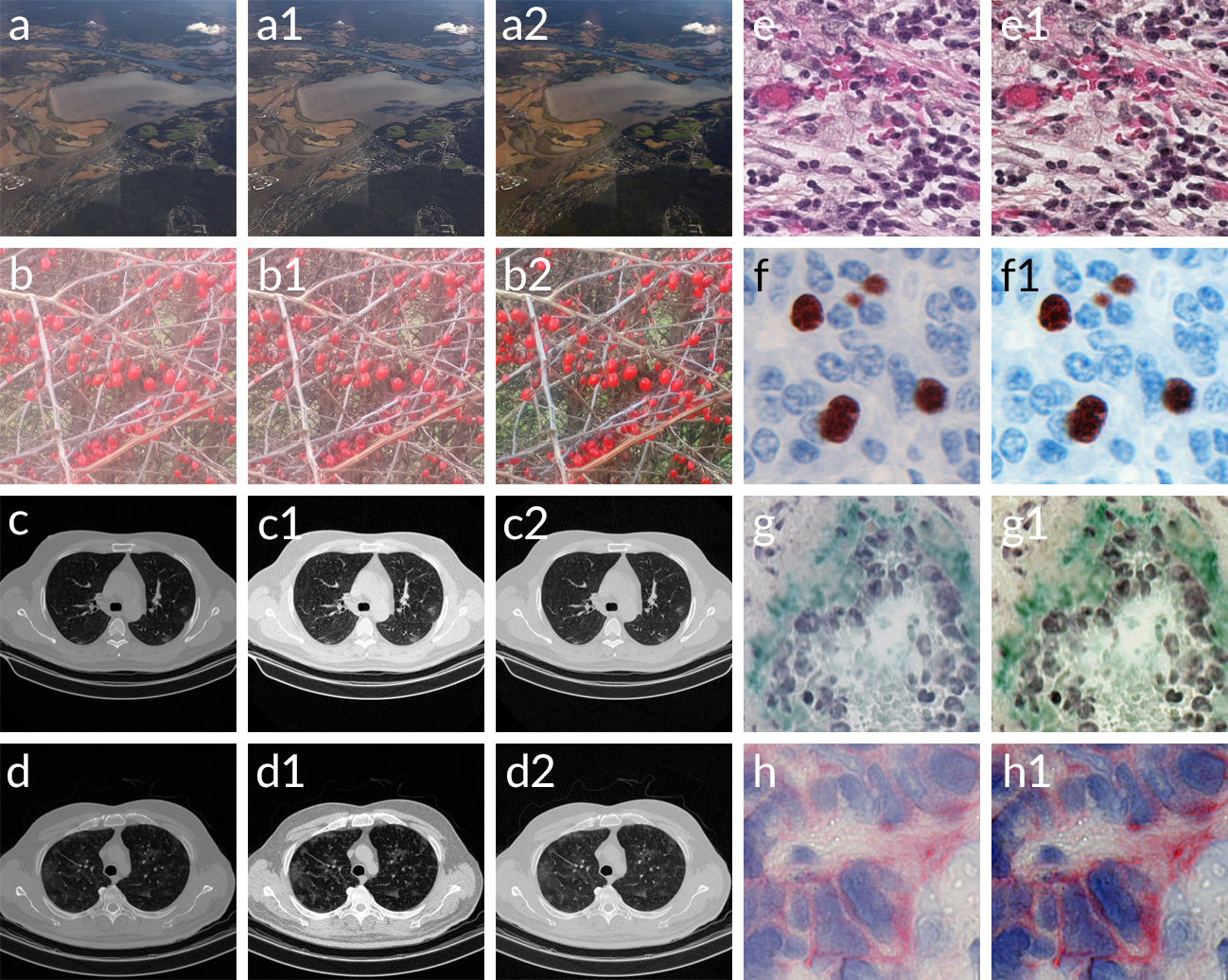}
\vspace{-12pt}
\caption{\footnotesize{Contrast enhancement in natural and medical (CT) images (24-bit, 8-bit). The original images (a-d) are examples from the CEED2016 \cite{Qureshi2016, Qureshi2017} and COVID-LDCT \cite{Afshar2021} datasets. The images in (a1-d1) derive from methods A-C \cite{Beghdadi1989, Zuiderveld1994, Mukherjee2008} (Table \ref{tab:contr_comp}). The images in (a2-d2) derive from the proposed method P (Table \ref{tab:contr_comp}). Histological close-up image sections (e-h) show an H\&E stain, and single immunostains that exhibit approx. brown, green, and red colors, respectively. For (f-h), the counterstain exhibits approx. blue color. The images (e1-h1) derive from the proposed method P (Table \ref{tab:contr_comp}). Natural and medical (CT) images have $512^2$ pixels, and histology images have $250^2$ pixels.}}
\label{fig:img_contr}
\end{figure}

Pertinent to previous/related work, the 2D orthogonal planes split (or $\pm$ split) was studied extensively, from a mathematical perspective, in Refs. \cite{HitzerSangwine2013, Hitzer2015}. The concept was also mentioned and/or studied in Refs. \cite{Ell2000, Sangwine2002-pk, Moxey2002, Hitzer2007, Ell2007b, Hitzer2009, Sangwine2013, Lan2016, Hitzer2016, Bahri2019}. More specifically, Ref. \cite{Sangwine2002-pk} utilized the concept for designing linear color-dependent filters based on the decomposition of an image into components parallel and perpendicular to a chosen direction in color space. Although the aim was to perform color image filtering, the example and overall presentation provided a good starting point for further developing the idea. Moreover, Ref. \cite{Lan2016} proposed a quaternion decomposition-based discriminant analysis method for face recognition. The authors mentioned the concept while providing an example of an image with changed colors as well as its grayscale version; this visualization signified that further exploration was worthwhile. Both Refs. \cite{Sangwine2002-pk, Lan2016} provided an excellent base and motivation for probing the application potential for natural and biomedical images. To the best of our knowledge, no prior research work - based on the 2D orthogonal planes split - showcases the range of computer vision and biomedical applications explored in this study. Aspects of this work related to image re-colorization and histological stain separation are covered under U.S. Patent No. 11,501,444 \cite{valous2022patent}.

The paper is organized as follows. Details about tissue preparation, software packages/libraries/code, datasets, performance metrics, and mathematical/computational workflow specifics along with definitions and introductions to quaternion algebra and the 2D orthogonal planes split concept are given in the \textit{Materials and Methods} section. This section provides valuable insights and background information that should be consulted to develop a comprehensive understanding of the results. In the \textit{Results} section, the outcomes of each workflow are visualized and (when possible) scrutinized with literature comparisons (using publicly available code/datasets). For fair comparisons, non-data-driven methods (which do not rely on large datasets for model training) were sought from the literature. In the \textit{Discussion} section, the workflows for natural and biomedical images are examined separately, accompanied by the study’s limitations, potential avenues for further development, and concluding remarks.

\begin{figure}[t]
\centering
\includegraphics[width=1.0\linewidth]{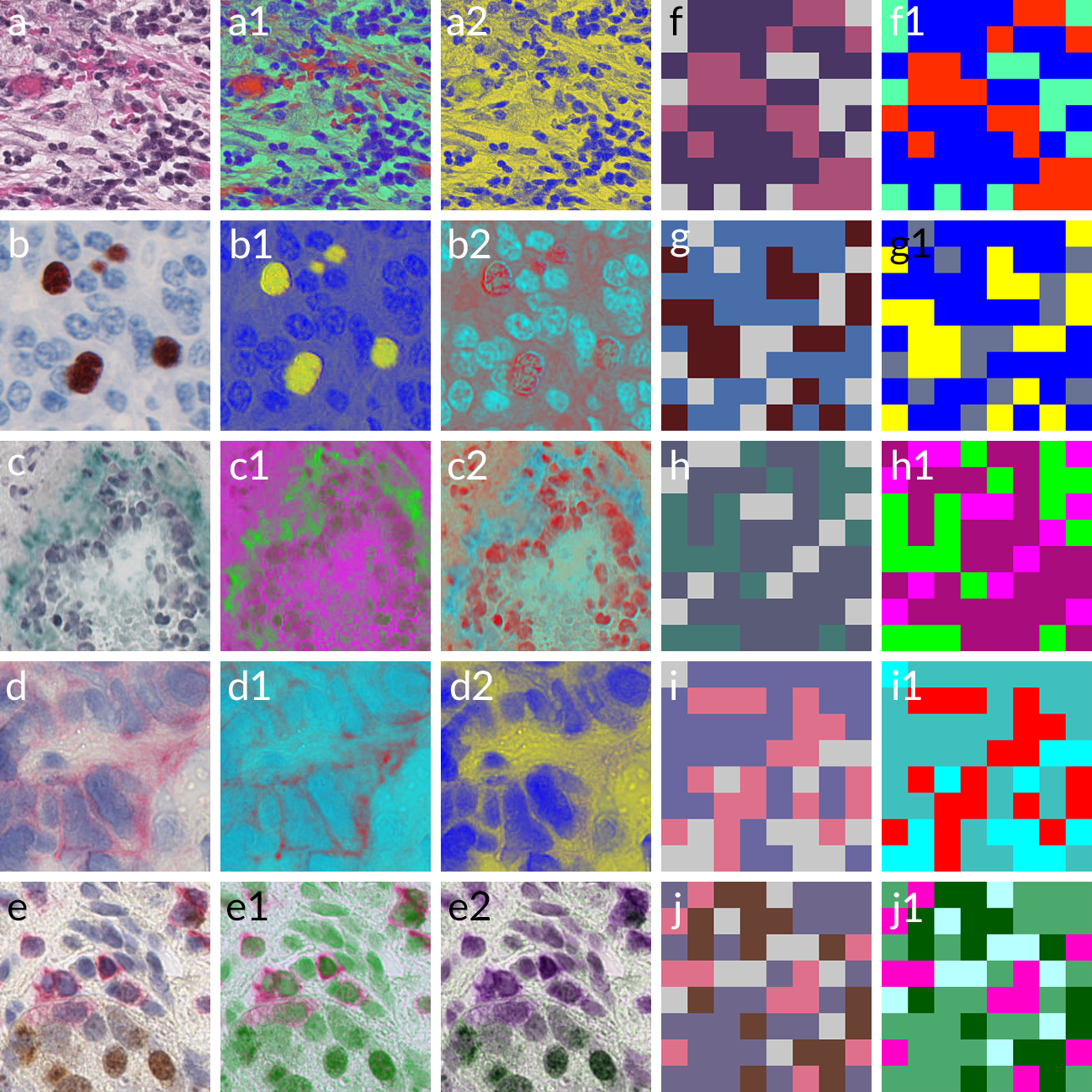}
\vspace{-12pt}
\caption{\footnotesize{Re-colorization in histological images and their synthetic models (24-bit). Histological close-up image sections (a-e) show an H\&E stain, single immunostains that exhibit approx. brown, green, and red colors, and a double immunostain exhibiting approx. brown and red colors, respectively. For (b-e), the counterstain exhibits approx. blue color. The sample renditions are: (a1) $q_+$, $\boldsymbol{f} = \boldsymbol{\mu_{5}}$, (a2) $q_+$, $\boldsymbol{f} = \boldsymbol{\mu_{9}}$, $\boldsymbol{g} = \boldsymbol{\mu_{13}}$, (b1) $q_+$, $\boldsymbol{f} = \boldsymbol{\mu_{3}}$, $\boldsymbol{g} = \boldsymbol{\mu_{13}}$, (b2) $q_+$, $\boldsymbol{f} = \boldsymbol{\mu_{1}}$, $\boldsymbol{g} = \boldsymbol{\mu_{11}}$, (c1) $q_-$, $\boldsymbol{f} = \boldsymbol{\mu_{10}}$, $\boldsymbol{g} = \boldsymbol{\mu_{12}}$, (c2) $q_-$, $\boldsymbol{f} = \boldsymbol{\mu_{8}}$, $\boldsymbol{g} = \boldsymbol{\mu_{11}}$, (d1) $q_-$, $\boldsymbol{f} = \boldsymbol{\mu_{11}}$, (d2) $q_+$, $\boldsymbol{f} = \boldsymbol{\mu_{4}}$, $\boldsymbol{g} = \boldsymbol{\mu_{9}}$, (e1) $q_-$, $\boldsymbol{f} = \boldsymbol{\mu_{1}}$, $\boldsymbol{g} = \boldsymbol{\mu_{10}}$, and (e2) $q_-$, $\boldsymbol{f} = \boldsymbol{\mu_{8}}$, $\boldsymbol{g} = \boldsymbol{\mu_{13}}$. Synthetic images (f-j) containing approx. color matching pixel blocks (block size: $32^2$ pixels) that correspond to previous stains, e.g. (f) simulates (a), (g) simulates (b), etc. The sample renditions are: (f1) $q_+$, $\boldsymbol{f} = \boldsymbol{\mu_{5}}$, (g1) $q_+$, $\boldsymbol{f} = \boldsymbol{\mu_{3}}$, $\boldsymbol{g} = \boldsymbol{\mu_{13}}$, (h1) $q_-$, $\boldsymbol{f} = \boldsymbol{\mu_{10}}$, $\boldsymbol{g} = \boldsymbol{\mu_{12}}$, (i1) $q_-$, $\boldsymbol{f} = \boldsymbol{\mu_{11}}$, and (j1) $q_-$, $\boldsymbol{f} = \boldsymbol{\mu_{1}}$, $\boldsymbol{g} = \boldsymbol{\mu_{10}}$. Synthetic images contain gray pixel blocks to simulate the microscopy slide backdrop. Pure unit quaternions $\boldsymbol{\mu_{1}}$-$\boldsymbol{\mu_{13}}$ are shown in Fig. \ref{fig:m1m13}. Images have $250^2$ pixels.}}
\label{fig:img_recol_bio}
\end{figure}

\section*{\large\uppercase{Results}}

\subsection*{\normalsize{Natural image re-colorization}}
Fig. \ref{fig:img_recol} illustrates image re-colorization using the proposed approaches on two images from the McGill color image database \cite{Olmos2004}, a painting from the open access artworks collection of the Metropolitan Museum of Art (New York), and a synthetic color image. Transformations may have geometric meaning (e.g. Fig. \ref{fig:m1m13}, $\boldsymbol{\mu_1}$, $\boldsymbol{\mu_2}$, and $\boldsymbol{\mu_3}$ refer to the red, green, and blue color axes, respectively), but can also be defined by the user (e.g. pixel sampling).

\subsection*{\normalsize{Natural image de-colorization}}
Fig. \ref{fig:img_decol1} exhibits image de-colorization using the proposed approaches on the images of Fig. \ref{fig:img_recol} and a synthetic color image. Table \ref{tab:decol_comp} shows comparisons among several de-colorization methods (A-O) with the proposed approaches (P1, P2a, and P2b) using the COLOR250 dataset \cite{Lu2014}. Transformations can have geometric meaning (e.g. Fig. \ref{fig:m1m13}, $\boldsymbol{\mu_7}$ is the direction corresponding to the luminance or gray-line axis \cite{Moxey2003, Ell2007a}), can be defined by the user or computed by automated methods. Methods A-K refer to different techniques from the literature \cite{Lu2014, CewuLu2012, Grundland2007, Smith2008, Gooch2005, Nafchi2017, QiegenLiu2015, Liu2017a, Xiong2017, Liu2016, Liu2017b}. Method L refers to the CIE Y color channel (CIE 1931 XYZ color space) \cite{Hunt2011}, and methods M-O to different recommendations for computing grayscale values \cite{InternationalTelecommunicationUnion2011, ITU-R_BT709, ITU-R_BT.2100}. Fig. \ref{fig:img_decol2} illustrates image de-colorization in selected examples from the COLOR250 dataset \cite{Lu2014}. Along with the original color images, renditions from literature methods and proposed approaches are shown (Table \ref{tab:decol_comp}).

\begin{table}[t]
\centering
\caption{\footnotesize{Comparisons of several de-colorization methods with the proposed approaches: methods P1, P2a, and P2b. Experiments ran on the COLOR250 dataset \cite{Lu2014} using the color-to-gray structural similarity metric (C2G-SSIM) (higher score is better) \cite{Ma2015}. The default or recommended values are utilized from the corresponding references (methods A-O). The numerical values displayed are the computed arithmetic means $\pm$ standard deviations, and medians. Values highlighted in boldface showcase the four best-performing methods; D, L, O, and proposed method P2b.}}
\vspace*{-0.75mm}
\resizebox{0.65\columnwidth}{!}{
\begin{tabular}{lclc}
\toprule
\multirow{1}*{\textbf{De-colorization}} & \multicolumn{2}{c}{\textbf{C2G-SSIM $\uparrow$}} \\
\textbf{methods} & \textbf{Mean $\pm$st.dev.} & \textbf{Median} \\
\midrule
A \cite{Lu2014} & 0.8907 $\pm$0.075 & 0.9047 \\
B \cite{CewuLu2012} & 0.8926 $\pm$0.073 & 0.9108 \\
C \cite{Grundland2007} & 0.8749 $\pm$0.083 & 0.8925 \\
D \cite{Smith2008} & \textbf{0.9069} $\pm$0.058 & \textbf{0.9211} \\
E \cite{Gooch2005} & 0.8935 $\pm$0.066 & 0.9045 \\
F \cite{Nafchi2017} & 0.8900 $\pm$0.067 & 0.8989 \\
G \cite{QiegenLiu2015} & 0.8825 $\pm$0.086 & 0.9010 \\
H \cite{Liu2017a} & 0.8903 $\pm$0.077 & 0.9079 \\
I \cite{Xiong2017} & 0.8961 $\pm$0.067 & 0.9101 \\
J \cite{Liu2016} & 0.8852 $\pm$0.068 & 0.8966 \\
K \cite{Liu2017b} & 0.8656 $\pm$0.094 & 0.8878 \\
L \cite{Hunt2011} & \textbf{0.9092} $\pm$0.058 & \textbf{0.9218} \\
M \cite{InternationalTelecommunicationUnion2011} & 0.9062 $\pm$0.060 & 0.9167 \\
N \cite{ITU-R_BT709} & 0.9062 $\pm$0.060 & 0.9170 \\
O \cite{ITU-R_BT.2100} & \textbf{0.9075} $\pm$0.060 & \textbf{0.9180} \\
P1 & 0.8881 $\pm$0.068 & 0.8941 \\
P2a & 0.9030 $\pm$0.062 & 0.9168 \\
P2b & \textbf{0.9082} $\pm$0.057 & \textbf{0.9219} \\
\bottomrule
\end{tabular}}
\label{tab:decol_comp}
\end{table}

\begin{table}[t]
\centering
\caption{\footnotesize{Comparisons of several contrast enhancement methods with the proposed approach P. Experiments ran on the CEED2016 dataset \cite{Qureshi2016, Qureshi2017} (d-ce1) and on the COVID-LDCT dataset \cite{Afshar2021} (d-ce2) using the visual information fidelity (VIF) metric (higher score is better) \cite{Sheikh2006}, the reduced-reference image quality metric for contrast change (RIQMC) (lower score is better) \cite{Gu2016}, the lightness order error (LOE) metric (lower score is better) \cite{Wang2013}, and the over-contrast measure (OCM) (lower score is better) \cite{Lee2019}. The default or recommended values are utilized from the corresponding references (methods A-G). The numerical values displayed are the computed medians. Values highlighted in boldface showcase the two best-performing methods; instance-wise A, B, C, or G, and proposed method P.}}
\vspace*{-0.75mm}
\resizebox{\columnwidth}{!}{
\begin{tabular}{lccccc}
\toprule
\textbf{Contrast enhancement} & \textbf{VIF $\uparrow$} & \textbf{RIQMC $\downarrow$} & \textbf{LOE $\downarrow$} & \textbf{OCM $\downarrow$} \\
\midrule
A (d-ce1) \cite{Beghdadi1989} & 0.7101 & \textbf{1.2205} & 355.8354 & 0.1384 \\
B (d-ce1) \cite{Zuiderveld1994} & \textbf{1.0301} & 1.9036 & \textbf{130.1682} & 0.0512 \\
C (d-ce1) \cite{Mukherjee2008} & 0.7088 & 1.6448 & 525.6769 & \textbf{0.0253} \\
D (d-ce1) \cite{Hummel1977} & 0.7373 & 6.8883 & 187.8670 & 0.1859 \\
E (d-ce1) \cite{Mukhopadhyay2000} & 0.6935 & 4.6629 & 413.6403 & 0.3431 \\
F (d-ce1) \cite{Chen2010} & 0.9446 & 4.0721 & 299.6194 & 0.0809 \\
G (d-ce1) \cite{KaimingHe2011} & 0.9088 & \textbf{-1.9562} & \textbf{85.4148} & 0.0312 \\
P (d-ce1) & \textbf{0.9572} & \textbf{-0.2779} & \textbf{60.5365} & \textbf{0.0017} \\
\hline
A (d-ce2) \cite{Beghdadi1989} & \textbf{1.2059} & 17.3208 & 2.4206 & 0.3969 \\
B (d-ce2) \cite{Zuiderveld1994} & 1.0400 & \textbf{14.9787} & 2.5928 & \textbf{0.3626} \\
C (d-ce2) \cite{Mukherjee2008} & 0.8487 & 18.9110 & \textbf{2.0581} & 0.7680 \\
D (d-ce2) \cite{Hummel1977} & 0.8077 & 22.7074 & 15.6578 & 1.4698 \\
E (d-ce2) \cite{Mukhopadhyay2000} & 1.1529 & 21.3537 & 4.5863 & 0.6662 \\
F (d-ce2) \cite{Chen2010} & 1.1730 & 18.6128 & 204.0700 & 0.6818 \\
G (d-ce2) \cite{KaimingHe2011} & 0.4702 & \textbf{-26.8082} & 107.9658 & \textbf{-0.1305} \\
P (d-ce2) & \textbf{1.1830} & \textbf{10.8442} & \textbf{0.2307} & \textbf{0.0971} \\
\bottomrule
\end{tabular}}
\label{tab:contr_comp}
\end{table}

\subsection*{\normalsize{Natural and biomedical image contrast enhancement}}
Table \ref{tab:contr_comp} shows comparisons among several contrast enhancement methods (A-G) with the proposed approach (P) using the CEED2016 \cite{Qureshi2016, Qureshi2017} and COVID-LDCT \cite{Afshar2021} datasets. Methods A-G refer to different methods from the literature \cite{Beghdadi1989, Zuiderveld1994, Mukherjee2008, Hummel1977, Mukhopadhyay2000, Chen2010, KaimingHe2011} corresponding to representative categories, respectively: adaptive edge-based contrast enhancement, contrast-limited adaptive histogram equalization, discrete cosine transform-based contrast enhancement, global histogram equalization, top hat transformation-based contrast enhancement, multiscale retinex, and haze removal using dark channel prior. Fig. \ref{fig:img_contr} exhibits image contrast enhancement in selected examples from both datasets. Along with the original images, renditions from literature methods and proposed approach are shown (Table \ref{tab:contr_comp}). Four histological close-up image sections are shown along with the corresponding contrast-enhanced versions using the proposed approach (P).

\subsection*{\normalsize{Histological image re-colorization}}
Fig. \ref{fig:img_recol_bio} illustrates image re-colorization using the proposed approaches on five histological close-up image sections and on synthetic color images. In this context, the transformations can have geometric meaning (e.g. Fig. \ref{fig:m1m13}, $\boldsymbol{\mu_1}$, $\boldsymbol{\mu_2}$, and $\boldsymbol{\mu_3}$ refer to the red, green, and blue color axes, respectively). The synthetic model images were generated to match approx. the color appearance of their real-world counterparts. The synthetic images in Fig. \ref{fig:img_recol_bio}(f-j) correspond to the histological images of Fig. \ref{fig:img_recol_bio}(a-e). Similarly, the applied transformations in Fig. \ref{fig:img_recol_bio}(f1-j1) correspond to the ones of Fig. \ref{fig:img_recol_bio}(a1-e1). The purpose of this approx. color matching is to demonstrate the lack of generated artifacts from the proposed approaches which can be readily perceived in synthetic model images.

\begin{figure}[t]
\centering
\includegraphics[width=1.0\linewidth]{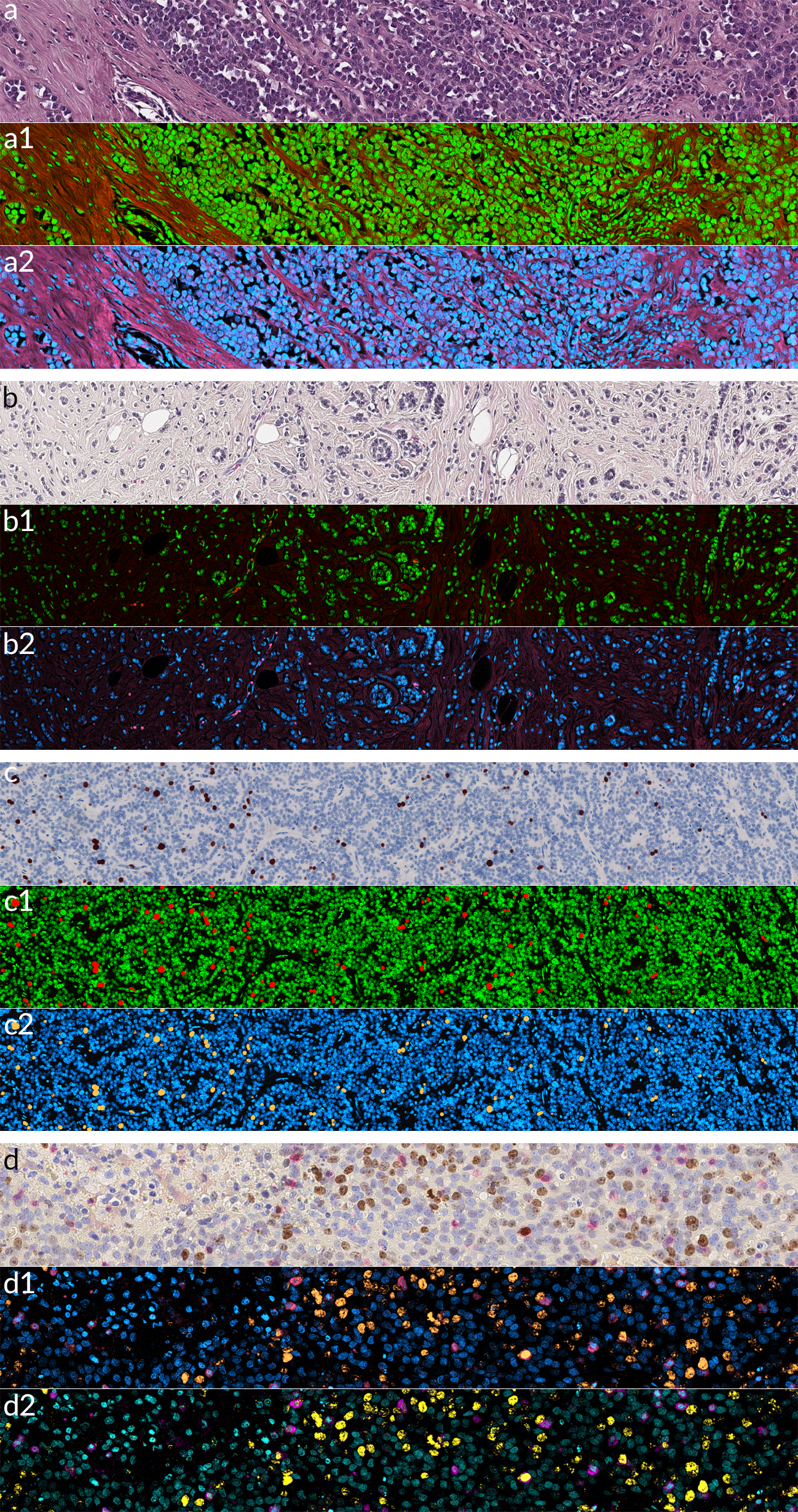}
\vspace{-12pt}
\caption{\footnotesize{Computational re-staining (targeted re-colorization) in histological images (24-bit). Histological close-up image sections (a-d) showing two H\&E stains, a single immunostain that exhibits approx. brown color, and a double immunostain exhibiting approx. brown and red colors, respectively. For (c-d), the counterstain exhibits approx. blue color. The sample renditions use: (a1)(b1)(c1) green (\#00FF00) and red color (\#FF0000), (a2)(b2) shades of blue (\#1E90FF) and pink color (\#DA467D), (c2) shades of blue (\#1E90FF) and orange color (\#D27D2D), (d1) shades of blue (\#1E90FF) pink (\#DA467D) and orange color (\#D27D2D), and (d2) cyan (\#00FFFF) magenta (\#FF00FF) and yellow color (\#FFFF00). The values in parentheses are the hexadecimal triplet representation of colors. Images have $1298 \times 200$ pixels.}}
\label{fig:img_restain_bio}
\end{figure}

\begin{figure}[t]
\centering
\includegraphics[width=1.0\linewidth]{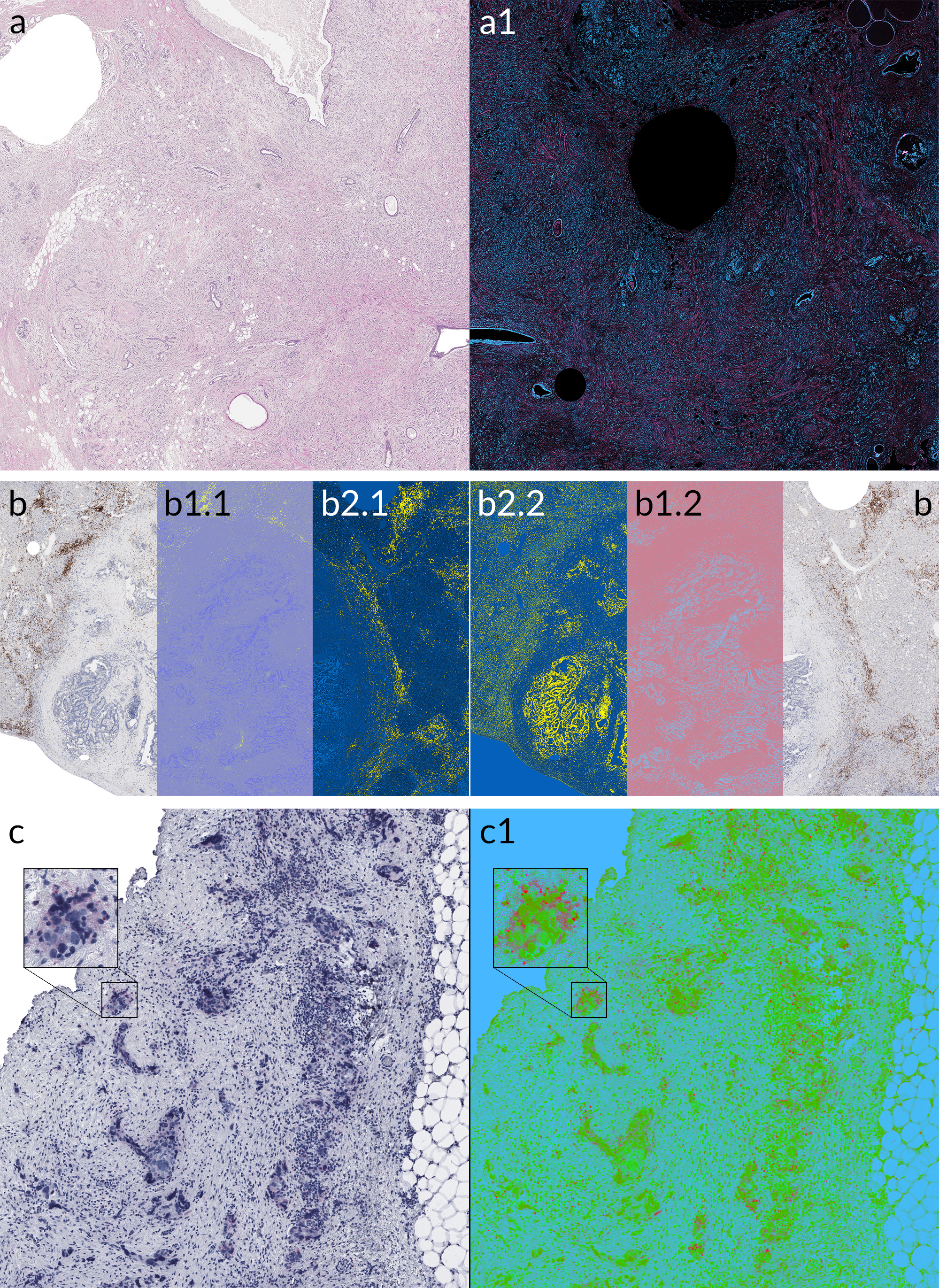}
\vspace{-12pt}
\caption{\footnotesize{Stain visualization of histological images (24-bit) using computational re-colorization and re-staining approaches. The H\&E image (a) shows invasive lobular carcinoma with the original (a) and computationally re-stained image (a1) using shades of blue (\#1E90FF) and pink color (\#DA467D), where the first half is the original H\&E image and the second half is the re-stained version. The single immunostain (b) shows human colorectal cancer liver metastasis where the first and last \nicefrac{1}{3} is the original image (b). The second \nicefrac{1}{3} (b1.1, b1.2) are the re-colorized versions, respectively: (b1.1) $q_+$, $\boldsymbol{f} = \boldsymbol{\mu_{1}}$, $\boldsymbol{g} = \boldsymbol{\mu_{11}}$ and (b2.2) $q_+$, $\boldsymbol{f} = \boldsymbol{\mu_{3}}$, $\boldsymbol{g} = \boldsymbol{\mu_{13}}$. The third \nicefrac{1}{3} (b2.1, b2.2) are the re-stained versions of (b1.1, b1.2) respectively, using shades of blue (\#1E90FF) and yellow color (\#FFDE21). The values in parentheses are the hexadecimal triplet representation of colors. The single immunostain (c) (in situ hybridization) shows human epithelial ovarian carcinoma with the original (c) and computationally re-colorized image (c1): $q_+$, $\boldsymbol{f} = \boldsymbol{\mu_{4}}$. The embedded images in (c) and (c1) show magnified areas. For this re-colorization process, a small exemplar image was selectively cropped to include shades of blue and red present in the original image (c). Images (a), (b), and (c) have $20000 \times 10024$, $12494 \times 8384$, and $6768 \times 6907$ pixels, respectively. Pure unit quaternions $\boldsymbol{\mu_{1}}$-$\boldsymbol{\mu_{13}}$ are shown in Fig. \ref{fig:m1m13}. All renditions are best appreciated when viewed up close.}}
\label{fig:img_large_scale_vis_bio}
\end{figure}

\subsection*{\normalsize{Computational re-staining}}
Fig. \ref{fig:img_restain_bio} illustrates computational re-staining using the proposed approaches on four histological close-up image sections. Computational re-staining is essentially targeted re-colorization on selected image colors that can simulate or alter the original colors to produce similar or alternative visual representations. The new renditions are obtained with a rather uniform dark background color.

\begin{figure}[t]
\centering
\includegraphics[width=1.0\linewidth]{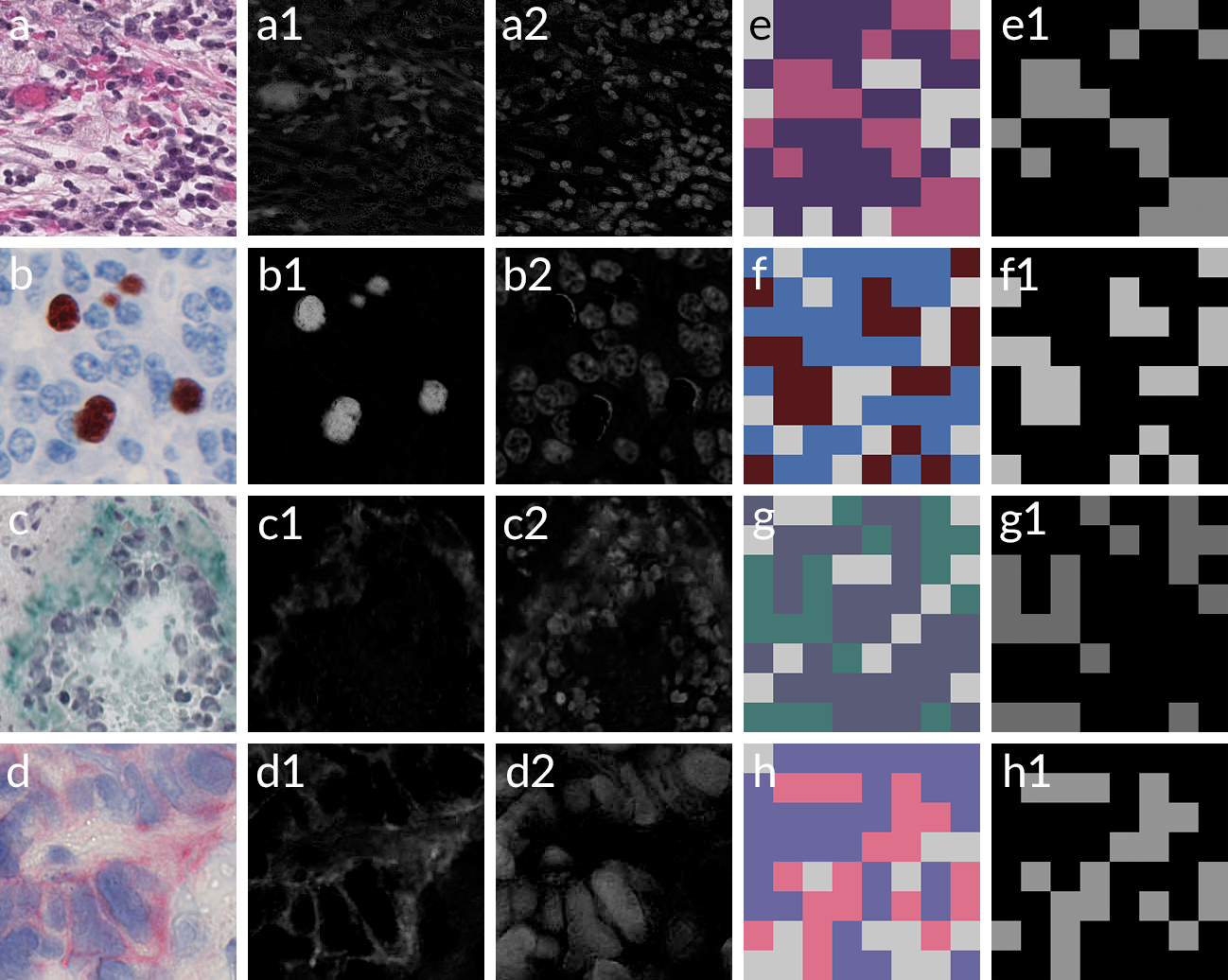}
\vspace{-12pt}
\caption{\footnotesize{Computational stain separation in histological images and their synthetic models (24-bit) based on transformations that have geometric meaning ($\boldsymbol{\mu_{7}}$). Histological close-up image sections (a-d) show an H\&E stain, and single immunostains that exhibit approx. brown, green, and red colors, respectively. For (b-d), the counterstain exhibits approx. blue color. The resulting images (8-bit) of the proposed approaches (a1-d1, a2-d2, and e1-h1) are provided as a direct output of the computation, without any modification or further post-processing. Images (e1-h1) correspond to images (a1-d1). Synthetic images contain gray pixel blocks to simulate the microscopy slide backdrop. Images have $250^2$ pixels.}}
\label{fig:img_unsupstainsep_bio}
\end{figure}

\begin{figure}[t]
\centering
\includegraphics[width=1.0\linewidth]{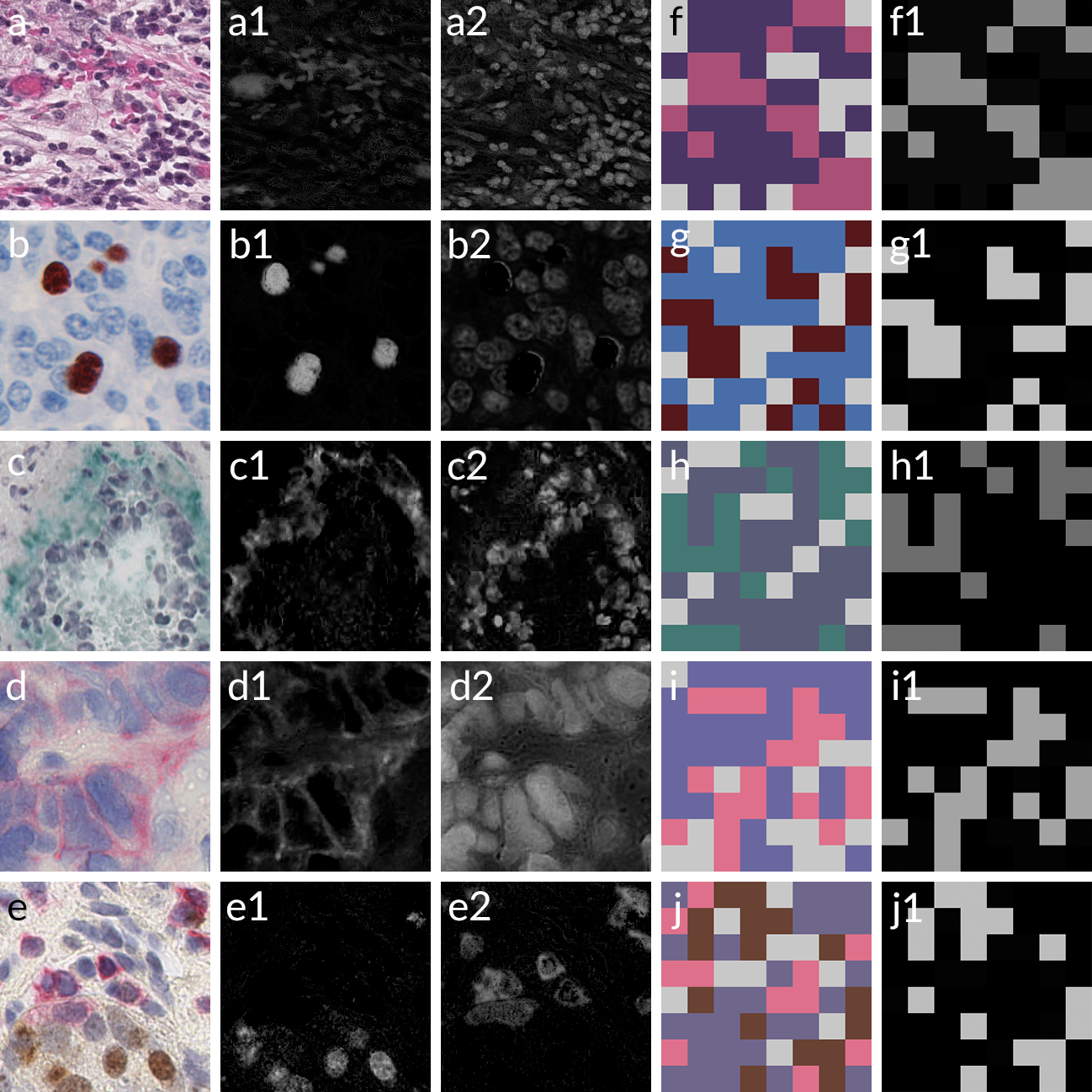}
\vspace{-12pt}
\caption{\footnotesize{Computational stain separation in histological images and their synthetic models (24-bit) with user-defined values. Histological close-up image sections (a-e) show an H\&E stain, single immunostains that exhibit approx. brown, green, and red colors, and a double immunostain exhibiting approx. brown and red colors, respectively. For (b-e), the counterstain exhibits approx. blue color. The resulting images (8-bit) of the proposed approaches (a1-e1, a2-e2, and f1-j1) are provided as a direct output of the computation, without any modification or further post-processing. Images (f1-j1) correspond to images (a1-e1). Synthetic images contain gray pixel blocks to simulate the microscopy slide backdrop. Images have $250^2$ pixels.}}
\label{fig:img_supstainsep_bio}
\end{figure}

\subsection*{\normalsize{Stain visualization}}
Fig. \ref{fig:img_large_scale_vis_bio} exhibits examples of stain visualization using computational re-colorization and re-staining approaches on three histological images. The H\&E image in Fig. \ref{fig:img_large_scale_vis_bio}(a) shows invasive lobular carcinoma which is frequently characterized by a dissolute growth pattern that makes it difficult to identify both radiologically and histologically. The re-stained rendition Fig. \ref{fig:img_large_scale_vis_bio}(a1) improves the viewing and identification of histological structures and tissue compartments, e.g. dissolute tumor cells. The single immunostain in Fig. \ref{fig:img_large_scale_vis_bio}(b) shows a human liver metastasis core, its invasive margin into the liver, and the adjacent liver. T lymphocytes are detected using immunohistochemistry for CD3 (brown color). In these re-colorized and re-stained renditions, CD3 staining is visualized as a standalone feature in the invasive margin and adjacent liver Fig. \ref{fig:img_large_scale_vis_bio}(b1.1, b2.1). The staining can also be visually suppressed showing only surrounding counterstained histological structures Fig. \ref{fig:img_large_scale_vis_bio}(b1.2, b2.2). The single immunostain in Fig. \ref{fig:img_large_scale_vis_bio}(c) shows an omentum metastasis of a human high-grade serous ovarian cancer (little differentiated). The tumor is located on the left part of the image, and its interface with the visceral adipose tissue (white structures) of the omentum is shown on the right part of the image. The RNA molecules encoding the NLRP3 protein are detected using in situ hybridization and appear as subcellular dots (red color) in the cytoplasm. In the re-colorized rendition Fig. \ref{fig:img_large_scale_vis_bio}(c1), the subcellular staining (red color) can be readily perceived, thus expediting the task of assessing the spatial localization of the RNA molecules.

\begin{figure}[t]
\centering
\includegraphics[width=1.0\linewidth]{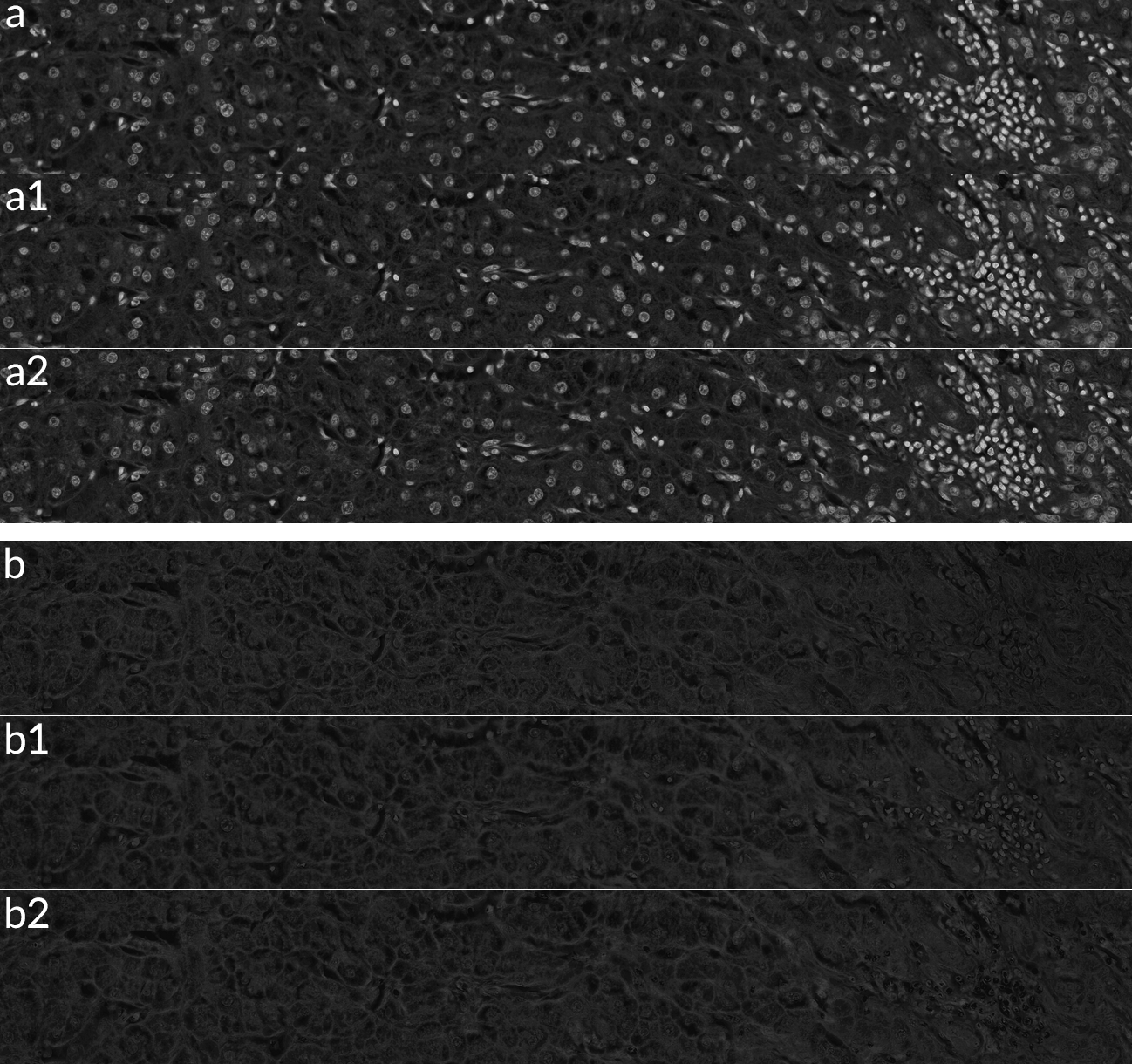}
\vspace{-12pt}
\caption{\footnotesize{Computational stain separation in an H\&E image (8-bit). Close-up image sections (a) and (b) were cropped from the lab ground truth 8-bit H-only and E-only images (d-he1), respectively. The images in (a1-a2) derive from method A \cite{Ruifrok2001} and proposed method P2a, respectively (Table \ref{tab:Hstain_sep_comp}). The images in (b1-b2) derive from method D \cite{Khan2014, stain_normalisation_toolbox_v2_2, Macenko2009} and proposed method P2a, respectively (Table \ref{tab:Estain_sep_comp}). Images have $1298 \times 200$ pixels. Stain-separated images are best appreciated when viewed up close.}}
\label{fig:img_stainsep_bio}
\end{figure}

\begin{table}[t]
\centering
\caption{\footnotesize{Comparisons of hematoxylin (H) stain separation methods with the proposed approaches: P1, P2a, and P2b. Experiments ran on an in-house H\&E staining dataset (d-he1, d-he2, and d-he3) using full-reference quality metrics such as: mean-squared error (MSE; lower score is better) \cite{matlab_toolbox_ref}, peak signal-to-noise ratio (PSNR; higher score is better) \cite{matlab_toolbox_ref}, structural similarity index (SSIM; value closer to 1 is better) \cite{matlab_toolbox_ref}, and multiscale structural similarity index (MS-SSIM; value closer to 1 is better) \cite{matlab_toolbox_ref}. The default or recommended values are utilized from the corresponding references (methods A-F). Values highlighted in boldface showcase the three best-performing methods; A and instance-wise proposed methods P1, P2a, or P2b.}}
\vspace*{-0.75mm}
\resizebox{\columnwidth}{!}{
\begin{tabular}{lccccc}
\toprule
\textbf{Stain separation for H} & \textbf{MSE $\downarrow$} & \textbf{PSNR $\uparrow$} & \textbf{SSIM $\uparrow$} & \textbf{MS-SSIM $\uparrow$} \\
\midrule
A (d-he1) \cite{Ruifrok2001} & \textbf{0.0006} & \textbf{32.2253} & \textbf{0.8666} & \textbf{0.9328} \\
B (d-he1) \cite{Ruifrok2001} & 0.0007 & 31.5233 & 0.8474 & 0.9177 \\
C (d-he1) \cite{Khan2014, stain_normalisation_toolbox_v2_2, Ruifrok2001} & 0.0009 & 30.4350 & 0.7972 & 0.8795 \\
D (d-he1) \cite{Khan2014, stain_normalisation_toolbox_v2_2, Macenko2009} & 0.0021 & 26.8602 & 0.7797 & 0.7761 \\
E (d-he1) \cite{Khan2014, stain_normalisation_toolbox_v2_2} & 0.0020 & 26.9967 & 0.7651 & 0.7764 \\
F (d-he1) \cite{Alsubaie2017} & 0.0028 & 25.5854 & 0.7586 & 0.7842 \\
P1 (d-he1) & \textbf{0.0006} & \textbf{32.3560} & \textbf{0.8783} & \textbf{0.9417} \\
P2a (d-he1) & \textbf{0.0006} & \textbf{31.8829} & \textbf{0.8580} & \textbf{0.9295} \\
P2b (d-he1) & 0.0007 & 31.8557 & 0.8568 & 0.9286 \\
\hline
A (d-he2) \cite{Ruifrok2001} & \textbf{0.0009} & \textbf{30.6051} & \textbf{0.8628} & \textbf{0.9394} \\
B (d-he2) \cite{Ruifrok2001} & 0.0009 & 30.3994 & 0.8557 & 0.9355 \\ 
C (d-he2) \cite{Khan2014, stain_normalisation_toolbox_v2_2, Ruifrok2001} & 0.0009 & 30.2897 & 0.8281 & 0.9245 \\
D (d-he2) \cite{Khan2014, stain_normalisation_toolbox_v2_2, Macenko2009} & 0.0039 & 24.1145 & 0.6434 & 0.7460 \\
E (d-he2) \cite{Khan2014, stain_normalisation_toolbox_v2_2} & 0.0051 & 22.8979 & 0.6150 & 0.6835 \\
F (d-he2) \cite{Alsubaie2017} & 0.0284 & 15.4652 & 0.4018 & 0.2880 \\
P1 (d-he2) & \textbf{0.0009} & \textbf{30.5770} & \textbf{0.8661} & \textbf{0.9418} \\
P2a (d-he2) & \textbf{0.0009} & \textbf{30.5150} & 0.8570 & \textbf{0.9381} \\
P2b (d-he2) & 0.0009 & 30.5115 & \textbf{0.8572} & 0.9380 \\
\hline
A (d-he3) \cite{Ruifrok2001} & 0.0020 & 27.0573 & 0.8166 & 0.8804 \\
B (d-he3) \cite{Ruifrok2001} & 0.0025 & 26.0150 & 0.7880 & 0.8574 \\
C (d-he3) \cite{Khan2014, stain_normalisation_toolbox_v2_2, Ruifrok2001} & 0.0024 & 26.2740 & 0.7878 & 0.8690 \\
D (d-he3) \cite{Khan2014, stain_normalisation_toolbox_v2_2, Macenko2009} & 0.0028 & 25.4885 & 0.7974 & 0.8693 \\
E (d-he3) \cite{Khan2014, stain_normalisation_toolbox_v2_2} & 0.0032 & 24.9928 & 0.7785 & 0.8571 \\
F (d-he3) \cite{Alsubaie2017} & 0.0033 & 24.7550 & 0.7525 & 0.8408 \\
P1 (d-he3) & \textbf{0.0018} & \textbf{27.3833} & \textbf{0.8275} & \textbf{0.8912} \\
P2a (d-he3) & \textbf{0.0018} & \textbf{27.3916} & \textbf{0.8318} & \textbf{0.8943} \\
P2b (d-he3) & \textbf{0.0018} & \textbf{27.3886} & \textbf{0.8312} & \textbf{0.8935} \\
\bottomrule
\end{tabular}}
\label{tab:Hstain_sep_comp}
\end{table}

\begin{table}[t]
\centering
\caption{\footnotesize{Comparisons of eosin (E) stain separation methods with the proposed approaches: P1, P2a, and P2b. Experiments ran on an in-house H\&E staining dataset (d-he1, d-he2, and d-he3) using full-reference quality metrics such as: mean-squared error (MSE; lower score is better) \cite{matlab_toolbox_ref}, peak signal-to-noise ratio (PSNR; higher score is better) \cite{matlab_toolbox_ref}, structural similarity index (SSIM; value closer to 1 is better) \cite{matlab_toolbox_ref}, and multiscale structural similarity index (MS-SSIM; value closer to 1 is better) \cite{matlab_toolbox_ref}. The default or recommended values are utilized from the corresponding references (methods A-F). Values highlighted in boldface showcase the three best-performing methods; instance-wise C, D, E, or F and instance-wise proposed methods P1, P2a, or P2b.}}
\vspace*{-0.75mm}
\resizebox{\columnwidth}{!}{
\begin{tabular}{lccccc}
\toprule
\textbf{Stain separation for E} & \textbf{MSE $\downarrow$} & \textbf{PSNR $\uparrow$} & \textbf{SSIM $\uparrow$} & \textbf{MS-SSIM $\uparrow$} \\
\midrule
A (d-he1) \cite{Ruifrok2001} & 0.0010 & 30.0729 & 0.7199 & 0.7758 \\
B (d-he1) \cite{Ruifrok2001} & 0.0010 & 30.0825 & 0.7201 & 0.7768 \\
C (d-he1) \cite{Khan2014, stain_normalisation_toolbox_v2_2, Ruifrok2001} & 0.0008 & 31.0277 & 0.7651 & 0.8172 \\
D (d-he1) \cite{Khan2014, stain_normalisation_toolbox_v2_2, Macenko2009} & \textbf{0.0007} & \textbf{31.2651} & \textbf{0.7768} & \textbf{0.8298} \\
E (d-he1) \cite{Khan2014, stain_normalisation_toolbox_v2_2} & 0.0010 & 29.9629 & 0.7435 & 0.7551 \\
F (d-he1) \cite{Alsubaie2017} & 0.0012 & 29.3336 & 0.6982 & 0.6986 \\
P1 (d-he1) & 0.0012 & 29.0773 & 0.7388 & 0.7270 \\
P2a (d-he1) & \textbf{0.0007} & \textbf{31.3994} & \textbf{0.7693} & \textbf{0.8376} \\
P2b (d-he1) & \textbf{0.0007} & \textbf{31.3985} & \textbf{0.7792} & \textbf{0.8418} \\
\hline
A (d-he2) \cite{Ruifrok2001} & 0.0018 & 27.4201 & 0.6019 & 0.6304 \\
B (d-he2) \cite{Ruifrok2001} & 0.0018 & 27.5423 & 0.6110 & 0.6378 \\
C (d-he2) \cite{Khan2014, stain_normalisation_toolbox_v2_2, Ruifrok2001} & \textbf{0.0010} & \textbf{29.9397} & \textbf{0.7179} & \textbf{0.7578} \\
D (d-he2) \cite{Khan2014, stain_normalisation_toolbox_v2_2, Macenko2009} & \textbf{0.0008} & \textbf{31.0731} & \textbf{0.7582} & \textbf{0.8053} \\
E (d-he2) \cite{Khan2014, stain_normalisation_toolbox_v2_2} & \textbf{0.0013} & \textbf{28.7390} & 0.7100 & 0.6921 \\
F (d-he2) \cite{Alsubaie2017} & 0.0015 & 28.2329 & 0.6574 & 0.6804 \\
P1 (d-he2) & 0.0040 & 24.0306 & \textbf{0.7289} & \textbf{0.7011} \\
P2a (d-he2) & 0.0019 & 27.2997 & 0.6827 & 0.6489 \\
P2b (d-he2) & 0.0018 & 27.4599 & 0.6877 & 0.6689 \\
\hline
A (d-he3) \cite{Ruifrok2001} & 0.0005 & 33.1732 & 0.8716 & 0.8738 \\
B (d-he3) \cite{Ruifrok2001} & 0.0005 & \textbf{33.2060} & 0.8723 & 0.8748 \\
C (d-he3) \cite{Khan2014, stain_normalisation_toolbox_v2_2, Ruifrok2001} & 0.0005 & 32.9978 & 0.8741 & 0.8746 \\
D (d-he3) \cite{Khan2014, stain_normalisation_toolbox_v2_2, Macenko2009} & 0.0005 & 33.0405 & 0.8742 & 0.8750 \\
E (d-he3) \cite{Khan2014, stain_normalisation_toolbox_v2_2} & \textbf{0.0005} & 33.1844 & \textbf{0.8885} & \textbf{0.8827} \\
F (d-he3) \cite{Alsubaie2017} & \textbf{0.0004} & \textbf{33.9079} & \textbf{0.8797} & \textbf{0.8851} \\
P1 (d-he3) & \textbf{0.0005} & \textbf{33.4580} & \textbf{0.8981} & \textbf{0.8951} \\
P2a (d-he3) & 0.0005 & 32.6683 & 0.8694 & 0.8661 \\
P2b (d-he3) & 0.0006 & 32.1628 & 0.8669 & 0.8609 \\
\bottomrule
\end{tabular}}
\label{tab:Estain_sep_comp}
\end{table}

\begin{table}[t]
\centering
\caption{\footnotesize{Comparisons of four classification methods (A-D) - using feature extraction on aggregated features computed from 91 combinations ($q_-$ and $q_+$, maps $\boldsymbol{f}()\boldsymbol{f}$ and $\boldsymbol{f}()\boldsymbol{g}$ where $\boldsymbol{f} = \boldsymbol{\mu_{i}}$ and $\boldsymbol{g} = \boldsymbol{\mu_{j}}$, with $\boldsymbol{i} < \boldsymbol{j}$ and for $\boldsymbol{\mu_{1}}$ to $\boldsymbol{\mu_{13}}$ --- e.g. $\boldsymbol{\mu_{3}}()\boldsymbol{\mu_{3}}$, $\boldsymbol{\mu_{3}}()\boldsymbol{\mu_{4}}$, but not $\boldsymbol{\mu_{3}}()\boldsymbol{\mu_{2}}$) of pure unit quaternions shown in Fig. \ref{fig:m1m13} - on a collection of histological images (H\&E staining) of human colorectal cancer \cite{Kather2016} (d-ml). Machine learning (ML) methods A-D are, respectively: random subspace ensemble \cite{TinKamHo1998}, \textit{k}-nearest neighbors \cite{Sproull1991}, support vector machines \cite{ChihWeiHsu2002}, and a feedforward fully connected neural network \cite{Liu1989}. The displayed classification metrics are reported for the test set. Higher scores are better for sensitivity (I), specificity (II), precision (III), and F1 score (IV); lower scores are better for fall-out (V). Values highlighted in boldface showcase the best-performing classification methods for $q_-$ and $q_+$.}}
\vspace*{-0.75mm}
\resizebox{\columnwidth}{!}{
\begin{tabular}{lcccc}
\toprule
\multirow{1}*{\textbf{Datasets}} & \multicolumn{4}{c}{\textbf{Supervised classification methods}} \\
\midrule
d-ml: original dataset \cite{Kather2016} & \textbf{A} & \textbf{B} & \textbf{C} & \textbf{D} \\
Sensitivity (\textbf{I}) $\uparrow$ & 0.9128 & 0.8448 & 0.9079 & 0.9096 \\
Specificity (\textbf{II}) $\uparrow$ & 0.9875 & 0.9778 & 0.9869 & 0.9871 \\
Precision (\textbf{III}) $\uparrow$ & 0.9161 & 0.8501 & 0.9077 & 0.9111 \\
F1 score (\textbf{IV}) $\uparrow$ & 0.9137 & 0.8445 & 0.9072 & 0.9095 \\
Fall-out (\textbf{V}) $\downarrow$ & 0.0125 & 0.0222 & 0.0131 & 0.0129 \\
\hline
$q_-$, output features & \textbf{A} & \textbf{B} & \textbf{C} & \textbf{D} \\
Sensitivity (\textbf{I}) $\uparrow$ & 0.9247 & \textbf{0.9208} & 0.9424 & 0.9143 \\
Specificity (\textbf{II}) $\uparrow$ & 0.9893 & \textbf{0.9887} & 0.9918 & 0.9878 \\
Precision (\textbf{III}) $\uparrow$ & 0.9281 & \textbf{0.9215} & 0.9429 & 0.9142 \\
F1 score (\textbf{IV}) $\uparrow$ & 0.9256 & \textbf{0.9203} & 0.9425 & 0.9141 \\
Fall-out (\textbf{V}) $\downarrow$ & 0.0107 & \textbf{0.0113} & 0.0082 & 0.0122 \\
\hline
$q_+$, output features & \textbf{A} & \textbf{B} & \textbf{C} & \textbf{D} \\
Sensitivity (\textbf{I}) $\uparrow$ & \textbf{0.9335} & 0.9200 & \textbf{0.9536} & \textbf{0.9336} \\
Specificity (\textbf{II}) $\uparrow$ & \textbf{0.9905} & 0.9886 & \textbf{0.9934} & \textbf{0.9905} \\
Precision (\textbf{III}) $\uparrow$ & \textbf{0.9350} & 0.9202 & \textbf{0.9543} & \textbf{0.9345} \\
F1 score (\textbf{IV}) $\uparrow$ & \textbf{0.9337} & 0.9198 & \textbf{0.9538} & \textbf{0.9338} \\
Fall-out (\textbf{V}) $\downarrow$ & \textbf{0.0095} & 0.0114 & \textbf{0.0066} & \textbf{0.0095} \\
\bottomrule
\end{tabular}}
\label{tab:ml_comp1}
\end{table}

\begin{table}[t]
\centering
\caption{\footnotesize{Comparisons of original data and proposed transformations - computed from 91 combinations ($q_-$ and $q_+$, maps $\boldsymbol{f}()\boldsymbol{f}$ and $\boldsymbol{f}()\boldsymbol{g}$ where $\boldsymbol{f} = \boldsymbol{\mu_{i}}$ and $\boldsymbol{g} = \boldsymbol{\mu_{j}}$, with $\boldsymbol{i} < \boldsymbol{j}$ and for $\boldsymbol{\mu_{1}}$ to $\boldsymbol{\mu_{13}}$ --- e.g. $\boldsymbol{\mu_{3}}()\boldsymbol{\mu_{3}}$, $\boldsymbol{\mu_{3}}()\boldsymbol{\mu_{4}}$, but not $\boldsymbol{\mu_{3}}()\boldsymbol{\mu_{2}}$) of pure unit quaternions shown in Fig. \ref{fig:m1m13} - using the pre-trained and untrained ResNet-101 \cite{HeZRS15} and EfficientNet-b0 \cite{Tan2019EfficientNetRM} convolutional neural networks on a collection of histological images (H\&E staining) of human colorectal cancer \cite{Kather2016} (d-ml). The displayed classification metrics are reported for the test set and are arithmetic means $\pm$ standard deviations of three iterations. Higher scores are better for sensitivity (I), specificity (II), precision (III), and F1 score (IV); lower scores are better for fall-out (V). Values highlighted in boldface showcase the two best-performing transformations for each dataset.}}
\vspace*{-0.75mm}
\resizebox{\columnwidth}{!}{%
\begin{tabular}{lccccc}
\toprule
{\textbf{Datasets}} & \multicolumn{5}{c}{\textbf{Classification metrics}}  \\
\multirow{1}*{\textit{pre-trained ResNet-101}} & \textbf{I $\uparrow$} & \textbf{II $\uparrow$} & \textbf{III $\uparrow$} & \textbf{IV $\uparrow$} & \textbf{V $\downarrow$} \\
\midrule
Original data (d-ml) \cite{Kather2016} & 0.9410 & 0.9916 & 0.9461 & 0.9407 & 0.0084 \\
 & $\pm$0.0095 & $\pm$0.0014 & $\pm$0.0072 & $\pm$0.0098 & $\pm$0.0014 \\
$q_+$, $\boldsymbol{f} = \boldsymbol{\mu_{4}}$, $\boldsymbol{g} = \boldsymbol{\mu_{5}}$ & \textbf{0.9661} & \textbf{0.9952} & \textbf{0.9664} & \textbf{0.9659} & \textbf{0.0048} \\
 & $\pm$0.0079 & $\pm$0.0011 & $\pm$0.0078 & $\pm$0.0081 & $\pm$0.0011 \\
$q_+$, $\boldsymbol{f} = \boldsymbol{\mu_{8}}$, $\boldsymbol{g} = \boldsymbol{\mu_{11}}$ & \textbf{0.9645} & \textbf{0.9949} & \textbf{0.9648} & \textbf{0.9644} & \textbf{0.0051} \\
 & $\pm$0.0005 & $\pm$0.0001 & $\pm$0.0003 & $\pm$0.0004 & $\pm$0.0001 \\
\hline
{} & \multicolumn{5}{c}{} \\
\multirow{1}*{\textit{untrained ResNet-101}} & \textbf{I $\uparrow$} & \textbf{II $\uparrow$} & \textbf{III $\uparrow$} & \textbf{IV $\uparrow$} & \textbf{V $\downarrow$} \\
\midrule
Original data (d-ml) \cite{Kather2016} & 0.8325 & 0.9761 & 0.8571 & 0.8310 & 0.0239 \\
 & $\pm$0.0242 & $\pm$0.0035 & $\pm$0.0152 & $\pm$0.0242 & $\pm$0.0035 \\
$q_-$, $\boldsymbol{f} = \boldsymbol{\mu_{8}}$, $\boldsymbol{g} = \boldsymbol{\mu_{13}}$ & \textbf{0.8835} & \textbf{0.9834} & \textbf{0.8927} & \textbf{0.8830} & \textbf{0.0166} \\
 & $\pm$0.0143 & $\pm$0.0020 & $\pm$0.0080 & $\pm$0.0142 & $\pm$0.0020 \\
$q_+$, $\boldsymbol{f} = \boldsymbol{\mu_{8}}$, $\boldsymbol{g} = \boldsymbol{\mu_{11}}$ & \textbf{0.8806} & \textbf{0.9829} & \textbf{0.8936} & \textbf{0.8795} & \textbf{0.0171} \\
 & $\pm$0.0312 & $\pm$0.0045 & $\pm$0.0163 & $\pm$0.0045 & $\pm$0.0342 \\
\hline
{} & \multicolumn{5}{c}{} \\
\multirow{1}*{\textit{pre-trained EfficientNet-b0}} & \textbf{I $\uparrow$} & \textbf{II $\uparrow$} & \textbf{III $\uparrow$} & \textbf{IV $\uparrow$} & \textbf{V $\downarrow$} \\
\midrule
Original data (d-ml) \cite{Kather2016} & 0.9370 & 0.9910 & 0.9388 & 0.9371 & 0.0090 \\
 & $\pm$0.0057 & $\pm$0.0008 & $\pm$0.0050 & $\pm$0.0055 & $\pm$0.0008 \\
$q_+$, $\boldsymbol{f} = \boldsymbol{\mu_{8}}$, $\boldsymbol{g} = \boldsymbol{\mu_{9}}$ & \textbf{0.9634} & \textbf{0.9948} & \textbf{0.9637} & \textbf{0.9634} & \textbf{0.0052} \\
 & $\pm$0.0005 & $\pm$0.0001 & $\pm$0.0005 & $\pm$0.0005 & $\pm$0.0001 \\
$q_+$, $\boldsymbol{f} = \boldsymbol{\mu_{1}}$, $\boldsymbol{g} = \boldsymbol{\mu_{2}}$ & \textbf{0.9631} & \textbf{0.9947} & \textbf{0.9634} & \textbf{0.9632} & \textbf{0.0053} \\
 & $\pm$0.0000 & $\pm$0.0000 & $\pm$0.0000 & $\pm$0.0000 & $\pm$0.0000 \\
\hline
{} & \multicolumn{5}{c}{} \\
\multirow{1}*{\textit{untrained EfficientNet-b0}} & \textbf{I $\uparrow$} & \textbf{II $\uparrow$} & \textbf{III $\uparrow$} & \textbf{IV $\uparrow$} & \textbf{V $\downarrow$} \\
\midrule
Original data (d-ml) \cite{Kather2016} & 0.8109 & 0.9730 & 0.8219 & 0.8098 & 0.0270 \\
 & $\pm$0.0245 & $\pm$0.0035 & $\pm$0.0229 & $\pm$0.0265 & $\pm$0.0035 \\
$q_+$, $\boldsymbol{f} = \boldsymbol{\mu_{5}}$, $\boldsymbol{g} = \boldsymbol{\mu_{6}}$ & \textbf{0.8689} & \textbf{0.9813} & \textbf{0.8812} & \textbf{0.8699} & \textbf{0.0187} \\
 & $\pm$0.0044 & $\pm$0.0006 & $\pm$0.0045 & $\pm$0.0043 & $\pm$0.0006 \\
$q_+$, $\boldsymbol{f} = \boldsymbol{\mu_{4}}$, $\boldsymbol{g} = \boldsymbol{\mu_{11}}$ & \textbf{0.8571} & \textbf{0.9796} & \textbf{0.8712} & \textbf{0.8543} & \textbf{0.0204} \\
 & $\pm$0.0352 & $\pm$0.0050 & $\pm$0.0247 & $\pm$0.0390 & $\pm$0.0050 \\
\bottomrule
\end{tabular}}
\label{tab:ml_comp2}
\end{table}

\subsection*{\normalsize{Computational stain separation}}
Figs. \ref{fig:img_unsupstainsep_bio}-\ref{fig:img_supstainsep_bio} illustrate computational stain separation using the proposed approaches on histological close-up image sections and on synthetic color images. In this context, transformations can have geometric meaning (e.g. Fig. \ref{fig:m1m13}, $\boldsymbol{\mu_7}$ is the direction corresponding to the luminance or gray-line axis \cite{Moxey2003, Ell2007a}), but also can be based on a computational method that is utilized to estimate the stain separation matrix (Macenko's method) \cite{Macenko2009, stain_normalisation_toolbox_v2_2}. The synthetic model images validated the proposed approaches for stain separation, thus approx. matching the outcome of their real-world counterparts. The synthetic images in Fig. \ref{fig:img_unsupstainsep_bio}(e-h) correspond to the histological images of Fig. \ref{fig:img_unsupstainsep_bio}(a-d). The stain-separated images in Fig. \ref{fig:img_unsupstainsep_bio}(e1-h1) correspond to the ones of Fig. \ref{fig:img_unsupstainsep_bio}(a1-d1). Similarly, the synthetic images in Fig. \ref{fig:img_supstainsep_bio}(f-j) correspond to the histological images of Fig. \ref{fig:img_supstainsep_bio}(a-e). The stain-separated images in Fig. \ref{fig:img_supstainsep_bio}(f1-j1) correspond to the ones of Fig. \ref{fig:img_supstainsep_bio}(a1-e1). For example, the immunostain separation result in Fig. \ref{fig:img_supstainsep_bio}(j1) shows the first component stain (approx. brown color), and Figs. \ref{fig:img_supstainsep_bio}(e1-e2) show the first two component stains (approx. brown and red color). For the double immunostain (Fig. \ref{fig:img_supstainsep_bio}(e)), two re-colorization steps are required because the number of dominant colors in this image is greater than two. 

Tables \ref{tab:Hstain_sep_comp}-\ref{tab:Estain_sep_comp} show comparisons among several stain separation methods (A-F) with the proposed approaches (P1, P2a, and P2b) using an in-house dataset (d-he1, d-he2, and d-he3) comprising three hematoxylin only (H-only), three eosin only (E-only), and three H\&E whole slide images. The H-only and E-only images obtained from the lab were utilized as ground truth references. Again, transformations can have geometric meaning, but also be defined by the user, or computed by automated methods. For presentation purposes, Table \ref{tab:Hstain_sep_comp} presents the results for hematoxylin (H), while Table \ref{tab:Estain_sep_comp} presents the results for eosin (E). Methods A-B are based on Ref. \cite{Ruifrok2001} with two different built-in stain vectors. Methods C-E are based on Refs. \cite{Khan2014, stain_normalisation_toolbox_v2_2} with: i) a single built-in stain vector \cite{Ruifrok2001}, ii) Macenko's method (for estimating the stain separation matrix) \cite{Macenko2009}, and iii) the stain color descriptor method (for estimating the stain separation matrix) \cite{Khan2014, stain_normalisation_toolbox_v2_2}, respectively. Method F is based on Ref. \cite{Alsubaie2017} in which a multi-resolution wavelet representation of the image is used for estimating the stain mixing matrix data. Fig. \ref{fig:img_stainsep_bio} exhibits computational stain separation in an H\&E image (d-he1). Along with the ground truth images (H-only and E-only) obtained from the lab, renditions from literature methods and proposed approaches are also shown (Tables \ref{tab:Hstain_sep_comp}-\ref{tab:Estain_sep_comp}).

\subsection*{\normalsize{Enhancing machine/deep learning workflows}}
Table \ref{tab:ml_comp1} shows comparisons between original and transformed image data with four classification methods (random subspace ensemble \cite{TinKamHo1998}, \textit{k}-nearest neighbors \cite{Sproull1991}, support vector machines \cite{ChihWeiHsu2002}, and a feedforward fully connected neural network \cite{Liu1989}) utilizing aggregated feature extraction on a public histological image dataset \cite{Kather2016}. Table \ref{tab:ml_comp2} shows comparisons between original and transformed image data with two deep learning pipelines (ResNet-101 \cite{HeZRS15} and EfficientNet-b0 \cite{Tan2019EfficientNetRM}) utilizing different transformation combinations on the same public dataset. Table \ref{tab:ml_comp3} shows comparisons between transformed (utilized as training set) and original (utilized as test set) image data with the same deep learning pipelines on the same public dataset. The aim was not to compare the efficiency of the tested classifiers but to assess the performance of the proposed transformations.

\begin{table}[t]
\centering
\caption{\footnotesize{Comparisons of varying amounts of transformed data (100$\%$, 50$\%$, and 20$\%$) - computed from 91 combinations ($q_-$ and $q_+$, maps $\boldsymbol{f}()\boldsymbol{f}$ and $\boldsymbol{f}()\boldsymbol{g}$ where $\boldsymbol{f} = \boldsymbol{\mu_{i}}$ and $\boldsymbol{g} = \boldsymbol{\mu_{j}}$, with $\boldsymbol{i} < \boldsymbol{j}$ and for $\boldsymbol{\mu_{1}}$ to $\boldsymbol{\mu_{13}}$ --- e.g. $\boldsymbol{\mu_{3}}()\boldsymbol{\mu_{3}}$, $\boldsymbol{\mu_{3}}()\boldsymbol{\mu_{4}}$, but not $\boldsymbol{\mu_{3}}()\boldsymbol{\mu_{2}}$) of pure unit quaternions shown in Fig. \ref{fig:m1m13} - using the pre-trained ResNet-101 \cite{HeZRS15} and EfficientNet-b0 \cite{Tan2019EfficientNetRM} convolutional neural networks on a collection of histological images (H\&E staining) of human colorectal cancer \cite{Kather2016} (d-ml). The transformed data were utilized as the training set and the original data were the test set (75:25). The displayed classification metrics are reported for the test set and are arithmetic means $\pm$ standard deviations of three iterations. Higher scores are better for sensitivity (I), specificity (II), precision (III), and F1 score (IV); lower scores are better for fall-out (V). Values highlighted in boldface showcase better performance than the original dataset reported in Table \ref{tab:ml_comp2}.}}
\vspace*{-0.75mm}
\resizebox{\columnwidth}{!}{%
\begin{tabular}{lccccc}
\toprule
{\textbf{Datasets} (d-ml) \cite{Kather2016}} & \multicolumn{5}{c}{\textbf{Classification metrics}}  \\
\multirow{1}*{\textit{pre-trained ResNet-101}} & \textbf{I $\uparrow$} & \textbf{II $\uparrow$} & \textbf{III $\uparrow$} & \textbf{IV $\uparrow$} & \textbf{V $\downarrow$} \\
\midrule
Transformed data, 100$\%$, $q_+$ & \textbf{0.9861} & \textbf{0.9980} & \textbf{0.9867} & \textbf{0.9862} & \textbf{0.0020} \\
 & $\pm$0.0044 & $\pm$0.0006 & $\pm$0.0041 & $\pm$0.0044 & $\pm$0.0006 \\
Transformed data, 50$\%$, $q_-$ & \textbf{0.9762} & \textbf{0.9966} & \textbf{0.9774} & \textbf{0.9762} & \textbf{0.0034} \\
 & $\pm$0.0028 & $\pm$0.0004 & $\pm$0.0025 & $\pm$0.0028 & $\pm$0.0004 \\
Transformed data, 20$\%$, $q_-$ & \textbf{0.9655} & \textbf{0.9951} & \textbf{0.9675} & \textbf{0.9656} & \textbf{0.0049} \\
 & $\pm$0.0029 & $\pm$0.0004 & $\pm$0.0023 & $\pm$0.0030 & $\pm$0.0004 \\
\hline
{} & \multicolumn{5}{c}{} \\
\multirow{1}*{\textit{pre-trained EfficientNet-b0}} & \textbf{I $\uparrow$} & \textbf{II $\uparrow$} & \textbf{III $\uparrow$} & \textbf{IV $\uparrow$} & \textbf{V $\downarrow$} \\
\midrule
Transformed data, 100$\%$, $q_-$ & \textbf{0.9907} & \textbf{0.9987} & \textbf{0.9909} & \textbf{0.9906} & \textbf{0.0013} \\
 & $\pm$0.0032 & $\pm$0.0005 & $\pm$0.0031 & $\pm$0.0032 & $\pm$0.0005 \\
Transformed data, 50$\%$, $q_-$ & \textbf{0.9880} & \textbf{0.9983} & \textbf{0.9883} & \textbf{0.9880} & \textbf{0.0017} \\
 & $\pm$0.0032 & $\pm$0.0005 & $\pm$0.0031 & $\pm$0.0032 & $\pm$0.0005 \\
Transformed data, 20$\%$, $q_-$ & \textbf{0.9506} & \textbf{0.9929} & \textbf{0.9552} & \textbf{0.9509} & \textbf{0.0071} \\
 & $\pm$0.0114 & $\pm$0.0016 & $\pm$0.0088 & $\pm$0.0111 & $\pm$0.0016 \\
\bottomrule
\end{tabular}}
\label{tab:ml_comp3}
\end{table}

\subsection*{\normalsize{Time complexity}}
\begin{figure}[t]
\centering
\includegraphics[width=1.0\linewidth]{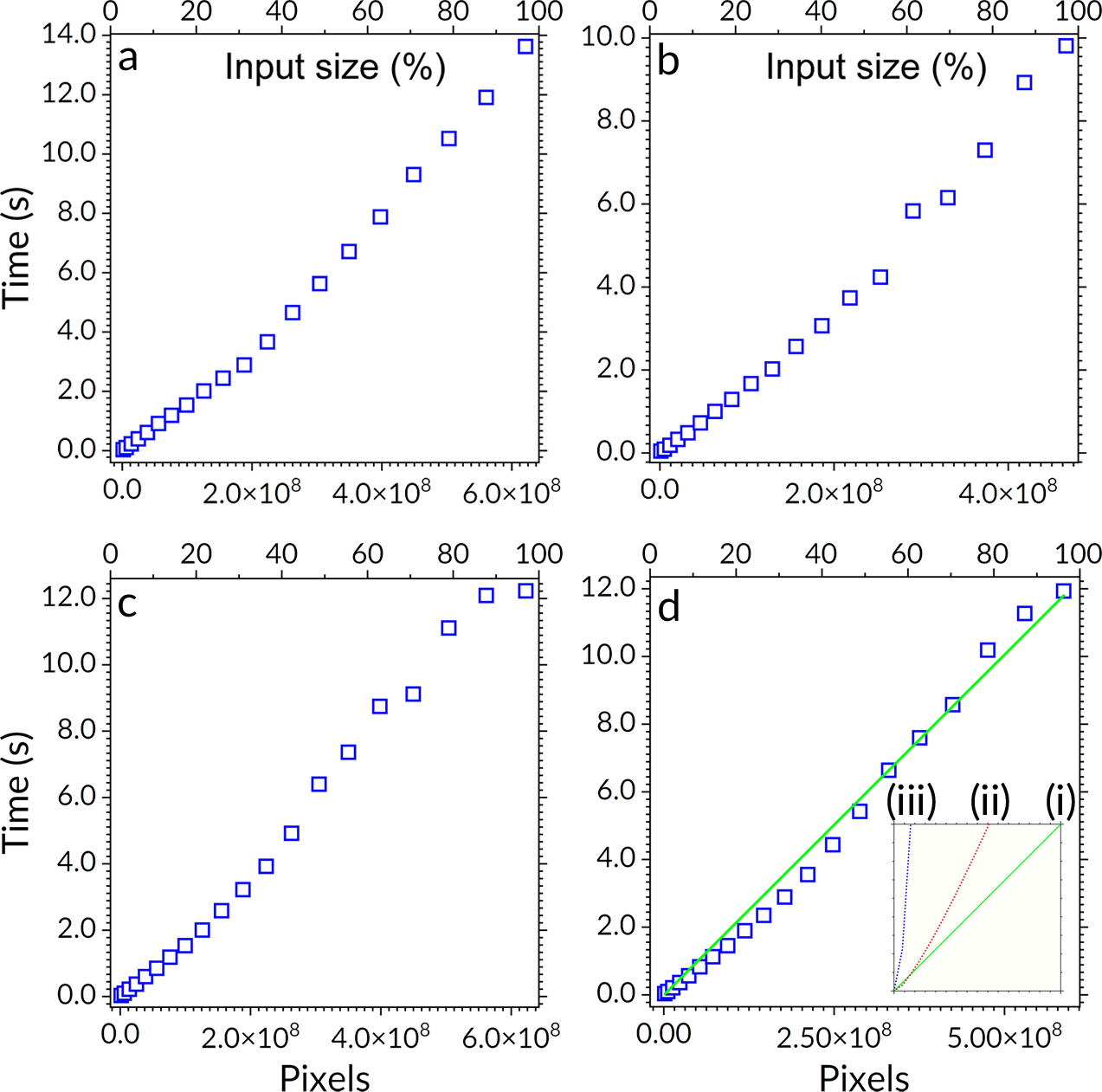}
\vspace{-12pt}
\caption{\footnotesize{Processing time of the 2D orthogonal planes split on three H\&E whole slide images (a-c). The bottom $x$-axis shows the variable input image size measured in pixels (width$\times$height), while the top $x$-axis shows the same expressed as a percentage. The $y$-axis shows the time required for processing in seconds (s). Graphs (a-c) show the average processing time computed from re-colorization runs for $q_+$ and $q_-$, $\boldsymbol{f} = \boldsymbol{\mu_{7}}$. Graph (d) shows results from the three histological images (average number of pixels and processing time) fitted to a linear model with a coefficient of determination $R^2 = 0.995$. The embedded graph in (d) shows commonly encountered time complexities: (i) linear time $O(n)$, (ii) linearithmic time $O(n\log{}n)$, and (iii) quadratic time $O(n^2)$. Pure unit quaternions $\boldsymbol{\mu_{1}}$-$\boldsymbol{\mu_{13}}$ are shown in Fig. \ref{fig:m1m13}.}}
\label{fig:time_compl}
\end{figure}

The processing time required by the 2D orthogonal planes split (OPS) to run as a function of variable input image size is shown in Fig. \ref{fig:time_compl} for three H\&E whole slide images. Results indicate that time complexity matches a linear behavior $O(n)$. Essentially, the runtime of the OPS grows almost linearly with the input image size, meaning that it would take proportionally longer to produce output as the input grows. In general, any algorithm that performs a constant number of operations on each element has linear time complexity \cite{Havill2020}.

\section*{\large\uppercase{Discussion}}
\noindent
The 2D orthogonal planes split framework provides clear advantages over standard image processing approaches, including mathematical generality, ability to process color geometrically, computational efficiency, and versatility across application domains. By decomposing each pixel, expressed as a quaternion, into pairs of orthogonal 2D planes, the method enables direct and coherent geometric processing of color data, supporting a broad set of workflows that are effective for both natural and biomedical images. Unlike data-driven techniques, these workflows rely solely on basic arithmetic and matrix operations, yet achieve results that match or surpass established methods. Leveraging the underlying algebraic structure of image data, the proposed workflows allow for transformations that are not only mathematically consistent but also often lead to interpretable outcomes.
\parskip=5pt

\noindent
\textbf{Workflows for natural images}. With basic quaternionic algebraic operations, different visual representations of natural images can be obtained (Fig. \ref{fig:img_recol}). By regulating color, the proposed approach can contribute to distinct, vivid, and aesthetically relevant visuals. The process can modulate how colors are perceived thus affecting complex scene perception. This can be useful in a variety of fields such as multimedia, graphic design, social media, and other visual media industries, where alternative visual representations may engage viewers and aid in the creation of impactful content. Image de-colorization is still an open problem in computer vision. The proposed approach (Figs. \ref{fig:img_decol1}-\ref{fig:img_decol2}, and Table \ref{tab:decol_comp}) offers an uncomplicated way to convert color images into grayscale that allows one to shift from color-based elements to other visual aspects such as texture, contrast, and light. This can reveal intricate details or emphasize certain features that might be overshadowed by color and can play a role in photography, graphic design, and visual storytelling. In the context of scientific visualization or accessibility, it can be useful for simplifying complex images, ensuring clarity, and improving visual comprehension for audiences with color vision deficiencies. Moreover, producing an effective grayscale representation can reduce the computational burden of machine/deep learning models employed in edge computing applications. Table \ref{tab:decol_comp} shows that even the proposed method P1 that uses the pure unit quaternion $\boldsymbol{\mu_7}$ shown in Fig. \ref{fig:m1m13} (direction corresponding to the luminance or gray-line axis \cite{Moxey2003, Ell2007a}) outputs a grayscale image that is computationally inexpensive to obtain with satisfactory conversion performance. In addition, the proposed approach offers an adjustable method to enhance the contrast of natural images (Fig. \ref{fig:img_contr} and Table \ref{tab:contr_comp}). Boosting contrast can improve visual quality and interpretability, e.g. in photography and remote sensing. Utilizing histology images as controls (Fig. \ref{fig:img_contr}) shows that there are no noticeable chrominance or luminance distortions in the enhanced images. Histology images are typically less complex in visual texture and distribution of hue and luminance values compared to natural images, thus distortions from contrast adjustment operations can be readily perceived.
\parskip=5pt

\noindent
\textbf{Workflows for biomedical images}. Image contrast enhancement (Fig. \ref{fig:img_contr} and Table \ref{tab:contr_comp}) in medical images (e.g. in radiology) improves the distinction of anatomical structures, tissues, and potential abnormalities. This is especially true for low-dose computed tomography (CT) images (utilized in this study), which provide fertile ground for computational methods that can improve image quality and diagnostic accuracy \cite{Shi2016}. By increasing the distinction between lighter and darker regions, contrast enhancement highlights important details and features that may otherwise be difficult to discern. In a more practical context, contrast enhancement can also aid in the detection and analysis of objects or areas of interest. Tuning the parameters $\alpha$, $\beta$, $\gamma$, and $\delta$ (refer to \textit{Materials and Methods}, Eq. \ref{eq:contrastenhancement2}) allows for utilizing the proposed method in the biomedical setting, thus offering flexibility by accommodating different types of images. The workflows that deal with biomedical re-colorization and re-staining (targeted recolorization) as well as their use in the visualization of histological images can be discussed under the same umbrella. In the case of re-colorization (Fig. \ref{fig:img_recol_bio}), results showed that certain cellular structures or tissue components in histology images become more distinguishable and others become part of the background, similar to switching on and off histological features (emphasizing specific biomarkers or cellular patterns) without any other prior operation, e.g. segmentation. In the case of re-staining, as an example, the eosin stain (Fig. \ref{fig:img_restain_bio}(b)) is visually amplified making it more easily identifiable in the re-stained versions (Fig. \ref{fig:img_restain_bio}(b1, b2)). In principle, both approaches have the potential to highlight different stains and assist in the identification of key histological features, providing flexibility for research and educational purposes. Furthermore, an obvious application of re-staining is to produce colorblind-friendly renditions (e.g. Fig. \ref{fig:img_restain_bio}(a2, b2, c2)); this adaptation often involves substituting colors with hues that are distinct to all viewers. Colorblind-friendly palettes often incorporate colors like blue and red, and various shades derived from these hues, e.g. blue and orange. For most types of color blindness, the perception of blue is largely unaffected. Colorblind-friendly images can be useful in digital pathology, promoting inclusivity and benefiting research and education. Both re-colorization and re-staining approaches can be utilized to visualize histological images (Fig. \ref{fig:img_large_scale_vis_bio}), e.g. highlighting a biomarker and revealing the surrounding histological structures. For example in Fig. \ref{fig:img_large_scale_vis_bio}(b1.1, b1.2, b2.1, b2.2), re-colorization followed by re-staining produced a more uniform and optimized visual outcome for observing the spatial organization of T lymphocytes (CD3) and adjacent histological structures. In addition, in the case of in situ hybridization (Fig. \ref{fig:img_large_scale_vis_bio}(c1)) or faint staining, the proposed approach can improve the viewing and identification of biomarker expression and cellular localization. Turning to stain separation, its application to tissue samples allows for the isolation of different stains, enabling further analysis of distinct cellular components. It is valuable in automated digital pathology workflows, enhancing the precision of quantitative analyses, e.g. cell counting, biomarker intensity evaluation, etc. thus providing a better understanding of tissue architecture and histopathological features. The proposed approach partitioned the contextual stain information of histological stains (such as hematoxylin and eosin; H\&E) as well as immunohistochemical markers into distinct channels representing each stain (Figs. \ref{fig:img_unsupstainsep_bio}-\ref{fig:img_stainsep_bio}, and Tables \ref{tab:Hstain_sep_comp}-\ref{tab:Estain_sep_comp}). Versatility was shown in Figs. \ref{fig:img_unsupstainsep_bio}-\ref{fig:img_supstainsep_bio} with transformations that have geometric meaning, as well as transformations whose values derive from known computational methods for obtaining the stain separation matrix. Another option would have been to utilize the stain information sampled from the image itself, in cases where other automated methods (for obtaining the stain separation matrix) produce sub-optimal results. It was demonstrated, using an in-house H\&E dataset (with ground truth images obtained from the lab), that the proposed approach was comparable to different algorithms (widely) used in the literature, e.g. color deconvolution (Tables \ref{tab:Hstain_sep_comp}-\ref{tab:Estain_sep_comp} and Fig. \ref{fig:img_stainsep_bio}). For the machine/deep learning computational experiments on histology images, three main approaches were tested: i) the conventional way with feature extraction and machine learning (Table \ref{tab:ml_comp1}), ii) the use of deep learning pipelines on the transformed images as the test set (Table \ref{tab:ml_comp2}), and iii) the use of deep learning pipelines on the transformed images as the training set and the original images as the test set (Table \ref{tab:ml_comp3}). Pertinent to the second approach, some of the proposed transformations were more effective at enhancing deep learning model performance by changing the visual representation of the input images. These models could interpret data patterns more effectively, reducing noise, and making training more efficient. The third approach integrated all proposed transformations to resolve the issue of evaluating them individually. For example, using half of the transformed images for training resulted in better performance at distinguishing the different classes of the original histological dataset, compared to utilizing a portion of the original images as the training set.
\parskip=5pt

\noindent
\textbf{Time complexity}. Linear-time algorithms are generally considered to be very fast, with the underlying issue of scalability \cite{Havill2020}. On one hand, gigapixel photography produces natural images with ultra-high resolution. On the other hand, whole-slide imaging of histology enables the detailed spatial interrogation of the entire tissue landscape in expansive high-resolution images containing billions of pixels \cite{Valous2020}. Considering that the running time of the 2D orthogonal planes split increases, at most, linearly with the size of the input (Fig. \ref{fig:time_compl}) and also given that system memory resources are typically finite, it follows that (for larger images and for achieving near real-time processing performance) further optimizations should be sought, e.g. image tiling, processing pixels in parallel, etc.
\parskip=5pt

\begin{figure}[t]
\centering
\includegraphics[width=1.0\linewidth]{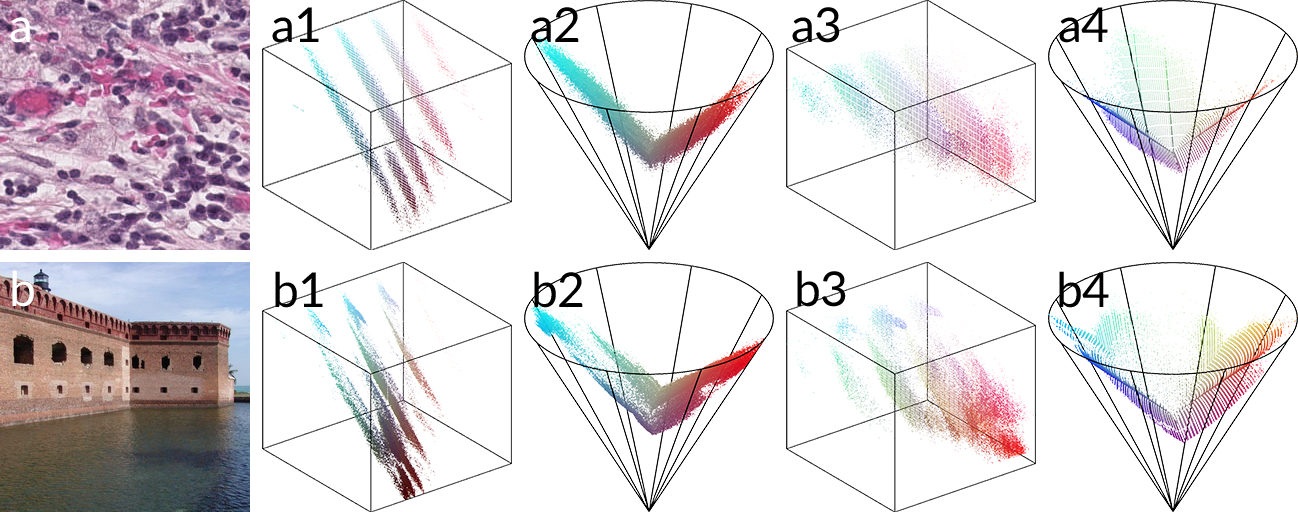}
\vspace{-12pt}
\caption{\footnotesize{3D color gamut visualizations as point clouds in the RGB and HSV color spaces of an H\&E close-up image section (a) and a natural image (b) from the outdoor scene dataset \cite{Wang2018}. The visualizations correspond to the following renditions: (a1, b1) $q_-$, $\boldsymbol{f} = \boldsymbol{\mu_{5}}$, $\boldsymbol{g} = \boldsymbol{\mu_{9}}$, (a2, b2) $q_-$, $\boldsymbol{f} = \boldsymbol{\mu_{7}}$, $\boldsymbol{g} = \boldsymbol{\mu_{8}}$, (a3, b3) $q_+$, $\boldsymbol{f} = \boldsymbol{\mu_{12}}$, $\boldsymbol{g} = \boldsymbol{\mu_{13}}$, and (a4, b4) $q_+$, $\boldsymbol{f} = \boldsymbol{\mu_{5}}$. Pure unit quaternions $\boldsymbol{\mu_{1}}$-$\boldsymbol{\mu_{13}}$ are shown in Fig. \ref{fig:m1m13}. Images have $250^2$ pixels (natural image was resized from its original dimensions).}}
\label{fig:img_color_vis}
\end{figure}

\noindent
\textbf{Study limitations and further development}. This study utilized a variety of pure unit quaternions (Fig. \ref{fig:m1m13}) for altering the color appearance of natural and biomedical images. Certainly, other quaternions can be employed for such transformations. A valuable approach for further development in identifying color patterns and selecting the most suitable transformations would be to visualize the color gamut of the transformed images. Fig. \ref{fig:img_color_vis} shows sample 3D color gamut visualizations of a histological and natural image - for different re-colorizations - in the RGB and HSV color spaces. These point clouds signify that similarities exist between transformations (as evidenced by the visualizations in the two color spaces) regardless of the unrelated visual content of the two images. The shape (spatial distribution), density (point concentration), and topology (point connectivity) of the color gamut point cloud may correspond to differences in the visual appearance of the transformed images, e.g. distinct color patterns, color smoothness, color vividness, etc. Furthermore, the re-colorization and re-staining methods would benefit from the development of a whole-slide viewer software package that can combine efficient image visualization, user-friendly annotation tools, and processing capabilities pertinent to the proposed approaches. The software would incorporate useful presets for common histological images that would allow fast visualization of the original and rendered slides with synchronized views. Naturally, the viewer would take advantage of the parallel architectures in modern computing systems to process the pixel-based proposed methods efficiently. Likewise, for stain separation, further development of a plugin for ImageJ/Fiji \cite{Schneider2012, Schindelin2012} or QuPath \cite{Bankhead2017} would allow researchers to decompose histological images into their stained components and then utilize them in quantification workflows. The proposed approaches for the modified representation of histological images can be utilized in research and education, however, further development is required to ensure their clinical utility and effectiveness. Pathologists are trained to interpret standard histological images, which have well-established visual features and patterns critical for accurate diagnosis. Introducing novel visualizations or altering the conventional appearance of histology images could disrupt this process, necessitating further retraining for pathologists to adapt to the new format. Moreover, any changes to standard practices must be rigorously validated to demonstrate that they preserve or enhance diagnostic accuracy and efficiency. This requires a larger clinical study - which is outside the scope of this work - that would refine the proposed methods so that they align closely with established histological conventions or incorporate clear, intuitive enhancements that would minimize the learning curve and foster acceptance in clinical settings. In this context, one possible visualization avenue to be studied further would be the implementation of alternative renditions of stained biomarkers as dynamic transparent overlays upon certain histological components of the original whole slide images. As for histological stain separation, further development is required for directly separating stains where the respective colors have a close distance in the hue values (angular position on a color space coordinate diagram), e.g. red and pink. Highly multiplexed images - leveraging multiplexed fluorescence, DNA, RNA, and isotope labeling \cite{Lewis2021} - provide fertile ground for such experimentation. Highly multiplexed imaging enables the visualization of numerous biomarkers within a single tissue sample, providing unprecedented insights into cellular and molecular heterogeneity \cite{Valous2022}. For example, Ref. \cite{Radtke2022} provides such high-quality multiplexed imaging data with the corresponding ground truth images; the latter are essential for validating any proposed method. Regarding machine/deep learning, results showed that the proposed transformations enhanced model performance by changing the visual representation of the input images. This can be useful if a particular type of dataset (e.g. H\&E) is utilized for which optimal transformations are known/computed a priori (e.g. like those shown in this work). In general though, rather than relying on fixed transformations, a more dynamic approach would be to integrate learnable transformations within the pipeline. For example, models can be trained with a neural architecture that includes layers or modules dedicated to learning the most effective transformations. This ensures that the model can automatically adjust during training, thereby improving its ability to generalize across different types of images.
\parskip=5pt

\noindent
\textbf{Concluding remarks}. This work leveraged quaternions and the 2D orthogonal planes split framework for a variety of natural and biomedical image analysis workflows. The splitting of a quaternion (representing a pixel) into pairs of orthogonal two-dimensional planes has demonstrated significant effectiveness in producing insightful results across different image types. Primarily, it demonstrated notable versatility, making it a valuable tool in computer vision and biomedical applications. Furthermore, the proposed non-data-driven methods achieved comparable or better results (particularly in cases involving well-known or widely used methods, e.g. in de-colorization and stain separation) to those reported in the literature, showcasing the potential of robust theoretical frameworks with practical effectiveness, even in the absence of large-scale data. This can enable an improved understanding of how outcomes are derived, making the proposed approaches particularly valuable in contexts where interpretability is significant. Working in the hypercomplex setting will continue to drive innovation and provide effective computational tools for a broad range of applications in image-driven fields.

\section*{\large\uppercase{Materials and Methods}}
In this section, all essential definitions and descriptions are provided to facilitate a clear understanding of the subject matter. Readers should check the relevant literature references to explore specific aspects in greater detail.

\subsection*{\normalsize{Tissue preparation}}
Microscope slides were imaged in brightfield mode at 20$\times$ and 40$\times$ magnifications using the Aperio AT2 (Leica Biosystems, Germany) and VS200 (Evident, Japan) whole slide scanners. The scanners automatically detected the regions of interest that contain tissue and also determined a valid focal plane for scanning.

For the hematoxylin and eosin (H\&E) stain shown in Fig. \ref{fig:img_contr}(e)/Fig. \ref{fig:img_recol_bio}(a)/Fig. \ref{fig:img_unsupstainsep_bio}(a)/Fig. \ref{fig:img_supstainsep_bio}(a)/Fig. \ref{fig:img_color_vis}(a), the sample originated from human ovarian adenocarcinoma \cite{SuarezCarmona2019}. After surgical resection, the tissue was formalin-fixed and paraffin-embedded; the tissue was sectioned into 3 $\mu$m thick slices and was placed on SuperFrost Plus Slides (Thermo Fisher Scientific, USA) before the H\&E stain \cite{SuarezCarmona2019}.

The H\&E stains shown in Fig. \ref{fig:img_restain_bio}(a-b)/Fig. \ref{fig:img_large_scale_vis_bio}(a) along with those utilized for Fig. \ref{fig:time_compl}, derived from early lobular breast carcinoma cases (human) treated by surgical resection. The tissue was formalin-fixed and paraffin-embedded, subsequently cut into 4 $\mu$m sections, and mounted on SuperFrost Plus Slides (Thermo Fisher Scientific, USA); sections were stained for H\&E according to standard protocols.

For the single immunostain shown in Fig. \ref{fig:img_contr}(f)/Fig. \ref{fig:img_recol_bio}(b)/Fig. \ref{fig:img_restain_bio}(c)/Fig. \ref{fig:img_unsupstainsep_bio}(b)/Fig. \ref{fig:img_supstainsep_bio}(b), primary pancreatic neuroendocrine neoplasms (human) were available as formalin‐fixed paraffin‐embedded tissue \cite{Valous2016}. Immunohistochemical (IHC) staining for Ki‐67 (murine monoclonal antibody MIB‐1, dilution 1:100) was performed on sections (5 $\mu$m thick) taken from original tissue blocks using the avidin‐biotin‐peroxidase detection system on a fully automated staining facility (DAKO, Germany) \cite{Valous2016}.

For the single immunostain shown in Fig. \ref{fig:img_contr}(g)/Fig. \ref{fig:img_recol_bio}(c)/Fig. \ref{fig:img_unsupstainsep_bio}(c)/Fig. \ref{fig:img_supstainsep_bio}(c), EpCAM IHC was performed on a human liver metastasis cryosection using Vina Green for visualization \cite{Berthel2019}. A 5 $\mu$m thick cryosection was prepared and incubated overnight with a biotin-murine-anti-EpCAM antibody (dilution 1:50) (BioLegend, USA) \cite{Berthel2019}. Then, the section was incubated in HRP Streptavidin (dilution 1:500) followed by 10 min in Vina Green solution (Biocare Medical, USA) \cite{Berthel2019}.

For the single immunostain shown in Fig. \ref{fig:img_contr}(h)/Fig. \ref{fig:img_recol_bio}(d)/Fig. \ref{fig:img_unsupstainsep_bio}(d)/Fig. \ref{fig:img_supstainsep_bio}(d), EpCAM IHC was performed on a human liver metastasis cryosection using AP-Red for visualization. A 5 $\mu$m thick cryosection was prepared and stained with a rabbit-anti-EpCAM antibody (dilution 1:2000) (Cell Signaling, Germany) using the Bond Polymer Refine Red Detection system (Leica Biosystems, Germany).

For the double immunostain shown in Fig. \ref{fig:img_recol_bio}(e)/Fig. \ref{fig:img_restain_bio}(d)/Fig. \ref{fig:img_supstainsep_bio}(e), the sample originated from human ovarian adenocarcinoma \cite{SuarezCarmona2019}. After surgical resection, the tissue was formalin-fixed and paraffin-embedded; the tissue was sectioned into 3 $\mu$m thick slices and was placed on SuperFrost Plus Slides (Thermo Fisher Scientific, USA) \cite{SuarezCarmona2019}. IHC for CD3 (clone Sp7) and Ki-67 (clone MIB-1) were performed on BOND-MAX (Leica Biosystems, Germany) using the BOND Polymer Refine Red Detection kit and the BOND Polymer Refine Detection kit, respectively \cite{SuarezCarmona2019}.

For the single immunostain shown in Fig. \ref{fig:img_large_scale_vis_bio}(b), a specimen from human colorectal cancer liver metastasis was available as formalin-fixed paraffin-embedded (FFPE) tissue. 3 $\mu$m thick sections were stained for CD3 (clone Sp7) on the BOND-MAX (Leica Biosystems, Germany) automated with the DAB Bond Polymer Refine Detection kit (Leica Biosystems, Germany) after appropriate epitope retrieval.

For the single immunostain shown in Fig. \ref{fig:img_large_scale_vis_bio}(c), a specimen from human epithelial ovarian carcinoma was available as formalin-fixed paraffin-embedded (FFPE) tissue; this was sectioned in RNase-free conditions. Sections (6 $\mu$m thick) were used within 24 hours of sectioning for in situ hybridization with the RNAscope 2.5 HD Assay-RED system with fast red detection (Advanced Cell Diagnostics, USA) following the manufacturer’s instructions. The RNAscope Probe - Hs-NLRP3 (Advanced Cell Diagnostics, USA) was utilized, including a positive (RNAscope Probe - Hs-PPIB-sense) and negative control (RNAscope 2.5 LS Negative Control Probe DapB). The slide was counterstained with hematoxylin and embedded in Aquatex mounting medium.

\begin{figure}[t]
\centering
\includegraphics[width=1.0\linewidth]{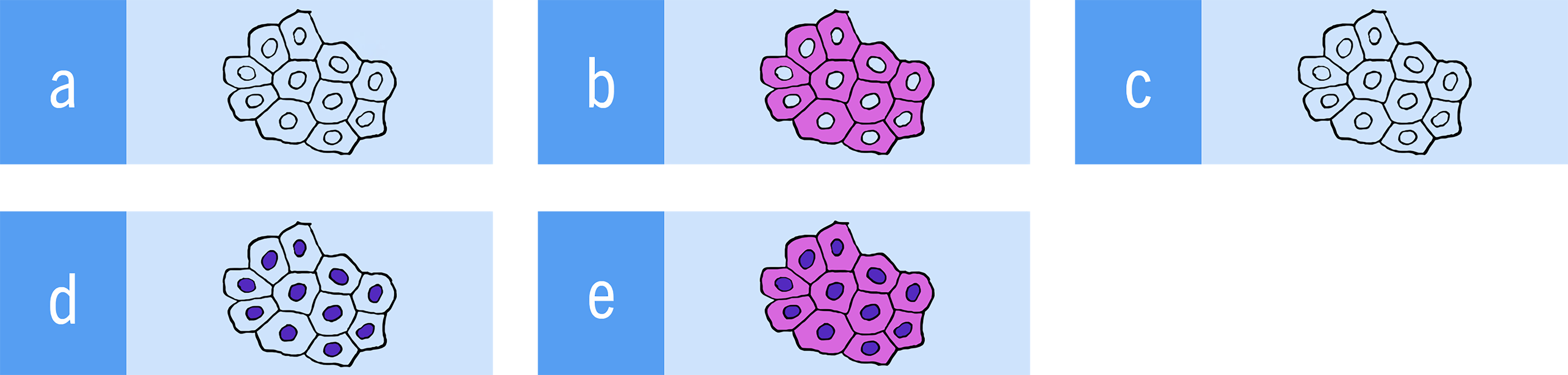}
\vspace{-12pt}
\caption{\footnotesize{Lab protocol schematic utilized for computational stain separation comparisons (H\&E stains of Tables \ref{tab:Hstain_sep_comp}-\ref{tab:Estain_sep_comp} and Fig. \ref{fig:img_stainsep_bio}). More specifically: a) an unstained tissue section was initially scanned, b) the eosin stain (Eosin Y; Sigma-Aldrich, USA) was performed and the slide was re-scanned, c) the sample was re-hydrated using ethanol baths (100\%, 96\%, 90\%, and 70\%) followed by deionized water, then dehydrated through an ethanol bath (from 70\% to 100\%) and xylol, and finally mounted before re-scanning the slide, d) re-hydration was carried out, followed by hematoxylin staining (Shandon Gill Hematoxylin 3; Thermo Fisher Scientific, USA), dehydration, and slide re-scanning, and finally e) the eosin stain was performed (as described in the previous step) thus obtaining the H\&E slide which was scanned as well.}}
\label{fig:h_e_he_protocol}
\end{figure}

For the H\&E stains utilized in Tables \ref{tab:Hstain_sep_comp}-\ref{tab:Estain_sep_comp} and Fig. \ref{fig:img_stainsep_bio}, the samples originated from primary pancreatic neuroendocrine neoplasms (human). The tissues were formalin-fixed and paraffin-embedded, subsequently cut into 3 $\mu$m sections, and mounted on SuperFrost Plus Slides (Thermo Fisher Scientific, USA). The lab protocol \cite{Tadrous2010, McCann2014} is visualized schematically in Fig. \ref{fig:h_e_he_protocol} (the unstained and de-stained whole slides were utilized as controls). The protocol was applied to three tissue sections, consequently, hematoxylin only (H-only), eosin only (E-only), and H\&E whole slide images were obtained.

\subsection*{\normalsize{Software, computing libraries, and code}}
The software package MATLAB R2023 (MathWorks, USA) was utilized for prototyping and visualization with no additional third-party commercial or open-source toolboxes. Mathematica v11 (Wolfram Research, USA) was used for the visualization of mathematical objects. Data plotting was performed using Origin (OriginLab, USA). Figure editing and visualization were performed using Adobe Photoshop (Adobe, USA). High-performance CPU and GPU computing libraries were developed in Python that were able to handle entire images as well as image tiles. Specifically, for this library: i) OpenCV \cite{opencv_library} and Pillow \cite{clark2015pillow} were used for reading and writing image data, ii) NumPy \cite{Harris2020} and NumPy-quaternion were used for working with arrays, and iii) a CUDA \cite{cuda} kernel was designed with Numba \cite{lam2015numba} for the Hamilton product. Codes ran on an Intel Xeon W-2265 central processing unit, an NVIDIA RTX A4000, and two NVIDIA Quadro P6000 graphics processing units. Python scripts, Jupyter notebooks, MATLAB code, and test images are publicly available in the following Figshare repository: \url{https://doi.org/10.6084/m9.figshare.29683382}. The code demonstrates functionalities such as re-colorization, de-colorization, contrast enhancement, re-staining, and stain separation. The code offers insight into the workflows and their resulting outputs; it demonstrates the methodologies presented in this work and helps to illustrate key implementation concepts. While the code is functional, it is designed primarily for understanding and exploration rather than direct application in production environments. The output files generated by the code include results from both Python and MATLAB implementations; these output images are provided as validation, demonstrating that both implementations produce matching results.

\subsection*{\normalsize{Datasets}}
The public McGill color image database was utilized for demonstrating re-/de-colorization, comprising a large number of natural scenes \cite{Olmos2004}. Regarding Fig. \ref{fig:img_recol}(d), the image originated from the open-access artworks collection of the Metropolitan Museum of Art, New York: \textit{Cider Making} by William Sidney Mount, 1840–1841 \url{(https://www.metmuseum.org/art/collection/search/11619}. Regarding Fig. \ref{fig:img_color_vis}(b), the image originated from the public outdoor scene dataset \cite{Wang2018}.

The public COLOR250 dataset (250 images, $260^2$ to $1024^2$ pixels) was introduced by Ref. \cite{Lu2014} for evaluating de-colorization performance; it contains natural images as well as digital charts, logos, and illustrations \cite{Lu2014}. 

The public CEED2016 dataset (d-ce1) comprising 30 images ($512^2$ pixels each) was introduced by Refs. \cite{Qureshi2016, Qureshi2017} for contrast enhancement evaluation. The authors selected images with different textures, color distributions, and contrast variations using three quantitative measures \cite{Qureshi2017}. The public COVID-LDCT dataset (d-ce2) contains chest computed tomography (CT) images using low-dose and ultra-low-dose scan protocols that reduce radiation exposure close to that of a single X-ray \cite{Afshar2021}; in this work, the extra set of 100 positive COVID-19 patients was utilized for contrast enhancement evaluation comprising 14687 DICOM images ($512^2$ pixels) \cite{Afshar2021}.

The in-house generated dataset (d-he1, d-he2, and d-he3) for stain separation comparisons comprised three H-only, three E-only, and three H\&E whole slide images (24-bit). Large sections were cropped from the whole slide images (respective sizes (width$\times$height): $9287\times3642$, $9319\times3675$, and $9312\times3653$ pixels) followed by image registration of the H-only and E-only image sections to the H\&E image sections (designated as reference). In this way, each H-only and E-only image section was aligned to the corresponding H\&E image section. In addition, 8-bit versions of these image sections were obtained, hence the 8-bit H-only and E-only image sections were utilized as lab ground truth for comparisons with the tested images.

A public biomedical image dataset was utilized for assessing performance gains pertinent to machine learning workflows. This balanced dataset (d-ml) is a collection of 5000 fully anonymized histological images (H\&E staining, 0.495 $\mu$m/pixel, 20$\times$ magnification) of formalin-fixed paraffin-embedded human colorectal adenocarcinomas, and contains 8 classes:  tumor epithelium, simple stroma (homogeneous composition including tumor stroma, extra-tumoral stroma, and smooth muscle), complex stroma (containing single tumor cells and/or few immune cells), immune cells (including immune-cell conglomerates and sub-mucosal lymphoid follicles), debris (including necrosis, hemorrhage, and mucus), normal mucosal glands, adipose tissue, and background (no tissue); each class comprises 625 images ($150^2$ pixels, 74$\times$74 $\mu$m) \cite{Kather2016}.

\subsection*{\normalsize{Performance metrics}}
Regarding the de-colorization metrics, the color-to-gray structural similarity metric (C2G-SSIM) was introduced by Ref. \cite{Ma2015}; it evaluates luminance, contrast, and structural similarities between a reference color image and the corresponding grayscale version. The authors showed that C2G-SSIM has a close agreement with subjective rankings and outperforms existing objective quality metrics for color-to-gray conversion \cite{Ma2015}.

Regarding the contrast enhancement metrics, the full-reference \cite{Wang2006} visual information fidelity (VIF) metric quantifies the loss of original image information due to processing or transmission; it can be used as a quality metric for both degraded and enhanced images \cite{Sheikh2006, Qureshi2017}. The reduced-reference \cite{Wang2011} image quality metric for contrast change (RIQMC) quantifies image contrast and naturalness; it combines the entropy of the phase congruency image and four statistical features computed from the image histogram \cite{Gu2016, Qureshi2017}. The lightness order error (LOE) metric measures naturalness preservation in the enhanced image \cite{Wang2013, Qureshi2017}; naturalness is defined as the degree of correspondence between the visual representation of the image and the knowledge of reality (colors of familiar objects) \cite{Qureshi2017}. The over-contrast measure (OCM) evaluates the amount of over-contrast in an image; it uses a local standard deviation map to detect uniform regions and employs a guided filter \cite{He2013} to measure the details in uniform regions \cite{Lee2019}.

Regarding the histological stain separation metrics, the full reference mean-squared error (MSE) measures the average squared difference between the tested and reference pixel values \cite{matlab_toolbox_ref}. The full reference peak signal-to-noise ratio (PSNR) derives from the MSE and indicates the ratio of the maximum pixel intensity to the power of the distortion \cite{matlab_toolbox_ref}. The full reference structural similarity (SSIM) index combines local image structure, luminance, and contrast into a single local quality score (structures are patterns of pixel intensities, especially among neighboring pixels, after normalizing for luminance and contrast) \cite{matlab_toolbox_ref}. The full reference multiscale structural similarity (MS-SSIM) index expands on the SSIM index by combining luminance information at the highest resolution level with structure and contrast information at several down-sampled resolutions \cite{matlab_toolbox_ref}.

\subsection*{\normalsize{Quaternion algebra}}
Quaternions were first described by W. R. Hamilton \cite{Grant2015} and their algebra is denoted by $\mathbb{H}$ \cite{BayroCorrochano2019}. An introduction to quaternionic calculus can be found in Ref. \cite{Morais2014}. Quaternions are vectors in four dimensions endowed with a rule for multiplication that is associative but not commutative \cite{Goldman2011}. Complex numbers are low-dimensional analogs of quaternions; many of the fundamental algebraic and geometric properties of quaternions appear in the complex numbers as well \cite{Goldman2011}. By defining, in addition to the complex number $i$, two additional imaginary units $j$ and $k$ such that the elements $1$, $i$, $j$, $k$ are linearly independent over the real numbers ($\mathbb{R}$), allows us to form a four-dimensional vector space \cite{Kramer2017}: $\mathbb{H}:=\{ q = a \cdot 1 + b\cdot i + c\cdot j + d \cdot k \quad | \quad a, b, c, d \in \mathbb{R}\}$. Essentially, the algebra $\mathbb{H}$ of real quaternions is a four-dimensional (4D) non-commutative algebra over the real number field $\mathbb{R}$ with canonical basis 1, $i$, $j$, $k$ satisfying the conditions ($\oast$: Hamilton product) \cite{Morais2014}:
    \begin{equation} \label{eq:definitions}
    \begin{aligned}
        &{i^2} = {j^2} = {k^2} = i \oast j \oast k =  - 1\\
        &i \oast j = - j \oast i = k\\
        &j \oast k = - k \oast j = i\\
        &k \oast i = - i \oast k = j
    \end{aligned}
    \end{equation}
    
\noindent
A real quaternion is often represented in the following form \cite{Morais2014}:
    \begin{equation} \label{eq:representation}
        q = {q_r} + {q_i}i + {q_j}j + {q_k}k
    \end{equation}
    
\noindent
where $q_{r}$, $q_{i}$, $q_{j}$, and $q_{k}$ $\in$ $\mathbb{R}$. From the previous equation, $q_{r}$ is called the real or scalar part $S(q)$, and $q_{i}i$ + $q_{j}j$ + $q_{k}k$ is called the vector part $V(q)$. The conjugate of a real quaternion leaves the scalar part $q_{r}$ unchanged \cite{Morais2014}:
    \begin{equation} \label{eq:conjugate}
        \overline q  = {q_r} - \left( {{q_i}i + {q_j}j + {q_k}k} \right)
    \end{equation}
    
\noindent
This leads to the norm $\in$ $\mathbb{R}$ of a real quaternion which is given by \cite{Morais2014}:
    \begin{equation} \label{eq:norm}
        \left| q \right| = \sqrt {q_r^2 + q_i^2 + q_j^2 + q_k^2}
    \end{equation}
    
\noindent
A pure quaternion $\boldsymbol{q}$ (with the real part being zero; $q_{r}=0$) is defined as \cite{Morais2014}:
    \begin{equation} \label{eq:purequaternion}
        {\boldsymbol{q}} = V(q) = {q_i}i + {q_j}j + {q_k}k
    \end{equation}
    
\noindent
Dividing a non-zero quaternion $q$ by its norm ($q/\left|q\right|$) produces a unit quaternion; thus, when the norm $|q| = 1$ and the real part is zero ($q_{r}=0$) then $\boldsymbol{q_{u}}$ is called a pure unit quaternion. The left and right inverse of a non-zero quaternion is defined as \cite{Morais2014}:
    \begin{equation} \label{eq:inverse}
        {q^{ - 1}} = {{{\overline q } \mathord{\left/ {\vphantom {{\overline q } {\left| q \right|}}} \right. \kern-\nulldelimiterspace} {\left| q \right|}}^2}
    \end{equation}

\noindent
Let $q$ be a real quaternion such that $q \neq (0,0,0,0)$ and $\boldsymbol{q}$ its pure quaternion form, then $q$ is also associated with an angle $\theta$ via \cite{Morais2014}:
    \begin{equation} \label{eq:angle}
    \begin{aligned}
        &cos \theta = \frac{q_r}{\left| q \right|} \\
        &sin \theta = \frac{\left| \boldsymbol{q} \right|}{\left| q \right|}
    \end{aligned}
    \end{equation}

\noindent
Thus, the polar representation of every non-zero quaternion $q$ is given by \cite{Morais2014}:
    \begin{equation} \label{eq:polarrepresentation}
        q = \left| q \right| \left( cos \theta + \frac{\boldsymbol{q}}{\left| \boldsymbol{q} \right|} sin \theta \right)
    \end{equation}

\noindent
If $p$, $q$ $\in$ $\mathbb{H}$ with $p = a_{1} + b_{1}i + c_{1}j + d_{1}k$ and $q = a_{2} + b_{2}i + c_{2}j + d_{2}k$, then addition and subtraction are defined as follows \cite{Morais2014}:
    \begin{equation} \label{eq:additionsubtraction}
    \begin{aligned}
        &p + q = \left( {a_{1} + a_{2}} \right) + \left( {b_{1} + b_{2}} \right)i + \left( {c_{1} + c_{2}} \right)j + \left( {d_{1} + d_{2}} \right)k\\
        &p - q = \left( {a_{1} - a_{2}} \right) + \left( {b_{1} - b_{2}} \right)i + \left( {c_{1} - c_{2}} \right)j + \left( {d_{1} - d_{2}} \right)k
    \end{aligned}
    \end{equation}

\noindent
The quaternion multiplication of $p$ and $q$ ($\oast$: Hamilton product) is given by \cite{Morais2014}:
    \begin{equation} \label{eq:quaternionmultiplication}
    \begin{aligned}
        p \oast q = &\left( {{a_1}{a_2} - {b_1}{b_2} - {c_1}{c_2} - {d_1}{d_2}} \right) +\\
        &+ \left( {{a_1}{b_2} + {b_1}{a_2} + {c_1}{d_2} - {d_1}{c_2}} \right)i +\\
        &+\left( {{a_1}{c_2} - {b_1}{d_2} + {c_1}{a_2} + {d_1}{b_2}} \right)j +\\
        &+\left( {{a_1}{d_2} + {b_1}{c_2} - {c_1}{b_2} + {d_1}{a_2}} \right)k
    \end{aligned}
    \end{equation}
    
\noindent
Moreover, real multiplication with $\lambda$ $\in$ $\mathbb{R}$ is given by \cite{Morais2014}:
    \begin{equation} \label{eq:realmultiplication}
        \lambda p = \left( {\lambda a{_1}} \right) + \left( {\lambda b{_1}} \right)i + \left( {\lambda c{_1}} \right)j + \left( {\lambda d{_1}} \right)k
    \end{equation}

\noindent
The dot product of $p$, $q$ $\in$ $\mathbb{H}$ is a real number ($\mathbb{R}$), defined as the sum of each coefficient of $p$ multiplied by the corresponding coefficient of $q$ \cite{Morais2014}, namely:
    \begin{equation} \label{eq:dotproduct}
    \begin{aligned}
        &\langle p,q \rangle = p \cdot q = {{a_1}{a_2} + {b_1}{b_2} + {c_1}{c_2} + {d_1}{d_2}} \\
    \end{aligned}
    \end{equation}

\noindent
Given $q = {q_r} + {q_i}i + {q_j}j + {q_k}k$ and its pure quaternion form $\boldsymbol{q} = {q_i}i + {q_j}j + {q_k}k$, then the exponential is computed as \cite{Morais2014, 1711.02508}:
    \begin{equation} \label{eq:exponential}
        e^{q} = e^{q_r} \left(cos\left| \boldsymbol{q} \right| + \frac{\boldsymbol{q}}{\left| \boldsymbol{q} \right|} sin\left| \boldsymbol{q} \right| \right)
    \end{equation}

\noindent
Furthermore, the quaternion natural logarithm function is given by \cite{Morais2014, 1711.02508}:
    \begin{equation} \label{eq:logarithm}
        ln(q) = ln \left| q \right|+\frac{arccos \left( \frac{q_{r}}{\left| q \right|} \right)}{\left| \boldsymbol{q} \right|}
        \boldsymbol{q}
    \end{equation}
    
\noindent
For a real quaternion $q$ $\in$ $\mathbb{H}$ and $n$ $\in$ $\mathbb{Z}$, the quaternion power function $q^n$ is defined as \cite{Morais2014}:
    \begin{equation} \label{eq:power1}
        q^n = e^{ln(q)n}
    \end{equation}

\noindent
Specifically, when $n$ $\in$ $\mathbb{R}$ and taking into account the polar form of a quaternion, then the quaternion power function is defined as \cite{Morais2014}:
    \begin{equation} \label{eq:power2}
        q^n = \left| q \right|^n e^{\frac{\boldsymbol{q}}{\left| \boldsymbol{q} \right|} n \theta} = 
        \left| q \right|^n \left(cos(n \theta) + \frac{\boldsymbol{q}}{\left| \boldsymbol{q} \right|} sin(n \theta) \right)
    \end{equation}

\noindent
If $A$ and $B$ are two quaternion matrices of same size and $\lambda$ $\in$ $\mathbb{R}$, then the sum or difference ($A + B$ or $A-B$) of $A$ and $B$ is computed by adding or subtracting entry-wise \cite{Morais2014}. Furthermore, the multiplication $\lambda A$ is given by multiplying every entry of A on the left-hand side by $\lambda$ \cite{Morais2014}.

\begin{figure}[t]
\centering
\includegraphics[width=1.0\linewidth]{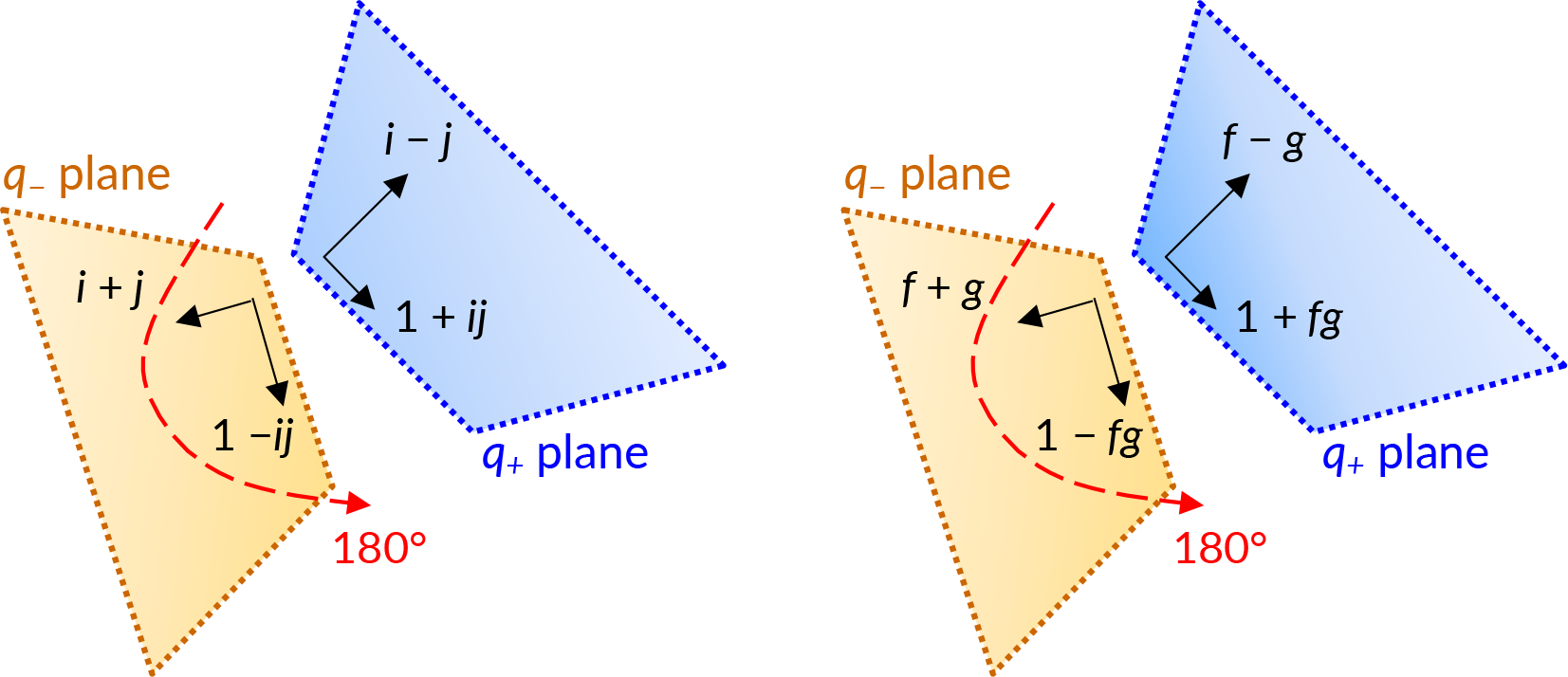}
\vspace{-12pt}
\caption{\footnotesize{Geometric properties of the maps $i()j$ (left) and $\boldsymbol{f}()\boldsymbol{g}$ (right) as half-turns. The map $i()j$ rotates the $q_{-}$ plane by $\pi$ around the 2D $q_{+}$ axis plane (left). Similarly for the $\boldsymbol{f}()\boldsymbol{g}$ map (right). Adapted from Refs. \cite{HitzerSangwine2013, Hitzer2015}.}}
\label{fig:ops_geo}
\end{figure}

\subsection*{\normalsize{2D orthogonal planes split}}
A thorough mathematical investigation of the 2D orthogonal planes split (or $\pm$ split) with definitions and proofs can be found in Refs. \cite{HitzerSangwine2013, Hitzer2015}. The split of quaternions is motivated by the consistent appearance of two terms in the quaternion Fourier transform \cite{HitzerSangwine2013, Hitzer2015}:
    \begin{equation} \label{eq:Fourier}
        F\{f\}(u,v) = \int_{\mathbb{R}^2} e^{-ixu} f(x,y) e^{-jyv} \,dxdy
    \end{equation}

\noindent
This observation ($i$ is on the left and $j$ is on the right) and that every quaternion can be rewritten as \cite{HitzerSangwine2013, Hitzer2015}:
    \begin{equation} \label{eq:rewrite}
        q = {q_r} + {q_i}i + {q_j}j + {q_k}k = {q_r} + {q_i}i + {q_j}j + {q_k}ij
    \end{equation}

\noindent
motivated the quaternion split with respect to the pair of orthonormal pure unit quaternions $i$, $j$:
    \begin{equation} \label{eq:opsij}
    \begin{aligned}
        &q = q_{+}+q_{-} \\
        &q_{\pm}^{\left(i,j\right)} = \frac{1}{2} \left(q \pm i \oast q \oast j \right)
    \end{aligned}
    \end{equation}

\noindent
where $\oast$ is the quaternion multiplication operator. Replacing e.g. $i$ with $j$, and $j$ with $k$ throughout would merely change the notation but not the fundamental observation \cite{HitzerSangwine2013, Hitzer2015}. This quaternion split is called the two-dimensional (2D) orthogonal planes split (OPS) \cite{HitzerSangwine2013, Hitzer2015}. Given any two quaternions $p$, $q$ $\in$ $\mathbb{H}$ and applying the 2D orthogonal planes split of Eq. \ref{eq:opsij}, then the resulting two parts are orthogonal, i.e. $p_{+}$ $\perp$ $q_{-}$ and $p_{-}$ $\perp$ $q_{+}$ \cite{HitzerSangwine2013, Hitzer2015}. The map $i()j$ leads to an adapted orthogonal basis of $\mathbb{H}$: $i \otimes q \otimes j = q_{+} - q_{-}$ \cite{HitzerSangwine2013, Hitzer2015}. Under the map $i()j$, the $q_{+}$ part is invariant but the $q_{-}$ part changes sign \cite{HitzerSangwine2013, Hitzer2015}. Both parts are 2D and they span two completely orthogonal planes, hence the name orthogonal planes split \cite{HitzerSangwine2013, Hitzer2015}. The $q_{+}$ and $q_{-}$ plane has the orthogonal quaternion basis \cite{HitzerSangwine2013, Hitzer2015}:
    \begin{equation} \label{eq:basisij}
    \begin{aligned}
        &q_{+} = \{i-j,1+ij \}\\
        &q_{-} = \{i+j,1-ij \}
    \end{aligned}
    \end{equation}

\noindent
Now, assume an arbitrary pair of pure unit quaternions $\boldsymbol{f}$ and $\boldsymbol{g}$ with $\boldsymbol{f}^2 = \boldsymbol{g}^2 = -1$. The orthogonal 2D planes split is then defined with respect to any two pure unit quaternions $\boldsymbol{f}$ and $\boldsymbol{g}$ as \cite{HitzerSangwine2013, Hitzer2015}:
    \begin{equation} \label{eq:opsfg}
    \begin{aligned}
        &q_{\pm}^{\left(f,g\right)} = \frac{1}{2} \left(q \pm \boldsymbol{f} \oast q \oast \boldsymbol{g} \right)\\
        &\boldsymbol{f} \oast q \oast \boldsymbol{g} = q_{+}-q_{-}
    \end{aligned}
    \end{equation}

\noindent
where $\oast$ is the quaternion multiplication operator. Similarly, under the map $\boldsymbol{f}()\boldsymbol{g}$ the $q_{+}$ part is invariant, but the $q_{-}$ part changes sign; both parts are 2D and span two completely orthogonal planes \cite{HitzerSangwine2013, Hitzer2015}. For $\boldsymbol{f} \neq \pm \boldsymbol{g}$, the $q_{+}$ and $q_{-}$ plane is spanned by two orthogonal quaternions, respectively \cite{HitzerSangwine2013, Hitzer2015}:
    \begin{equation} \label{eq:basisfg}
    \begin{aligned}
        &q_{+} = \{\boldsymbol{f}-\boldsymbol{g},1+\boldsymbol{f}\boldsymbol{g} \}\\
        &q_{-} = \{\boldsymbol{f}+\boldsymbol{g},1-\boldsymbol{f}\boldsymbol{g} \}
    \end{aligned}
    \end{equation}

\noindent
Given any two quaternions $p$ and $q$ and applying the 2D orthogonal planes split with respect to any two pure unit quaternions $\boldsymbol{f}$ and $\boldsymbol{g}$ we get zero for the scalar part of the mixed products \cite{HitzerSangwine2013, Hitzer2015}. A geometric picture of the split is obtained in Fig. \ref{fig:ops_geo} (adapted from Refs. \cite{HitzerSangwine2013, Hitzer2015}). The map $i()j$ rotates the $q_{-}$ plane by $\pi$ around the 2D $q_{+}$ axis plane \cite{HitzerSangwine2013, Hitzer2015}. This interpretation of the map $i()j$ is in agreement with Coxeter's notion of half-turn \cite{HitzerSangwine2013, Hitzer2015}. In agreement with the geometric interpretation, the map $i()j$ is an involution because applying it twice leads to identity \cite{HitzerSangwine2013, Hitzer2015}:
    \begin{equation} \label{eq:involution}
        i(iqj)j=i^2qj^2=(-1)^2q=q
    \end{equation}

\noindent
Similarly, the map $\boldsymbol{f}()\boldsymbol{g}$ rotates the $q_{-}$ plane by $\pi$ around the 2D $q_{+}$ axis plane (Fig. \ref{fig:ops_geo}), which is in agreement with Coxeter's notion of half-turn \cite{HitzerSangwine2013, Hitzer2015}. The pair of pure unit quaternions $\boldsymbol{f}, \boldsymbol{g}$ $\in$ $\mathbb{H}$ used in the general OPS (Eq. \ref{eq:opsfg}) leads to an explicit basis for the resulting two orthogonal 2D planes ($q_{+}$ plane and $q_{-}$ plane) \cite{HitzerSangwine2013}. Theorem 3.5 from Ref. \cite{HitzerSangwine2013} dictates how to determine from a given steerable pair of orthogonal 2D planes $\in$ $\mathbb{H}$, the pair of pure unit quaternions $\boldsymbol{f},\boldsymbol{g}$ $\in$ $\mathbb{H}$ which splits $\mathbb{H}$ exactly into this given pair of orthogonal 2D planes \cite{HitzerSangwine2013}.

\begin{figure}[t]
\centering
\includegraphics[width=1.0\linewidth]{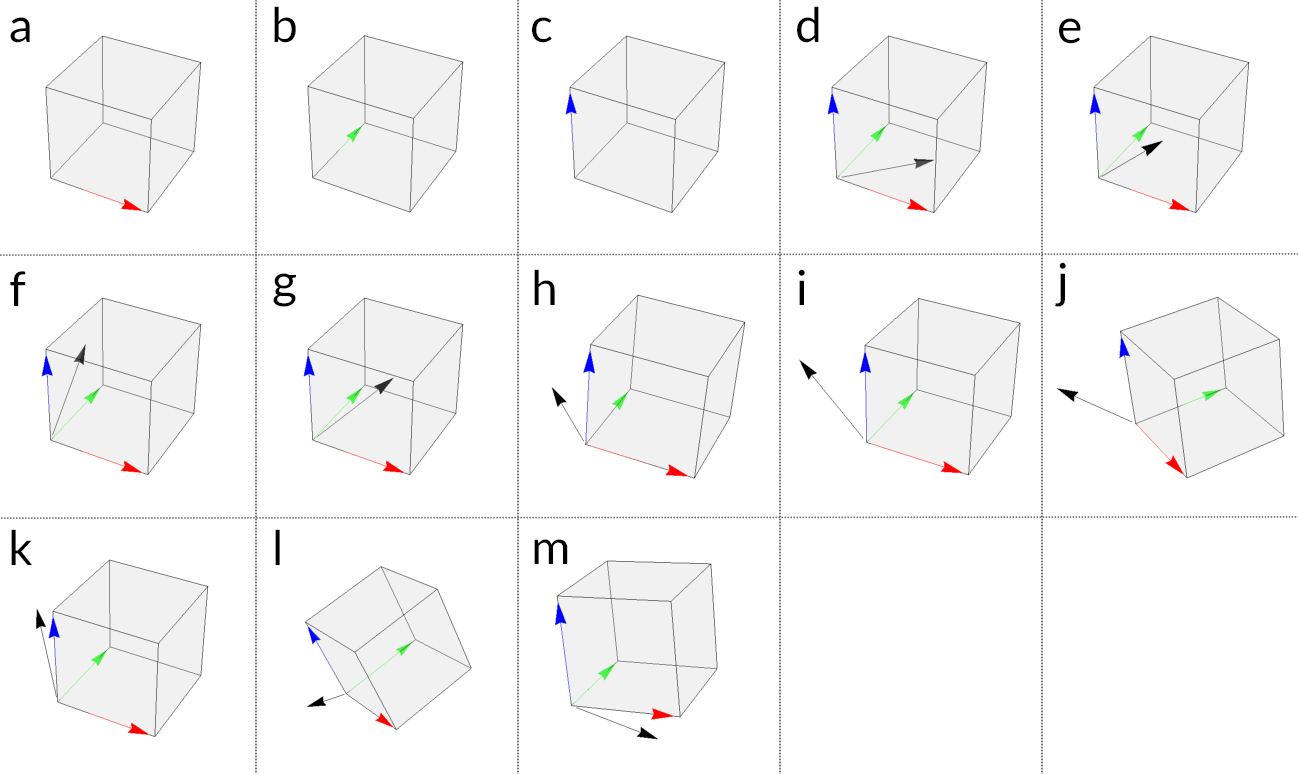}
\vspace{-12pt}
\caption{\footnotesize{Pure unit quaternions $\boldsymbol{\mu_{1}}$ to $\boldsymbol{\mu_{13}}$ utilized in this study: a) $\boldsymbol{\mu_{1}}=[i]$, b) $\boldsymbol{\mu_{2}}=[j]$, c) $\boldsymbol{\mu_{3}}=[k]$, d) $\boldsymbol{\mu_{4}}=[(i+j)/\sqrt{2}]$, e) $\boldsymbol{\mu_{5}}=[(i+k)/\sqrt{2}]$, f) $\boldsymbol{\mu_{6}}=[(j+k)/\sqrt{2}]$, g) $\boldsymbol{\mu_{7}}=[(i+j+k)/\sqrt{3}]$, h) $\boldsymbol{\mu_{8}}=[(-i+j)/\sqrt{2}]$, i) $\boldsymbol{\mu_{9}}=[(-i+k)/\sqrt{2}]$, j) $\boldsymbol{\mu_{10}}=[(-j+k)/\sqrt{2}]$, k) $\boldsymbol{\mu_{11}}=[(-i+j+k)/\sqrt{3}]$, l) $\boldsymbol{\mu_{12}}=[(i-j+k)/\sqrt{3}]$, and m) $\boldsymbol{\mu_{13}}=[(i+j-k)/\sqrt{3}]$.}}
\label{fig:m1m13}
\end{figure}

\subsection*{\normalsize{Quaternion matrix representation}}
A color image (24-bit, RGB) $I_C$ with color channels $cc1$ (red), $cc2$ (green), and $cc3$ (blue) having size $[n \times m]$ is a function $\{ I_C: F \longrightarrow [0,255]^3 \in \mathbb{R} \}$ where $F=[0;m-1] \times [0;n-1]$ are the pixels. A practical way for a color image, in the RGB color space, to be converted to a quaternion matrix is by placing the three color components (R, G, and B) into the three quaternionic imaginary parts ($i$, $j$, $k$) \cite{Moxey2003}. A pixel at image coordinates $(n, m)$ can be represented as $\boldsymbol{q(n,m)}=r(n,m)i+g(n,m)j+b(n,m)k$ where $r(n,m)$, $g(n,m)$, and $b(n,m)$ are the red, green, and blue components of the pixel, respectively. Thus, the image is represented as a matrix $\boldsymbol{I_{q}}$ whose elements are pure quaternions \cite{Ell2007a}.

\subsection*{\normalsize{Re-colorization}}
Results are shown in Figs. \ref{fig:img_recol}, \ref{fig:img_recol_bio}, and \ref{fig:img_large_scale_vis_bio}. Table \ref{tab:workflow_recol} provides an overview of the computational workflow. The 2D orthogonal planes split is carried out with respect to one or two pure unit quaternions $\boldsymbol{f}$ and $\boldsymbol{g}$:
    \begin{equation} \label{eq:recolorization}
    \begin{aligned}
        &q_{+}^{\left(f,f\right)} = \frac{1}{2} \left(\boldsymbol{I_{q}} + \boldsymbol{f} \oast \boldsymbol{I_{q}} \oast \boldsymbol{f} \right)\\
        &q_{-}^{\left(f,f\right)} = \frac{1}{2} \left(\boldsymbol{I_{q}} - \boldsymbol{f} \oast \boldsymbol{I_{q}} \oast \boldsymbol{f} \right)\\
        &q_{+}^{\left(f,g\right)} = \frac{1}{2} \left(\boldsymbol{I_{q}} + \boldsymbol{f} \oast \boldsymbol{I_{q}} \oast \boldsymbol{g} \right)\\
        &q_{-}^{\left(f,g\right)} = \frac{1}{2} \left(\boldsymbol{I_{q}} - \boldsymbol{f} \oast \boldsymbol{I_{q}} \oast \boldsymbol{g} \right)
    \end{aligned}
    \end{equation}

\noindent
where $\oast$ is the quaternion multiplication operator. Eq. \ref{eq:recolorization} contains multiplications of two quaternion numbers by a quaternion matrix $[n \times m]$, one quaternion matrix addition, and one real multiplication. In this work, the maps $\boldsymbol{f}()\boldsymbol{f}$ and $\boldsymbol{f}()\boldsymbol{g}$ are generalized to be: a) a combination of pure unit quaternions $\boldsymbol{\mu_{1}}$ to $\boldsymbol{\mu_{13}}$ that derive from pure quaternions, respectively: $(i), (j), (k), (i + j), (i + k), (j + k), (i + j + k), (-i + j), (-i + k), (-j + k), (-i + j + k), (i - j + k)$, and $(i + j - k)$ (Fig. \ref{fig:m1m13}), and b) a combination of values represented as pure unit quaternions that are determined by the user (e.g. pixel or area sampling), or computed by an automated method. The symbol $()$ refers to a color pixel represented as a pure quaternion. The pure unit quaternions $\boldsymbol{f}, \boldsymbol{g}$ can have geometric meaning in the RGB color space, e.g. $\boldsymbol{\mu_1}$, $\boldsymbol{\mu_2}$, and $\boldsymbol{\mu_3}$ refer to the red, green, and blue color axes respectively, while $\boldsymbol{\mu_7}$ is the direction corresponding to the luminance or gray-line axis \cite{Moxey2003, Ell2007a}. Color channel matrices $\in$ $\mathbb{R}$ are extracted from the resulting quaternion matrix of Eq. \ref{eq:recolorization}. Scalar parts are retained during computations but are discarded prior to image encoding. For image encoding, data normalization is carried out for each color channel in the range $[0,1]$. Thus, the transformed color image is obtained: $I_{T}$ (24-bit).

In certain cases, where some colors are grossly underrepresented in an image (e.g. sparse representation, faint appearance, etc.) or batch effects need to be eliminated for machine learning applications, an exemplar image (small or large depending on the application) is utilized. This image is either cropped from the original image to represent preferred color distributions or a completely new image is introduced that can optimally capture color variability. More specifically, data normalization is carried out for each color channel in the range $[0,1]$ on the exemplar image. The pairs comprising minimum (min) and maximum (max) values $[min,max]$ are computed very rapidly for the exemplar image and any desired combination of pure unit quaternions ($\boldsymbol{f}$, $\boldsymbol{g}$); these pre-computed normalization ranges are then utilized for the image encoding step of the transformed images. For example, in Fig. \ref{fig:img_large_scale_vis_bio}(c), data normalization was carried out for each color channel in the range $[0,1]$ on a small exemplar image ($50 \times 50$ pixels) that was cropped from the original image. The exemplar image was selectively cropped to include shades of blue and red present in the original image. The pair comprising minimum and maximum values were computed for the exemplar image ($\boldsymbol{f} = \boldsymbol{\mu_{4}}$) and then utilized for the encoding step of the transformed image Fig. \ref{fig:img_large_scale_vis_bio}(c1).

\begin{table}[t]
    \centering
    \caption{\raggedright{\textbf{Workflow 1}: Re-colorization}}
    \vspace*{-0.75mm}
    \scriptsize
    \renewcommand{\arraystretch}{0.98}
    \begin{tabular*}{\linewidth}{@{\extracolsep{3pt}}l l@{}}
        \toprule
        \textbf{Require}: & re-colorization of color image $I_C$ \\
        \textbf{Input}: & color image (24-bit, RGB) $I_C$ \\
        \textbf{Output}: & transformed color image (24-bit, RGB) $I_T$ \\
        \noalign{\vskip 0.5mm}
        \hline
        \noalign{\vskip 0.5mm}
        1: & Quaternion representation as $\boldsymbol{I_q}$ \\
        2: & 2D orthogonal planes split according to Eq. \ref{eq:recolorization} \\
        3: & Extraction of color channel matrices \\
        4: & Data normalization for each color channel in $[0,1]$ \\
        5: & Concatenation of transformed color channels \\
        \bottomrule
    \end{tabular*}
\label{tab:workflow_recol}
\end{table}

\subsection*{\normalsize{Re-staining}}
Results are shown in Figs. \ref{fig:img_restain_bio} and \ref{fig:img_large_scale_vis_bio}. Tables \ref{tab:workflow_restain1} and \ref{tab:workflow_restain2} provide overviews of the computational workflows. For histological images with two dominant colors (e.g. H\&E or single immunostain with counterstain, Fig. \ref{fig:img_restain_bio}(a-c)), let $\boldsymbol{I_{q}}$ be the pure quaternionic matrix representation of $I_C$ with size $[n \times m]$. An image pre-processing step is carried out; the input image $\boldsymbol{I_{q}}$ is transformed to $TI_{q}$ using the following equation (component-wise multiplications):
    \begin{equation} \label{eq:transform}
        TI_{q} = ln(\boldsymbol{I_{q}}) e^{\boldsymbol{I_{q}}}
    \end{equation}

\noindent
Color channel matrices $\in$ $\mathbb{R}$ are extracted from the resulting quaternion matrix of Eq. \ref{eq:transform}. To obtain the output, scalar parts are retained during computations but are discarded prior to image encoding. For image encoding, data normalization is carried out for all color channels jointly in the range $[0,1]$. The resulting image is transformed to a quaternion matrix $\boldsymbol{TI_{nq}}$, and then the 2D orthogonal planes split is carried out with respect to a pure unit quaternion $\boldsymbol{f}$ (with $\boldsymbol{f} = \boldsymbol{\mu_{7}}$):
    \begin{equation} \label{eq:restaining1}
        q_{+}^{\left(f,f\right)} = \frac{1}{2} \left(\boldsymbol{TI_{nq}} + \boldsymbol{f} \oast \boldsymbol{TI_{nq}} \oast \boldsymbol{f} \right)
    \end{equation}

\begin{table}[t]
    \centering
    \caption{\raggedright{\textbf{Workflow 2}: Re-staining, 2 colors}}
    \vspace*{-0.75mm}
    \scriptsize
    \renewcommand{\arraystretch}{0.98}
    \begin{tabular*}{\linewidth}{@{\extracolsep{3pt}}l l@{}}
        \toprule
        \textbf{Require}: & targeted re-colorization of color image $I_C$ \\
        \textbf{Input}: & color image (24-bit, RGB) $I_C$ \\
        \textbf{Output}: & transformed color image (24-bit, RGB) $I_{T}$ \\
        \noalign{\vskip 0.5mm}
        \hline
        \noalign{\vskip 0.5mm}
        1: & Quaternion representation as $\boldsymbol{I_{q}}$ \\
        2: & Pre-processing operation according to Eq. \ref{eq:transform} \\
        3: & Extraction of color channel matrices \\
        4: & Data normalization jointly for all color channels in $[0,1]$ \\
        5: & Quaternion representation as $\boldsymbol{TI_{nq}}$ \\
        6: & 2D orthogonal planes split according to Eq. \ref{eq:restaining1} \\
        7: & Extraction of color channel matrices $ccA$ and $ccB$ \\
        8: & Histogram operation on color channels $ccA$ and $ccB$ \\
        9: & Quaternion representation as $\boldsymbol{I_{ccA}}$ and $\boldsymbol{I_{ccB}}$ \\
        10: & 2D orthogonal planes split according to Eq. \ref{eq:restaining2} \\
        11: & Extraction of color channels $ccA_{n}$ and $ccB_{n}$ \\
        12: & Data truncation for color channels $ccA_{n}$ and $ccB_{n}$ in $[0,1]$ \\
        13: & Bitwise OR: $ccAB_{n} = ccA_{n} \lor ccB_{n}$ \\
        \bottomrule
    \end{tabular*}
\label{tab:workflow_restain1}
\end{table}

\begin{table}[t]
    \centering
    \caption{\raggedright{\textbf{Workflow 3}: Re-staining, 3 colors}}
    \vspace*{-0.75mm}
    \scriptsize
    \renewcommand{\arraystretch}{0.98}
    \begin{tabular*}{\linewidth}{@{\extracolsep{3pt}}l l@{}}
        \toprule
        \textbf{Require}: & targeted re-colorization of color image $I_C$ \\
        \textbf{Input}: & color image (24-bit, RGB) $I_C$ \\
        \textbf{Output}: & transformed color image (24-bit, RGB) $I_{T}$ \\
        \noalign{\vskip 0.5mm}
        \hline
        \noalign{\vskip 0.5mm}
        1: & Quaternion representation as $\boldsymbol{I_{q}}$ \\
        2: & Pre-processing operation according to Eq. \ref{eq:transform} \\
        3: & Extraction of color channel matrices \\
        4: & Data normalization jointly for all color channels in $[0,1]$ \\
        5: & Quaternion representation as $\boldsymbol{TI_{nq}}$ \\
        6: & 2D orthogonal planes split according to Eq. \ref{eq:restaining1} \\
        7: & Extraction of color channel matrix $ccA$ \\
        8: & Re-colorization operations on $I_C$ \\
        9: & Repeat steps 6-7 on re-colorized images \\
        10: & Extraction of color channel matrices $ccB$ and $ccC$ \\
        11: & Histogram operation on color channels $ccA$, $ccB$, and $ccC$ \\
        12: & Quaternion representation as $\boldsymbol{I_{ccA}}$, $\boldsymbol{I_{ccB}}$, and $\boldsymbol{I_{ccC}}$ \\
        13: & 2D orthogonal planes split according to Eq. \ref{eq:restaining2} \\
        14: & Extraction of color channel matrices $ccA_{n}$, $ccB_{n}$, and $ccC_{n}$ \\
        15: & Data truncation for channels $ccA_{n}$, $ccB_{n}$, and $ccC_{n}$ in $[0,1]$ \\
        16: & Bitwise OR: $ccAB_{n} = ccA_{n} \lor ccB_{n}$, $ccABC_{n} = ccAB_{n} \lor ccC_{n}$ \\
        \bottomrule
    \end{tabular*}
\label{tab:workflow_restain2}
\end{table}

\noindent
where $\oast$ is the quaternion multiplication operator. Depending on the histological image, different color channels are retained as output (corresponding to the stains). Therefore depending on the image, color channel matrices $ccA$ and $ccB$ $\in$ $\mathbb{R}$ are extracted from the resulting quaternion matrix of Eq. \ref{eq:restaining1}. Scalar parts are retained during computations but are discarded prior to data truncation which is carried out in the range $[0,1]$. The histograms of $ccA$ and $ccB$ are then linearly scaled: intensity values between 0 and some value $X$ (that depends on the image, e.g. 0.15-0.50) are mapped to values between 0 and 1. Then, $ccA$ and $ccB$ are converted to quaternion matrices by placing each color channel threefold into the three quaternionic imaginary parts ($i$, $j$, $k$), e.g. $\boldsymbol{I_{ccA}} = ccA(n,m)i+ccA(n,m)j+ccA(n,m)k$ and $\boldsymbol{I_{ccB}} = ccB(n,m)i+ccB(n,m)j+ccB(n,m)k$. A 2D orthogonal planes split step is carried out to $\boldsymbol{I_{ccA}}$ and $\boldsymbol{I_{ccB}}$ $\in$ $\mathbb{H}$ with respect to user-defined pure quaternion $\boldsymbol{f}$:
    \begin{equation} \label{eq:restaining2}
    \begin{split}
        & q_{-}^{\left(f,f\right)} = \frac{1}{2} \left(\boldsymbol{I_{ccA}} - \boldsymbol{f} \oast \boldsymbol{I_{ccA}} \oast \boldsymbol{f} \right) \\
        & q_{-}^{\left(f,f\right)} = \frac{1}{2} \left(\boldsymbol{I_{ccB}} - \boldsymbol{f} \oast \boldsymbol{I_{ccB}} \oast \boldsymbol{f} \right)
    \end{split}
    \end{equation}

\noindent
where $\oast$ is the quaternion multiplication operator. Regarding Fig. \ref{fig:img_restain_bio}(a1,b1,c1), the map $\boldsymbol{f}()\boldsymbol{f}$ is user-defined for the green ($\boldsymbol{f} = \boldsymbol{ud_{1}} = 0i + \frac{255}{255}j + 0k$) and red color ($\boldsymbol{f} = \boldsymbol{ud_{2}} = \frac{255}{255}i + 0j + 0k$). Regarding Fig. \ref{fig:img_restain_bio}(a2,b2), the map $\boldsymbol{f}()\boldsymbol{f}$ is user-defined for the shades of blue ($\boldsymbol{f} = \boldsymbol{ud_{1}} =  \frac{30}{255}i + \frac{144}{255}j + \frac{255}{255}k$) and pink color ($\boldsymbol{f} = \boldsymbol{ud_{2}} = \frac{218}{255}i + \frac{70}{255}j + \frac{125}{255}k$). Regarding Fig. \ref{fig:img_restain_bio}(c2), the map $\boldsymbol{f}()\boldsymbol{f}$ is user-defined for the shades of blue ($\boldsymbol{f} = \boldsymbol{ud_{1}} = \frac{30}{255}i + \frac{144}{255}j + \frac{255}{255}k$) and orange color ($\boldsymbol{f} = \boldsymbol{ud_{2}} = \frac{210}{255}i + \frac{125}{255}j + \frac{45}{255}k$). The symbol $()$ refers to a color pixel represented as a pure quaternion. Color channel matrices $ccA_{n}$ and $ccB_{n}$ $\in$ $\mathbb{R}$ are extracted from the resulting quaternion matrix of Eq. \ref{eq:restaining2}. Scalar parts are retained during computations but are discarded prior to data truncation which is carried out in the range $[0,1]$. Prior to image encoding, a bitwise OR operation is carried out, such that:  $ccAB_{n} = ccA_{n} \lor ccB_{n}$. Thus, the transformed color image is obtained: $I_{T} = ccAB_{n}$ (24-bit).

For histological images with more than two dominant colors (e.g. double immunostain with counterstain, Fig. \ref{fig:img_restain_bio}(d)), the initial pre-processing step of Eq. \ref{eq:transform} is carried out. The process continues as previously shown with Eq. \ref{eq:restaining1}, and the subsequent histogram operation on color channel $ccA$. Given that the number of dominant colors is more than two, additional re-colorization steps are required to highlight the remaining colors. This is especially true in cases where the colors do not have a sufficient distance in the hue values (angular position on the HSV color space coordinate diagram), e.g. red and brown shades. The number of required re-colorization steps is equal to $(n_{c}-1)$ where $n_{c}$ is the number of dominant colors. As an example, for the double immunostain with counterstain (3 dominant colors, Fig. \ref{fig:img_restain_bio}(d)), the number of re-colorization steps is two. After each re-colorization, all previous steps apply (except for the image pre-processing step) for obtaining color channel matrices $ccB$ and $ccC$. Color channels $ccA$, $ccB$, and $ccC$ $\in$ $\mathbb{R}$ are then converted to quaternion matrices by placing each color channel threefold into the three quaternionic imaginary parts as previously shown. A 2D orthogonal planes split step is carried out to $\boldsymbol{I_{ccA}}$, $\boldsymbol{I_{ccB}}$, and $\boldsymbol{I_{ccC}}$ with respect to user-defined pure quaternion $\boldsymbol{f}$ as in Eq. \ref{eq:restaining2}. Regarding Fig. \ref{fig:img_restain_bio}(d1), the map $\boldsymbol{f}()\boldsymbol{f}$ is user-defined for the shades of blue ($\boldsymbol{f} = \boldsymbol{ud_{1}} = \frac{30}{255}i + \frac{144}{255}j + \frac{255}{255}k$), pink ($\boldsymbol{f} = \boldsymbol{ud_{2}} = \frac{218}{255}i + \frac{70}{255}j + \frac{125}{255}k$), and orange color ($\boldsymbol{f} = \boldsymbol{ud_{3}} = \frac{210}{255}i + \frac{125}{255}j + \frac{45}{255}k$). Regarding Fig. \ref{fig:img_restain_bio}(d2), the map $\boldsymbol{f}()\boldsymbol{f}$ is user-defined for the cyan ($\boldsymbol{f} = \boldsymbol{ud_{1}} = 0i + \frac{255}{255}j + \frac{255}{255}k$), magenta ($\boldsymbol{f} = \boldsymbol{ud_{2}} = \frac{255}{255}i + 0j + \frac{255}{255}k$), and yellow color ($\boldsymbol{f} = \boldsymbol{ud_{3}} = \frac{255}{255}i + \frac{255}{255}j + 0k$). The symbol $()$ refers to a color pixel represented as a pure quaternion. For both Fig. \ref{fig:img_restain_bio}(d1,d2), the first and second re-colorization steps are performed according to Eq. \ref{eq:recolorization} ($q_+$, $\boldsymbol{f}()\boldsymbol{g}$) with $\boldsymbol{f} = \boldsymbol{\mu_{10}}$, $\boldsymbol{g} = \boldsymbol{\mu_{11}}$ and $\boldsymbol{f} = \boldsymbol{\mu_{7}}$, $\boldsymbol{g} = \boldsymbol{\mu_{8}}$, respectively. Color channel matrices $ccA_{n}$, $ccB_{n}$, and $ccC_{n}$ $\in$ $\mathbb{R}$ are extracted from the resulting quaternion matrix. Scalar parts are retained during computations but are discarded prior to data truncation which is carried out in the range $[0,1]$. Prior to image encoding, two bitwise OR operations are carried out, such that: $ccAB_{n} = ccA_{n} \lor ccB_{n}$, and $ccABC_{n} = ccAB_{n} \lor ccC_{n}$. Thus, the transformed color image is obtained: $I_{T} = ccABC_{n}$ (24-bit). In case of stain co-localization in the tissue, the bitwise OR operation could be replaced by image blending operations depending on the desired visualization outcome, e.g. color channel layers on top of each other.

\subsection*{\normalsize{De-colorization}}
Results are shown in Figs. \ref{fig:img_decol1}-\ref{fig:img_decol2}, and Table \ref{tab:decol_comp}. Table \ref{tab:workflow_decol} provides an overview of the computational workflow. Let $\boldsymbol{I_{q}}$ be the pure quaternionic matrix representation of $I_C$ with size $[n \times m]$. The 2D orthogonal planes split is carried out with respect to a pure unit quaternion $\boldsymbol{f}$:
    \begin{equation} \label{eq:decolorization}
        q_{-}^{\left(f,f\right)} = \frac{1}{2} \left(\boldsymbol{I_{q}} - \boldsymbol{f} \oast \boldsymbol{I_{q}} \oast \boldsymbol{f} \right)
    \end{equation}

\noindent
where $\oast$ is the quaternion multiplication operator. The map $\boldsymbol{f}()\boldsymbol{f}$ can be $\boldsymbol{f} = \boldsymbol{\mu_{7}}$, but also a combination of values represented as pure unit quaternions that are defined by the user (e.g. pixel or area sampling) or computed by an automated method. The symbol $()$ refers to a color pixel represented as a pure quaternion. Color channel matrices $\in$ $\mathbb{R}$ are extracted from the resulting quaternion matrix of Eq. \ref{eq:decolorization} and their arithmetic mean is obtained. Scalar parts are retained during computations but are discarded prior to image encoding. For image encoding, data normalization is carried out in the range $[0,1]$. Thus, the transformed grayscale image is obtained: $I_{G}$ (8-bit).

\begin{table}[t]
    \centering
    \caption{\raggedright{\textbf{Workflow 4}: De-colorization}}
    \vspace*{-0.75mm}
    \scriptsize
    \renewcommand{\arraystretch}{0.98}
    \begin{tabular*}{\linewidth}{@{\extracolsep{3pt}}l l@{}}
        \toprule
        \textbf{Require}: & de-colorization of color image $I_C$ \\
        \textbf{Input}: & color image (24-bit, RGB) $I_C$ \\
        \textbf{Output}: & transformed grayscale image (8-bit) $I_{G}$ \\
        \noalign{\vskip 0.5mm}
        \hline
        \noalign{\vskip 0.5mm}
        1: & Quaternion representation as $\boldsymbol{I_{q}}$ \\
        2: & 2D orthogonal planes split according to Eq. \ref{eq:decolorization} \\
        3: & Extraction of color channel matrices \\
        4: & Data normalization for each color channel in $[0,1]$ \\
        5: & Arithmetic mean of transformed color channels \\
        \bottomrule
    \end{tabular*}
\label{tab:workflow_decol}
\end{table}

Regarding the de-colorization methods of Table \ref{tab:decol_comp}, proposed method P1 is based on the following: $q_-$, $\boldsymbol{f} = \boldsymbol{\mu_{7}}$. Method P2a is based on defining fixed color channel weights; $q_-$, $\boldsymbol{p} = 0.30i + 0.50j + 0.05k$, $\boldsymbol{f} = \boldsymbol{p}/\left| \boldsymbol{p} \right |$. These weights reflect the different recommendations for computing grayscale values \cite{InternationalTelecommunicationUnion2011, ITU-R_BT709, ITU-R_BT.2100}.

Method P2b is based on metrics computed from each image. For this purpose, the 1-norm (maximum absolute column sum) values of each $[n \times m]$ input color channel $cc1$ (red), $cc2$ (green), and $cc3$ (blue) $\in$ $\mathbb{R}$, are computed such as: 
    \begin{equation} \label{eq:1-norm}
    \begin{aligned}
        &\norm{cc1}_{1} = \nicefrac{\left( \max_{1 \leq j \leq n} \sum_{i=1}^m |cc1_{ij}| \right)}{255} \\ 
        &\norm{cc2}_{1} = \nicefrac{\left( \max_{1 \leq j \leq n} \sum_{i=1}^m |cc2_{ij}| \right)}{255} \\
        &\norm{cc3}_{1} = \nicefrac{\left( \max_{1 \leq j \leq n} \sum_{i=1}^m |cc3_{ij}| \right)}{255}
    \end{aligned}
    \end{equation}
    
\noindent
The 2-norm (spectral norm) values of each $[n \times m]$ input color channel $cc1$ (red), $cc2$ (green), and $cc3$ (blue) $\in$ $\mathbb{R}$, are computed such as (with $\sigma_{\text{max}}()$ representing the largest singular value): 
    \begin{equation} \label{eq:2-norm}
    \begin{aligned}
        &\norm{cc1}_{2} = \nicefrac{\left[ \sigma_{\text{max}}(cc1) \right]}{255} \\ 
        &\norm{cc2}_{2} = \nicefrac{\left[ \sigma_{\text{max}}(cc2) \right]}{255} \\
        &\norm{cc3}_{2} = \nicefrac{\left[ \sigma_{\text{max}}(cc3) \right]}{255}
    \end{aligned}
    \end{equation}

\noindent
The Frobenius norm \cite{Golub1996} values of each $[n \times m]$ input color channel $cc1$ (red), $cc2$ (green), and $cc3$ (blue) $\in$ $\mathbb{R}$, are computed such as: 
    \begin{equation} \label{eq:frobeniusnorm}
    \begin{aligned}
        &\norm{cc1}_{F} = \nicefrac{\sqrt{\left( \sum_{i=1}^{m} \sum_{j=1}^{n} \left| cc1_{ij} \right|^2 \right)}}{255} \\ 
        &\norm{cc2}_{F} = \nicefrac{\sqrt{\left( \sum_{i=1}^{m} \sum_{j=1}^{n} \left| cc2_{ij} \right|^2 \right)}}{255} \\ 
        &\norm{cc3}_{F} = \nicefrac{\sqrt{\left( \sum_{i=1}^{m} \sum_{j=1}^{n} \left| cc3_{ij} \right|^2 \right)}}{255}
    \end{aligned}
    \end{equation}

\noindent
The arithmetic mean values of each $[n \times m]$ normalized ($[0,1]$) input color channel $cc1$ (red), $cc2$ (green), and $cc3$ (blue) $\in$ $\mathbb{R}$, are computed such as:
    \begin{equation} \label{eq:meanvalues}
    \begin{aligned}
        &cc1_{mn} = \nicefrac{\left( \frac{1}{mn} \sum_{i=1}^{m} \sum_{j=1}^{n} cc1_{ij} \right)}{255} \\ 
        &cc2_{mn} = \nicefrac{\left( \frac{1}{mn} \sum_{i=1}^{m} \sum_{j=1}^{n} cc2_{ij} \right)}{255} \\
        &cc3_{mn} = \nicefrac{\left( \frac{1}{mn} \sum_{i=1}^{m} \sum_{j=1}^{n} cc3_{ij} \right)}{255}
    \end{aligned}
    \end{equation}

\noindent
Then, the 1-norm, 2-norm, Frobenius norm, and arithmetic mean values are further computed with the following:
    \begin{equation} \label{eq:altogether}
    \scalebox{0.85}{\(\displaystyle
    \begin{aligned}
        &cc1_{mn} = \left( f_{a} \times cc1_{mn} \right) + \left( cc1_{mn} \times cc2_{mn} \times cc3_{mn} \right) \\ 
        &cc2_{mn} = \left( f_{a} \times cc2_{mn} \right) + \left( cc1_{mn} \times cc2_{mn} \times cc3_{mn} \right) \\
        &cc3_{mn} = \left( f_{a} \times cc3_{mn} \right) + \left( cc1_{mn} \times cc2_{mn} \times cc3_{mn} \right) \\
        &\norm{cc1}_{1} = \left( f_{a} \times \norm{cc1}_{1} \right) + \left( \norm{cc1}_{1} \times \norm{cc2}_{1} \times \norm{cc3}_{1} \right) \\ 
        &\norm{cc2}_{1} = \left( f_{a} \times \norm{cc2}_{1} \right) + \left( \norm{cc1}_{1} \times \norm{cc2}_{1} \times \norm{cc3}_{1} \right) \\
        &\norm{cc3}_{1} = \left( f_{a} \times \norm{cc3}_{1} \right) + \left( \norm{cc1}_{1} \times \norm{cc2}_{1} \times \norm{cc3}_{1} \right) \\
        &\norm{cc1}_{2} = \left(f_{a} \times \norm{cc1}_{2} \right) + \left( \norm{cc1}_{2} \times \norm{cc2}_{2} \times \norm{cc3}_{2} \right) \\ 
        &\norm{cc2}_{2} = \left( f_{a} \times \norm{cc2}_{2} \right) + \left( \norm{cc1}_{2} \times \norm{cc2}_{2} \times \norm{cc3}_{2} \right) \\
        &\norm{cc3}_{2} = \left( f_{a} \times \norm{cc3}_{2} \right) + \left( \norm{cc1}_{2} \times \norm{cc2}_{2} \times \norm{cc3}_{2} \right) \\
        &\norm{cc1}_{F} = \left( f_{a} \times \norm{cc1}_{F} \right) + \left( \norm{cc1}_{F} \times \norm{cc2}_{F} \times \norm{cc3}_{F} \right) \\
        &\norm{cc2}_{F} = \left( f_{a} \times \norm{cc2}_{F} \right) + \left( \norm{cc1}_{F} \times \norm{cc2}_{F} \times \norm{cc3}_{F} \right) \\
        &\norm{cc3}_{F} = \left( f_{a} \times \norm{cc3}_{F} \right) + \left( \norm{cc1}_{F} \times \norm{cc2}_{F} \times \norm{cc3}_{F} \right)
    \end{aligned}
    \)}
    \end{equation}

\noindent
Finally, the encoding values are computed per image using the following: 
    \begin{equation} \label{eq:encoding}
    \begin{aligned}
        &ev_{1} = f_{1} \times \left( cc1_{mn} \times \norm{cc1}_{1} + \norm{cc1}_{2} \times \norm{cc1}_{F} \right)^{f_{b}} \\
        &ev_{2} = f_{2} \times \left( cc2_{mn} \times \norm{cc2}_{1} + \norm{cc2}_{2} \times \norm{cc2}_{F} \right)^{f_{b}} \\
        &ev_{3} = f_{3} \times \left( cc3_{mn} \times \norm{cc3}_{1} + \norm{cc3}_{2} \times \norm{cc3}_{F} \right)^{f_{b}}
    \end{aligned}
    \end{equation}

\noindent
The fixed factors ($f_{a} = 0.68$, $f_{b} = 1.80$, $f_{1} = 0.36$, $f_{2} = 1.30$, and $f_{3} = 0.07$) were optimized for the COLOR250 dataset \cite{Lu2014}. Since the dataset contains a variety of images such as natural images, digital charts, logos, and illustrations, it is reasonable to assume that these values would work fairly well with other images or datasets. The encoding values (computed per image) are utilized in the computations as pure unit quaternions ($q_-$, $\boldsymbol{p} = ev_{1}i + ev_{2}j + ev_{3}k$, $\boldsymbol{f} = \boldsymbol{p}/\left| \boldsymbol{p} \right |$).

\subsection*{\normalsize{Contrast enhancement}}
Results are shown in Table \ref{tab:contr_comp} and Fig. \ref{fig:img_contr}. Table \ref{tab:workflow_contrenhanc} provides an overview of the computational workflow. Let $\boldsymbol{I_{q}}$ be the pure quaternionic matrix representation of $I_C$ with size $[n \times m]$. Eq. \ref{eq:recolorization} is modified in order to avoid chrominance and luminance distortions. Contrast enhancement is performed using the following:
    \begin{equation} \label{eq:contrastenhancement1}
        q_{+}^{\left(c_{u},c_{l}\right)} = \frac{1}{2} \left[h(\boldsymbol{I_{q}}) + \boldsymbol{c_{u}} \oast \boldsymbol{I_{q}} \oast \boldsymbol{c_{l}} \right]
    \end{equation}

\noindent
where $\oast$ is the quaternion multiplication operator. The terms $\mathbf{c_{u}} = 1i + 1j + 1k$ and $\mathbf{c_{l}} = 0.01i + 0.01j + 0.01k$ are pure quaternions. The term $h(\boldsymbol{I_{q}})$ is constructed in a modular fashion where the individual sub-terms provide sequential contrast adjustments and refinements, namely: 
    \begin{equation} \label{eq:contrastenhancement2}
    \begin{aligned}
        h(\boldsymbol{I_q}) = \alpha &\left[ \frac{\boldsymbol{I_q}}{2} \right] - \beta \left[ \frac{1}{e^{ \boldsymbol{I_q}}} \right] + \\
        & +\gamma \left[ \frac{e^{ \boldsymbol{I_q}}}{ \boldsymbol{I_q} + e^{ \boldsymbol{I_q}}} \right] + \delta \left[ \frac{\boldsymbol{I_q}^{1/2}}{\boldsymbol{I_q} + e^{\boldsymbol{I_q}}} \right]
    \end{aligned}
    \end{equation}

\noindent
The parameters $\alpha$, $\beta$, $\gamma$, and $\delta$ can be tuned for different types of images. For the CEED2016 dataset \cite{Qureshi2016, Qureshi2017} (natural color images) and histology images, the parameters have the following values: $\alpha=10$, $\beta=1$, $\gamma=-1$, and $\delta=1$. For the COVID-LDCT dataset \cite{Afshar2021} (medical grayscale images), the parameters have the following values: $\alpha=1$, $\beta=10^5$, $\gamma=10^4$, and $\delta=10^4$. Eq. \ref{eq:contrastenhancement1} contains real multiplications, quaternion additions/subtractions, and component-wise multiplications/divisions. Color channel matrices $\in$ $\mathbb{R}$ are extracted from the resulting quaternion matrix of Eq. \ref{eq:contrastenhancement1}. Scalar parts are retained during computations but are discarded prior to image encoding. For image encoding, data normalization is carried out for each color channel in the range $[0,1]$. Thus, the transformed color image is obtained: $I_{T}$ (24-bit).

\begin{table}[t]
    \centering
    \caption{\raggedright{\textbf{Workflow 5}: Contrast enhancement}}
    \vspace*{-0.75mm}
    \scriptsize
    \renewcommand{\arraystretch}{0.98}
    \begin{tabular*}{\linewidth}{@{\extracolsep{3pt}}l l@{}}
        \toprule
        \textbf{Require}: & enhance contrast of color image $I_C$ \\
        \textbf{Input}: & color image (24-bit, RGB) $I_C$ \\
        \textbf{Output}: & transformed color image (24-bit, RGB) $I_{T}$ \\
        \noalign{\vskip 0.5mm}
        \hline
        \noalign{\vskip 0.5mm}
        1: & Quaternion representation as $\boldsymbol{I_{q}}$ \\
        2: & 2D orthogonal planes split according to Eq. \ref{eq:contrastenhancement1} using Eq. \ref{eq:contrastenhancement2} \\
        3: & Extraction of color channel matrices \\
        4: & Data normalization for each color channel in $[0,1]$ \\
        5: & Concatenation of transformed color channels \\
        \bottomrule
    \end{tabular*}
\label{tab:workflow_contrenhanc}
\end{table}

\subsection*{\normalsize{Stain separation}}
Results are shown in Figs. \ref{fig:img_unsupstainsep_bio}, \ref{fig:img_supstainsep_bio}, and \ref{fig:img_stainsep_bio}, and Tables \ref{tab:Hstain_sep_comp}-\ref{tab:Estain_sep_comp}. Tables \ref{tab:workflow_stainsep1} and \ref{tab:workflow_stainsep2} provide overviews of the computational workflows. For histological images with two dominant colors (e.g. H\&E or single immunostain with counterstain, Fig. \ref{fig:img_unsupstainsep_bio}(a-d) and Fig. \ref{fig:img_supstainsep_bio}(a-d)), let $\boldsymbol{I_{q}}$ be the pure quaternionic matrix representation of $I_C$ with size $[n \times m]$. An image pre-processing step is carried out; the input image $\boldsymbol{I_{q}}$ is transformed to $TI_{q}$ using Eq. \ref{eq:transform}. Color channel matrices $\in$ $\mathbb{R}$ are extracted from the resulting quaternion matrix of Eq. \ref{eq:transform}. To obtain the output, scalar parts are retained during computations but are discarded prior to image encoding. For image encoding, data normalization is carried out for all color channels jointly in the range $[0,1]$. The resulting image is transformed to a quaternion matrix $\boldsymbol{TI_{nq}}$, and then the 2D orthogonal planes split is carried out with respect to a pure unit quaternion/pure quaternion $\boldsymbol{f}$ or any two pure unit quaternions/pure quaternions $\boldsymbol{f}$ and $\boldsymbol{g}$:
    \begin{equation} \label{eq:stainseparation}
    \begin{aligned}
        &q_{+}^{\left(f,f\right)} = \frac{1}{2} \left(\boldsymbol{TI_{nq}} + \boldsymbol{f} \oast \boldsymbol{TI_{nq}} \oast \boldsymbol{f} \right) \\
        &q_{+}^{\left(f,g\right)} = \frac{1}{2} \left(\boldsymbol{TI_{nq}} + \boldsymbol{f} \oast \boldsymbol{TI_{nq}} \oast \boldsymbol{g} \right)
    \end{aligned}
    \end{equation}

\noindent
where $\oast$ is the quaternion multiplication operator. The symbol $()$ refers to a color pixel represented as a pure quaternion. The map can be $\boldsymbol{f}()\boldsymbol{f}$ where $\boldsymbol{f} = \boldsymbol{\mu_{7}}$, but also a map $\boldsymbol{f}()\boldsymbol{g}$ based on a combination of values represented as pure quaternions that are defined by the user (e.g. pixel or area sampling) or computed by an automated method. Fig. \ref{fig:img_unsupstainsep_bio} shows results from transformations that have geometric meaning ($\boldsymbol{\mu_{7}}$). For Fig. \ref{fig:img_supstainsep_bio}, a computational method is utilized to estimate the stain separation matrix (Macenko's method), meaning that the algorithm can compute the stain vectors of a histological image \cite{Macenko2009, stain_normalisation_toolbox_v2_2}. The two triplets of the output stain separation matrix are encoded as two pure quaternions ($\boldsymbol{f}=\boldsymbol{s_{1}}$ and $\boldsymbol{g}=\boldsymbol{s_{2}}$). Color channel matrices $ccA$, $ccB$, and $ccC$ $\in$ $\mathbb{R}$ are extracted from the resulting quaternion matrix of Eq. \ref{eq:stainseparation}. Scalar parts are retained during computations but are discarded prior to data truncation which is carried out in the range $[0,1]$. Thus, the output grayscale images are obtained: $I_{Gs_{1}}$ and $I_{Gs_{2}}$ (8-bit). Depending on the stains present in the histological image, different color channels are retained as output, e.g. for H\&E in Fig. \ref{fig:img_supstainsep_bio}(a), $ccA$ and $ccC$ are written as output corresponding to the red and blue color channels, respectively.

\begin{table}[t]
    \centering
    \caption{\raggedright{\textbf{Workflow 6}: Stain separation, 2 colors}}
    \vspace*{-0.75mm}
    \scriptsize
    \renewcommand{\arraystretch}{0.98}
    \begin{tabular*}{\linewidth}{@{\extracolsep{3pt}}l l@{}}
        \toprule
        \textbf{Require}: & stain separation of color image $I_C$ \\
        \textbf{Input}: & color image (24-bit, RGB) $I_C$ \\
        \textbf{Output}: & grayscale images (8-bit) $I_{Gs_{1}}$, $I_{Gs_{2}}$ \\
        \noalign{\vskip 0.5mm}
        \hline
        \noalign{\vskip 0.5mm}
        1: & Quaternion representation as $\boldsymbol{I_{q}}$ \\
        2: & Pre-processing operation according to Eq. \ref{eq:transform} \\
        3: & Extraction of color channel matrices \\
        4: & Data normalization jointly for all color channels in $[0,1]$ \\
        5: & Quaternion representation as $\boldsymbol{TI_{nq}}$ \\
        6: & 2D orthogonal planes split according to Eq. \ref{eq:stainseparation} \\
        7: & Extraction of color channels $ccA$, $ccB$, and $ccC$ \\
        8: & Data truncation for color channels $ccA$, $ccB$, and $ccC$ in $[0,1]$ \\
        9: & Retention of color channels corresponding to stain information \\
        \bottomrule
    \end{tabular*}
\label{tab:workflow_stainsep1}
\end{table}

\begin{table}[t]
    \centering
    \caption{\raggedright{\textbf{Workflow 7}: Stain separation, 3 colors}}
    \vspace*{-0.75mm}
    \scriptsize
    \renewcommand{\arraystretch}{0.98}
    \begin{tabular*}{\linewidth}{@{\extracolsep{3pt}}l l@{}}
        \toprule
        \textbf{Require}: & stain separation of color image $I_C$ \\
        \textbf{Input}: & color image (24-bit, RGB) $I_C$ \\
        \textbf{Output}: & grayscale images (8-bit) $I_{Gs_{1}}$, $I_{Gs_{2}}$, $I_{Gs_{3}}$ \\
        \noalign{\vskip 0.5mm}
        \hline
        \noalign{\vskip 0.5mm}
        1: & Quaternion representation as $\boldsymbol{I_{q}}$ \\
        2: & Pre-processing operation according to Eq. \ref{eq:transform} \\
        3: & Extraction of color channel matrices \\
        4: & Data normalization jointly for all color channels in $[0,1]$ \\
        5: & Quaternion representation as $\boldsymbol{TI_{nq}}$ \\
        6: & 2D orthogonal planes split according to Eq. \ref{eq:stainseparation} \\
        7: & Extraction of color channels $ccA$, $ccB$, and $ccC$ \\
        8: & Data truncation for color channels $ccA$, $ccB$, and $ccC$ in $[0,1]$ \\
        9: & Retention of color channels corresponding to stain information \\
        10: & Re-colorization operations on $I_C$ \\
        11: & Repeat steps 1-9 on re-colorized images \\
        \bottomrule
    \end{tabular*}
\label{tab:workflow_stainsep2}
\end{table}

On certain occasions, e.g. Fig. \ref{fig:img_unsupstainsep_bio}(c) and Fig. \ref{fig:img_supstainsep_bio}(c), stain separation is not optimal due to the limited effectiveness of the approach or the similarity of the values in the stain separation matrix for the two stains. One solution is to obtain a better stain separation matrix with another computational method. If this is not possible, then an additional step is required before stain separation. This step comprises computational re-staining as described in a previous section. More specifically, for Fig. \ref{fig:img_supstainsep_bio}(c) virtual re-staining is carried out to the original histological image with the final rendition utilizing: green (\#00FF00) and blue color (\#0000FF). The values in parentheses are the hexadecimal triplet representation of colors. The output from the virtual re-staining step is then utilized as input for the separation step, as described previously in this section.

For histological images with more than two dominant colors (e.g. double immunostain with counterstain, Fig. \ref{fig:img_supstainsep_bio}(e)), the initial pre-processing step of Eq. \ref{eq:transform} is carried out. For the counterstain (approx. blue color), the process continues as previously shown in this section using Eq. \ref{eq:stainseparation}. Given that the number of dominant colors is more than two, additional re-colorization steps are required to highlight the remaining colors. This is especially true in cases where the colors do not have a sufficient distance in the hue values (angular position on the HSV color space coordinate diagram), e.g. red and brown shades. The number of required re-colorization steps is equal to $(n_{c}-1)$ where $n_{c}$ is the number of dominant colors. As an example, for the double immunostain with counterstain (3 dominant colors, Fig. \ref{fig:img_supstainsep_bio}(e)), the number of re-colorization steps is two. For the immunostain that exhibits approx. brown color, the re-colorization step is performed according to Eq. \ref{eq:recolorization} ($q_+$) with $\boldsymbol{f} = \boldsymbol{\mu_{3}}$, $\boldsymbol{g} = \boldsymbol{\mu_{8}}$. For the immunostain that exhibits approx. red color, the re-colorization step is performed according to Eq. \ref{eq:recolorization} ($q_+$, $\boldsymbol{f}()\boldsymbol{g}$) with $\boldsymbol{f} = \boldsymbol{\mu_{8}}$, $\boldsymbol{g} = \boldsymbol{\mu_{10}}$. After each re-colorization step, all previous steps apply for obtaining color channel matrices $ccA$, $ccB$, and $ccC$ $\in$ $\mathbb{R}$. Depending on the stains present and the re-colorization step, the corresponding color channel is retained as output, thus the output grayscale images are obtained for the remaining two stains.

Regarding the stain separation methods of Tables \ref{tab:Hstain_sep_comp}-\ref{tab:Estain_sep_comp} and Fig. \ref{fig:img_stainsep_bio}, proposed method P1 is as follows: $q_+$, $\boldsymbol{f} = \boldsymbol{\mu_{7}}$. In method P2a, a computational method is utilized to estimate the stain separation matrix (Macenko's method) \cite{Macenko2009, stain_normalisation_toolbox_v2_2}, hence the two triplets of the output stain separation matrix are encoded as two pure quaternions ($q_+$, $\boldsymbol{f}()\boldsymbol{g}$, $\boldsymbol{f}=\boldsymbol{s_{1}}$ and $\boldsymbol{g}=\boldsymbol{s_{2}}$). In method P2b, a computational method is utilized to estimate the stain separation matrix (stain color descriptor method) \cite{Khan2014, stain_normalisation_toolbox_v2_2}, hence the two triplets of the output stain separation matrix are encoded as two pure quaternions ($q_+$, $\boldsymbol{f}()\boldsymbol{g}$, $\boldsymbol{f}=\boldsymbol{s_{3}}$ and $\boldsymbol{g}=\boldsymbol{s_{4}}$). For comparison purposes, 8-bit versions of the output images were obtained.

\subsection*{\normalsize{Machine/deep learning workflows}}
\begin{figure}[t]
\centering
\includegraphics[width=1.0\linewidth]{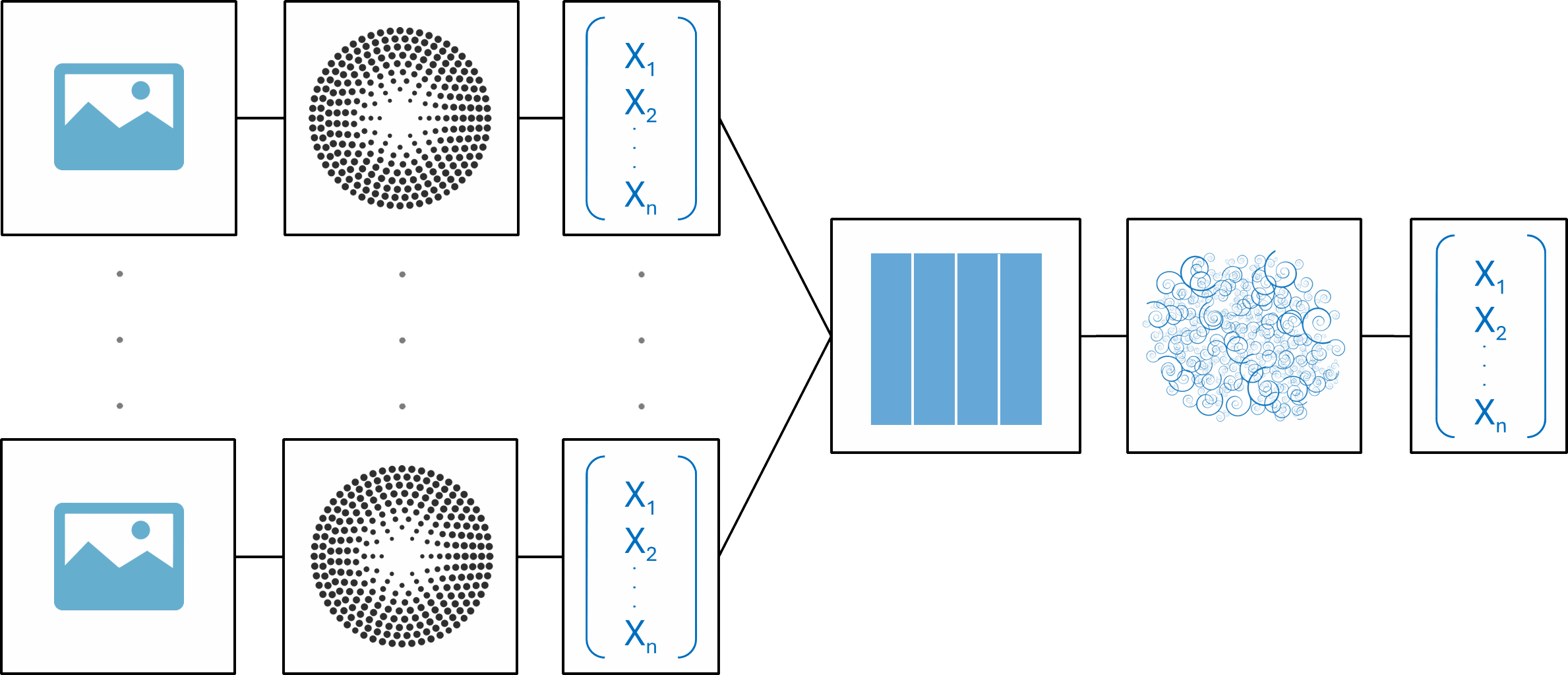}
\vspace{-12pt}
\caption{\footnotesize{Conceptual schematic showing: computation of the handcrafted features from the input images, aggregation of the feature sets, feature extraction using sparse filtering, and resulting output feature set. Results are shown in Table \ref{tab:ml_comp1}.}}
\label{fig:ml1}
\end{figure}

Results are shown in Tables \ref{tab:ml_comp1}, \ref{tab:ml_comp2}, and \ref{tab:ml_comp3}. Regarding Table \ref{tab:ml_comp1}, the machine learning methods (A-D) were the following, respectively: random subspace ensemble \cite{TinKamHo1998} (discriminant learner, 30 ensemble learning cycles), \textit{k}-nearest neighbors \cite{Sproull1991} (Euclidean distance, distance weighting function: inverse of the squared distance, 10 nearest neighbors), support vector machines \cite{ChihWeiHsu2002} (quadratic, one-vs-all design), and a feedforward fully connected neural network \cite{Liu1989} (one fully connected layer, first layer size: 100, ReLU activation function, loss function minimization technique: limited-memory BFGS quasi-Newton algorithm). The handcrafted features were the following: statistical descriptors \cite{RojasMoraleda2016} computed from wavelet coefficients by applying the 2D discrete wavelet transform, statistical descriptors \cite{RojasMoraleda2016} computed from wavelet coefficients by applying the dual-tree quaternionic 2D discrete wavelet transform \cite{WaiLamChan2008}, and textural descriptors from Gabor filtering \cite{Bianconi2007}; the total number of features was 1076. Wavelet decompositions can produce effective feature sets for image classification \cite{RojasMoraleda2016}. The data were split into training and test sets (75:25). Features were computed from 91 combinations of pure unit quaternions ($q_-$ and $q_+$, maps $\boldsymbol{f}()\boldsymbol{f}$ and $\boldsymbol{f}()\boldsymbol{g}$ where $\boldsymbol{f} = \boldsymbol{\mu_{i}}$ and $\boldsymbol{g} = \boldsymbol{\mu_{j}}$, with $\boldsymbol{i} < \boldsymbol{j}$ and for $\boldsymbol{\mu_{1}}$ to $\boldsymbol{\mu_{13}}$). For example, valid combinations included $\boldsymbol{\mu_{3}}()\boldsymbol{\mu_{3}}$ or $\boldsymbol{\mu_{9}}()\boldsymbol{\mu_{9}}$, and $\boldsymbol{\mu_{3}}()\boldsymbol{\mu_{4}}$ or $\boldsymbol{\mu_{3}}()\boldsymbol{\mu_{6}}$, but not $\boldsymbol{\mu_{3}}()\boldsymbol{\mu_{2}}$. The features were then aggregated separately for $q_-$ and $q_+$; the two aggregated datasets comprised 97916 features each. Feature extraction on the aggregated feature sets was carried out using sparse filtering \cite{Ngiam2011} which mapped the input feature sets to new output feature sets with decreased dimensionality (standardized input data: \textit{z}-scores, $L^2$ regularization coefficient: 1, relative convergence tolerance on gradient norm: $10^{-4}$, absolute convergence tolerance on step size: $10^{-4}$). The output feature set was defined a priori for $q_-$ and $q_+$ to 1076 features each, for comparison purposes. The process (excluding the final machine learning steps) is visualized in Fig. \ref{fig:ml1}.

\begin{figure}[t]
\centering
\includegraphics[width=1.0\linewidth]{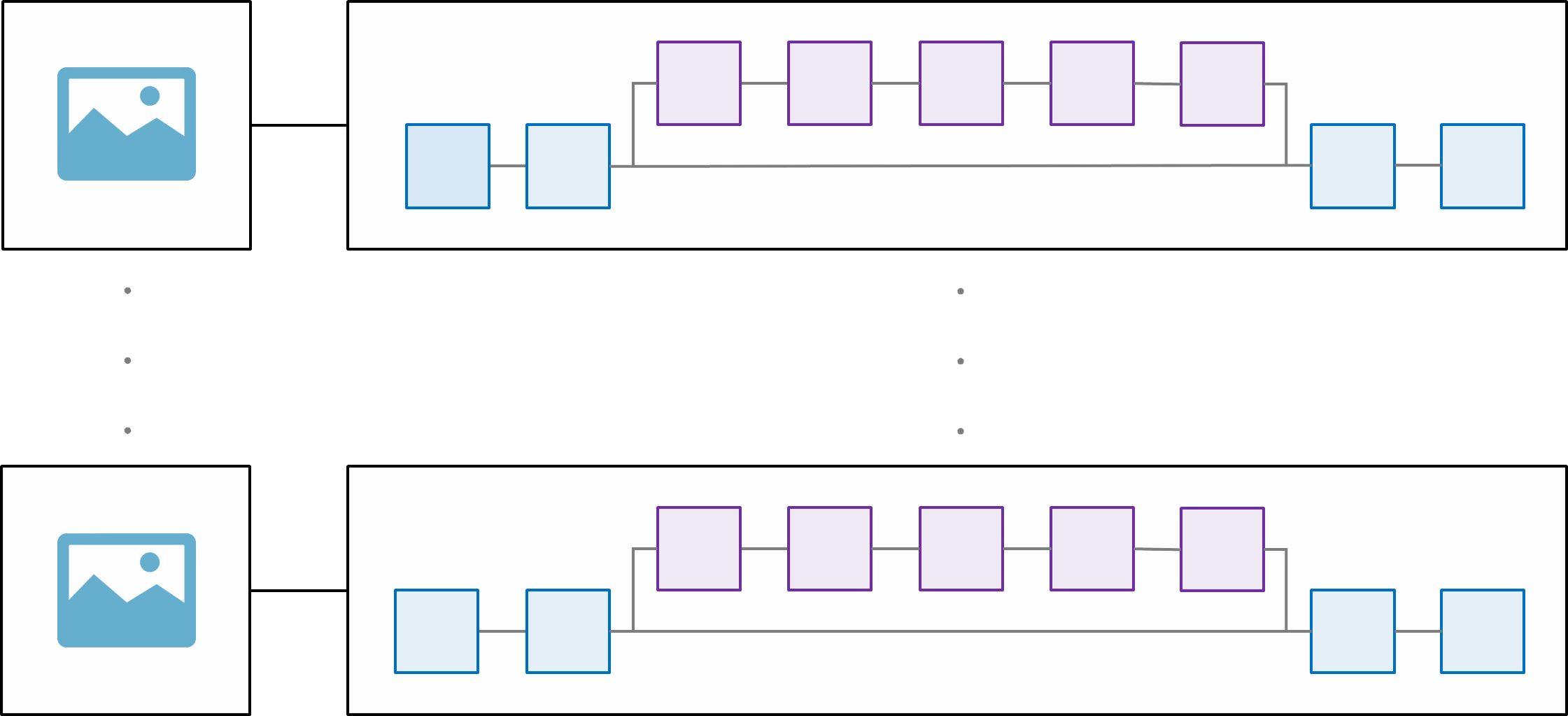}
\vspace{-12pt}
\caption{\footnotesize{Conceptual schematic showing that each transformation - computed from 91 combinations of pure unit quaternions ($q_-$ and $q_+$, maps $\boldsymbol{f}()\boldsymbol{f}$ and $\boldsymbol{f}()\boldsymbol{g}$ where $\boldsymbol{f} = \boldsymbol{\mu_{i}}$ and $\boldsymbol{g} = \boldsymbol{\mu_{j}}$, with $\boldsymbol{i} < \boldsymbol{j}$ and for $\boldsymbol{\mu_{1}}$ to $\boldsymbol{\mu_{13}}$) - was tested one by one using two deep learning pipelines. Results are shown in Table \ref{tab:ml_comp2}.}}
\label{fig:ml2}
\end{figure}

\begin{figure}[t]
\centering
\includegraphics[width=1.0\linewidth]{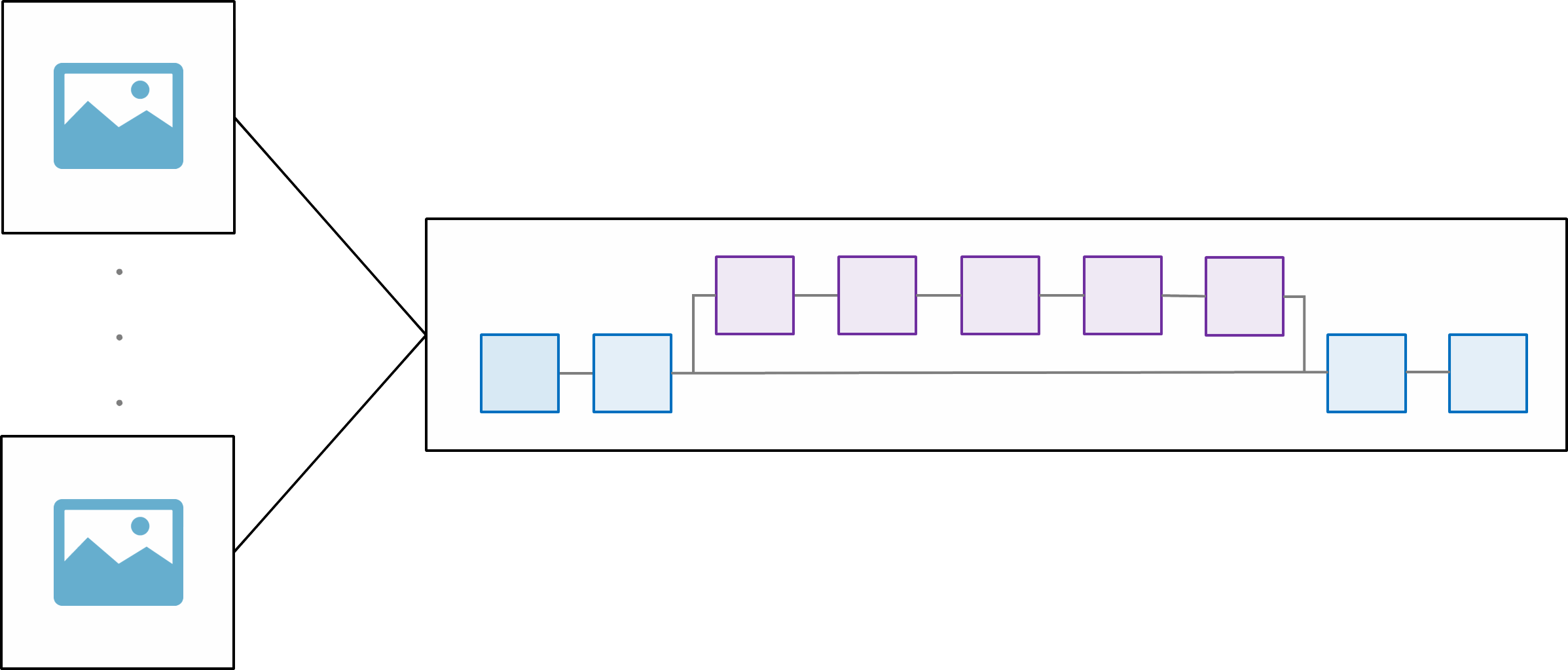}
\vspace{-12pt}
\caption{\footnotesize{Conceptual schematic showing transformed data - computed from 91 combinations of pure unit quaternions ($q_-$ and $q_+$, maps $\boldsymbol{f}()\boldsymbol{f}$ and $\boldsymbol{f}()\boldsymbol{g}$ where $\boldsymbol{f} = \boldsymbol{\mu_{i}}$ and $\boldsymbol{g} = \boldsymbol{\mu_{j}}$, with $\boldsymbol{i} < \boldsymbol{j}$ and for $\boldsymbol{\mu_{1}}$ to $\boldsymbol{\mu_{13}}$) - utilized as training data while original dataset was used as the test set; tests were carried out using two deep learning pipelines. Results are shown in Table \ref{tab:ml_comp3}.}}
\label{fig:ml3}
\end{figure}

Regarding Tables \ref{tab:ml_comp2}-\ref{tab:ml_comp3}, the ResNet-101 \cite{HeZRS15} and EfficientNet-b0 \cite{Tan2019EfficientNetRM} convolutional neural networks (CNNs) were utilized as untrained versions as well as pre-trained on the ImageNet dataset; 1.4M images and 1000 classes \cite{Deng2009}. The data were split into training and test sets (75:25). To prevent the CNNs from overfitting, data augmentation operations were performed on the training set: random flipping along the horizontal and vertical axis, random translation up to 30 pixels horizontally and vertically, random scaling up to 10$\%$ horizontally and vertically, and random rotation with an angle up to 90 degrees. The learnable parameters were updated using the stochastic gradient descent with the momentum algorithm. The untrained and pre-trained models may necessitate different training approaches; however, since the objective was not to compare classifiers and to maintain simplicity, the rest of the CNNs settings were the following: size of the mini-batch for each training iteration was set to 10, maximum number of epochs used for training was set to 6, initial learning rate used for training was set to \SI{3e-4}, and training data were shuffled before each epoch. 

For Table \ref{tab:ml_comp2}, each transformation - computed from 91 combinations of pure unit quaternions ($q_-$ and $q_+$, maps $\boldsymbol{f}()\boldsymbol{f}$ and $\boldsymbol{f}()\boldsymbol{g}$ where $\boldsymbol{f} = \boldsymbol{\mu_{i}}$ and $\boldsymbol{g} = \boldsymbol{\mu_{j}}$, with $\boldsymbol{i} < \boldsymbol{j}$ and for $\boldsymbol{\mu_{1}}$ to $\boldsymbol{\mu_{13}}$) - was assessed separately on the aforementioned untrained and pre-trained deep learning pipelines (Fig. \ref{fig:ml2}). For Table \ref{tab:ml_comp3}, all transformed data (56875 images per class) - computed from 91 combinations of pure unit quaternions ($q_-$ and $q_+$, maps $\boldsymbol{f}()\boldsymbol{f}$ and $\boldsymbol{f}()\boldsymbol{g}$ where $\boldsymbol{f} = \boldsymbol{\mu_{i}}$ and $\boldsymbol{g} = \boldsymbol{\mu_{j}}$, with $\boldsymbol{i} < \boldsymbol{j}$ and for $\boldsymbol{\mu_{1}}$ to $\boldsymbol{\mu_{13}}$) - were utilized as training data while the original dataset was used as the test set (Fig. \ref{fig:ml3}). Further tests were carried out with 50$\%$ and 80$\%$ of the transformed images being randomly deleted, resulting in a slightly unbalanced dataset.

Regarding Tables \ref{tab:ml_comp1}, \ref{tab:ml_comp2}, and \ref{tab:ml_comp3}, sensitivity or true positive rate is defined as $[TP/(TP+FN)]$, specificity or true negative rate as $[TN/(TN+FP)]$, precision or positive predictive value as $[TP/(TP+FP)]$, F1 score (harmonic mean of precision and sensitivity) as $[2TP/(2TP+FP+FN)]$, and fall-out or false positive rate as $[FP/(FP+TN)]$ \cite{Ting2017}, where $TP$ are the true positives, $TN$ the true negatives, $FP$ the false positives, and $FN$ the false negatives \cite{Tharwat2020}. Before employing the machine/deep learning pipelines to the transformed images, an additional step was carried out as previously mentioned in the re-colorization section. To avoid batch effects in the transformed images that can erroneously decrease or increase classification performance (due to the larger or smaller intra-class variance, respectively), data normalization was carried out for each color channel in the range $[0,1]$ on a larger histological exemplar image (24-bit, $5120 \times 2160$ pixels). The pairs comprising minimum (min) and maximum (max) values $[min,max]$ were computed very rapidly for the exemplar image for each combination ($q_-$ and $q_+$, maps $\boldsymbol{f}()\boldsymbol{f}$ and $\boldsymbol{f}()\boldsymbol{g}$ where $\boldsymbol{f} = \boldsymbol{\mu_{i}}$ and $\boldsymbol{g} = \boldsymbol{\mu_{j}}$, with $\boldsymbol{i} < \boldsymbol{j}$ and for $\boldsymbol{\mu_{1}}$ to $\boldsymbol{\mu_{13}}$); these pre-computed normalization ranges were then utilized for the image encoding step of the transformed images.

\subsection*{\normalsize{Time complexity}}
Results are shown in Fig. \ref{fig:time_compl}. Computing time experiments were carried out on three H\&E whole slide images (24-bit) with respective sizes (width$\times$height): $28000\times22193$, $28000\times18456$, and $22235\times28000$ pixels. These images were scaled down iteratively 19 times in 5$\%$ intervals using bicubic interpolation, maintaining the aspect ratio (ratio of image width to height) \cite{matlab_toolbox_ref}. Thus, after shrinking the images, twenty samples per histological image were available for the time complexity experiments: from the original size (mentioned previously) to the smaller image dimensions, respectively (width$\times$height): $1400\times1100$, $1400\times923$, and $1112\times1400$ pixels. The 2D orthogonal planes split was carried out (using re-colorization as an example) on these variably-sized images utilizing Eq. \ref{eq:recolorization} for $q_+$ and $q_-$, $\boldsymbol{f}()\boldsymbol{f}$, $\boldsymbol{f} = \boldsymbol{\mu_{7}}$. The elapsed time (s) was recorded only for running Eq. \ref{eq:recolorization} and did not include any other operations. Standard routines were utilized with no further optimizations.

\bigskip
\begin{center}
    \rule{0.40\linewidth}{0.5pt}
\end{center}

\bigskip
\noindent
\small{\textbf{Acknowledgments}. We wish to extend our gratitude to the scientists and mathematicians who provided feedback and comments during oral and poster presentations of this work.}

\medskip
\noindent
\small{{\textbf{Author contributions}: N.A.V. designed and performed research; E.H., D.D., R.R.M., F.P., M.S.-C., A.B., I.P., C.F., A.R., C.C.W., B.L., N.H., I.Z., and D.J. contributed resources and/or expertise in their respective domains including the provision of materials, wet-laboratory methods, clinical insights, and mathematical/computational reasoning; D.D. developed Python-based computational libraries; N.A.V. analyzed data and wrote the manuscript; All authors read, reviewed, and approved the manuscript.}}

\medskip
\noindent
\small{{\textbf{Declaration of interests}: N.A.V., N.H., I.Z., and D.J.: Segments of the work presented in the manuscript are protected by intellectual property rights and are covered under U.S. Patent No. 11,501,444 titled \textit{Method, Software Module, and Integrated Circuit for Color Visualization and Separation} (United States Patent and Trademark Office, 2022). The patented components pertain to image re-colorization and histological stain separation. The remaining authors declare no competing interests.}}

\renewcommand{\bibfont}{\footnotesize}
\bibliography{references}

\end{document}